%% file: main.tex
\documentclass{article}

\usepackage[preprint]{corl_2026} 

\usepackage{graphicx}
\usepackage{booktabs}
\usepackage{cuted}
\usepackage{caption}
\usepackage{multirow}
\usepackage{subfig}
\usepackage{wrapfig}
\usepackage{makecell}

\usepackage{amsmath}
\usepackage{amssymb}
\usepackage{mathtools}
\usepackage{bm}
\usepackage{color}
\usepackage{colortbl}
\usepackage{enumitem}
\usepackage{subcaption}

\definecolor{Gray}{gray}{0.9}

\newlength{\lhswidth}
\newcommand{\lhs}[1]{\makebox[\lhswidth][l]{\ensuremath{#1}}}

\usepackage[most]{tcolorbox}
\tcbuselibrary{listings,skins,breakable}

\definecolor{promptbg}{HTML}{F2F2F2}      
\definecolor{prompttitlebg}{HTML}{D9D9D9} 

\lstdefinestyle{promptstyle}{%
  basicstyle=\ttfamily\scriptsize,
  breaklines=true,
  breakatwhitespace=false,
  columns=fullflexible,
  keepspaces=true,
  upquote=true,
  morecomment=[l][\bfseries]{\#},
}

\newtcblisting{promptbox}[1]{%
  breakable, enhanced,
  colback=promptbg, colframe=prompttitlebg,
  colbacktitle=prompttitlebg, coltitle=black,
  fonttitle=\bfseries\ttfamily\small,
  title={#1},
  boxrule=0.4pt, arc=2pt,
  left=6pt, right=6pt, top=4pt, bottom=4pt,
  listing only,
  listing style=promptstyle,
}

\usepackage{hyperref}

\title{PRISM: Personalized Robotic Dataset Generation via Image-based Scene and Motion Synthesis}

\author{
  \textbf{Dogyu Ko}$^{*}$,
  \textbf{Haneul Kim}$^{*}$,
  \textbf{Chanyoung Yeo}$^{*}$,
  \textbf{Dowoon Lee},
  \textbf{Taeho Park},
  \textbf{Hyoseok Hwang}$^{\dagger}$ \\
  \\
  Kyung Hee University \\
  \texttt{\{kodogyu, sky9893, ducksdud08, pauldoun, katehoya, hyoseok\}@khu.ac.kr} \\
  $^{*}$Equal Contribution
  $^{\dagger}$Corresponding Author
}

\begin{document}
\maketitle


\vspace{-5pt}
\begin{figure}[htbp]
    \centering
    \includegraphics[width=\textwidth]{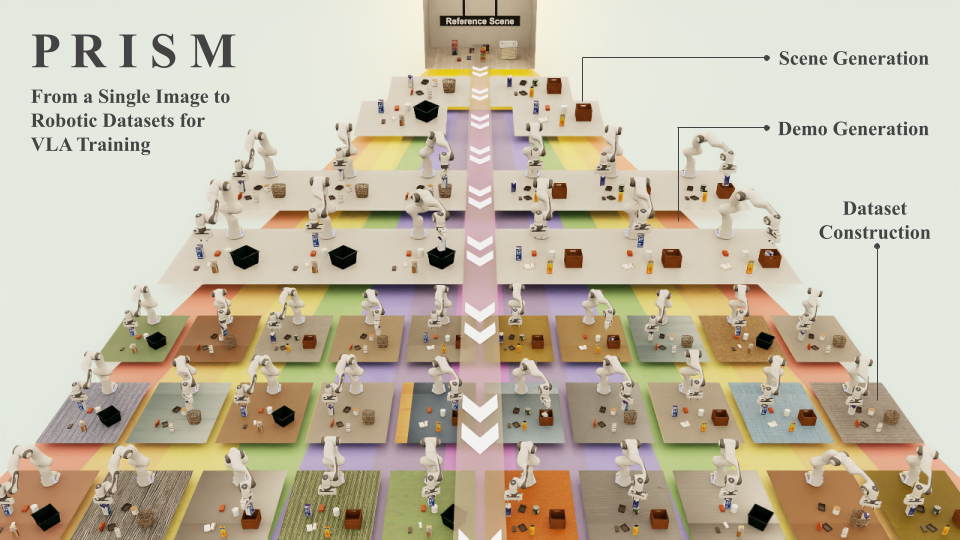}
    \caption{PRISM generates personalized robotic datasets from a single image and a task instruction of the target environment. Scenes and trajectories are synthesized while preserving semantic and geometric structure, and datasets are constructed through structured randomization.}
    \label{fig:introduction}
    \vspace{-2pt}
\end{figure}

\begin{abstract}
Recent advances in large-scale pretrained vision-language-action models have improved robot policy learning, but directly deploying such policies in user-specific environments remains challenging due to limited generalization, which inevitably requires collecting a dataset tailored to the target environment.
Teleoperation yields well-aligned data but is costly and difficult to scale, whereas simulation scales easily but struggles to resemble the target environment and generate task-specific trajectories. 
To meet both simultaneously, we propose PRISM, an end-to-end pipeline that generates personalized robotic datasets from a single image and a natural-language instruction. 
PRISM constructs digital cousin scenes that are semantically and geometrically aligned with the user environment yet diverse at the instance level, and synthesizes executable demonstrations without human teleoperation.
Extensive experiments show that policies trained on PRISM-generated datasets outperform those trained on baseline-generated datasets on LIBERO and LIBERO-Plus, achieve up to 100\% success rate on three real-world manipulation tasks, and maintain stronger performance when evaluated in environments that differ from those seen during training.

\end{abstract}


\section{Introduction}
\label{sec:intro}
\vspace{-3pt}
Vision-Language-Action (VLA) models~\cite{openvla, octo, pi05} have rapidly emerged as a leading paradigm for general-purpose robotic manipulation, adapting large vision-language foundation models to predict robot actions directly from images and natural-language instructions.
Trained jointly on web-scale vision-language data and robot demonstrations across many robots, tasks, and environments~\cite{openx, droid}, modern VLAs acquire broad zero-shot competence over unseen objects, scenes, and instructions.
Their success suggests a promising path toward general-purpose robotic manipulation.

Yet this broad competence does not translate to a user's deployment environment, where the performance of a pretrained VLA often degrades sharply on tasks it would otherwise handle reliably.
The cause is structural, since a VLA's competence is bounded above by the empirical distribution from which its training trajectories were drawn, and that distribution rarely covers any specific user's environment densely enough for reliable execution~\cite{rt2}.
Closing this gap therefore requires fine-tuning on data tailored to the target environment, and recent studies show that the resulting policy's quality is governed mostly by how closely the training data match the deployment target~\cite{rt2, compare_ft_ml, openvla-oft}.
The central question is therefore how to obtain such target-aligned data without prohibitive human effort.

Existing approaches to acquiring target-aligned data fall into three families.
Teleoperation in the target environment, by direct robot control~\cite{openx, droid} or with specialized hardware~\cite{umi}, yields data that faithfully reflect the target environment, achieving structural alignment with task supervision, since they are gathered by a human who performs the task in the deployment scene itself.
Simulation-based approaches sidestep the cost of physical data collection and split into two further families.
One line of work uses LLMs or VLMs to compose tasks and scenes from a large 3D asset library~\cite{robogen, gensim, gensim2, robotwin1.0, robotwin2.0}, automatically producing large quantities of diverse training data with minimal human effort.
Another rebuilds a single specific target environment from sensor data such as RGB-D video, most notably the digital twin paradigm~\cite{xsim}, producing a simulator that mirrors the actual target deployment scene and supports policy training within it.

Despite their progress, each family falls short in a different way.
Teleoperation aligns with the target but is hard to scale, being costly and labor-intensive.
Synthesizing scenes with LLMs or VLMs scales freely but draws scenes from a target-agnostic distribution, so alignment to a specific user environment is lost.
Reconstructing the target as a digital twin restores alignment but collapses the distribution to a single instance, eliminating the variation needed to generalize beyond that exact configuration.
Thus, existing simulation-based approaches provide either target alignment or instance-level diversity, but not both. 
A personalized dataset, however, must preserve the target structure while varying scene instances.
We therefore frame personalized dataset generation not as collecting more perfectly target-aligned data, but as constructing a target-conditioned distribution of scenes that share the target's structure, including object categories and spatial relations, while varying at the instance level and paired with task-completing demonstrations at sufficient scale.

In this work, we propose \textbf{PRISM}, an end-to-end pipeline that instantiates this principle from a single RGB-D image of the target environment and a task instruction.
From the image, PRISM extracts object categories and geometry, retrieves semantically and visually matched assets from a large 3D asset library, and composes digital cousin scenes that preserve the target's categories and spatial relations while sampling alternative instances per category.
It then parses the instruction and synthesizes executable demonstrations, eliminating human teleoperation.
PRISM incorporates two design choices that improve dataset quality and efficiency.
It selects grasps that induce low-complexity motions, producing demonstrations that are easy to imitate rather than merely kinematically valid.
Moreover, it applies visual randomization while keeping each trajectory fixed, making data generation more efficient and encouraging the policy to learn task motions invariant to visual appearance.

Our contributions are as follows: 
(1) We propose PRISM, an end-to-end pipeline that generates personalized robotic datasets from a single RGB-D image and a natural-language instruction;
(2) we introduce a motion-aware grasp selection strategy that yields natural, reliably executable demonstrations and improves task success; 
and (3) a trajectory-preserving visual randomization scheme that improves data efficiency and encourages appearance-invariant policies.

\vspace{-3pt}
\section{Related Work}
\label{sec:related_work}
\subsection{Vision-Language-Action Models for Robot Manipulation}
\label{sec:related_work1}
Imitation learning has become the dominant approach to robot manipulation, where behavior cloning maps observations directly to actions, and diffusion-based policies~\cite{dp, dp3} model multimodal action distributions for more complex tasks.
Building on this paradigm, Vision-Language-Action (VLA) models~\cite{openvla, octo, pi05, gr00t} adapt large vision-language foundation models to predict robot actions directly from images and natural-language instructions.
Trained jointly on web-scale vision-language data and large-scale robot demonstrations spanning many robots, tasks, and environments~\cite{openx, droid, rt1, roboturk}, VLAs acquire broad zero-shot competence and have become the default starting point for downstream manipulation policies.
Since collecting such demonstrations by teleoperation is costly, PRISM procedurally generates physically plausible demonstrations within a user-specific simulation environment, scalably expanding the data needed to adapt a VLA to its target environment.

\vspace{-1pt}
\subsection{Scene Generation for Robot Simulation}
\vspace{-1pt}
\label{sec:related_work2}
Simulation offers a scalable alternative to real-world teleoperation for collecting manipulation data, and a body of work focuses on automatically constructing the simulated scene.
URDFormer~\cite{urdformer} reconstructs articulated scene structures directly from images.
ACDC~\cite{acdc} first proposes the digital cousin concept and composes such scenes from a single real reference image, preserving the reference's high-level structure while substituting alternative asset instances, and GAIA~\cite{gaia} augments the scene with the objects needed to make the task instruction executable within it.
Another group of research generates scenes as one stage of a larger pipeline for producing robot training data or interactive environments.
RoboGen~\cite{robogen} and GenSim2~\cite{gensim2} prompt an LLM to propose a manipulation task and then compose the corresponding scene by selecting and arranging the objects the task requires.
Gen2Sim~\cite{gen2sim} lifts objects from a single image into simulation-ready 3D assets and assembles them into an interactive scene.
The resulting scenes, however, are drawn from a generic distribution and do not reflect a specific real-world target environment.

\subsection{Simulation-Based Demonstration Generation}
\label{sec:related_work3}
Recent works utilize foundation models to synthesize tasks and demonstrations, prompting LLMs to propose tasks and generate the supervision needed to solve them~\cite{robogen, gensim, gensim2, humanoidgen}, guiding long-horizon task-and-motion planning with commonsense priors~\cite{vlm-tamp}, or replacing physical simulation with VLM-driven video world models~\cite{dreamgen}.
Another line reconstructs the deployment scene as a digital twin, following the Real-to-Sim-to-Real paradigm to rebuild the target environment from sensor data for policy training~\cite{robotwin1.0, robotwin2.0, xsim}, or scaling a small set of seed demonstrations to many configurations within a given scene~\cite{mimicgen, dexmimicgen, skillmimicgen, demogen}.
In contrast, PRISM generates digital cousin scenes from a single image and synthesizes demonstrations for executing the task instruction.

\begin{figure*}[t]
    \centering 
    \includegraphics[width=1.0\linewidth]{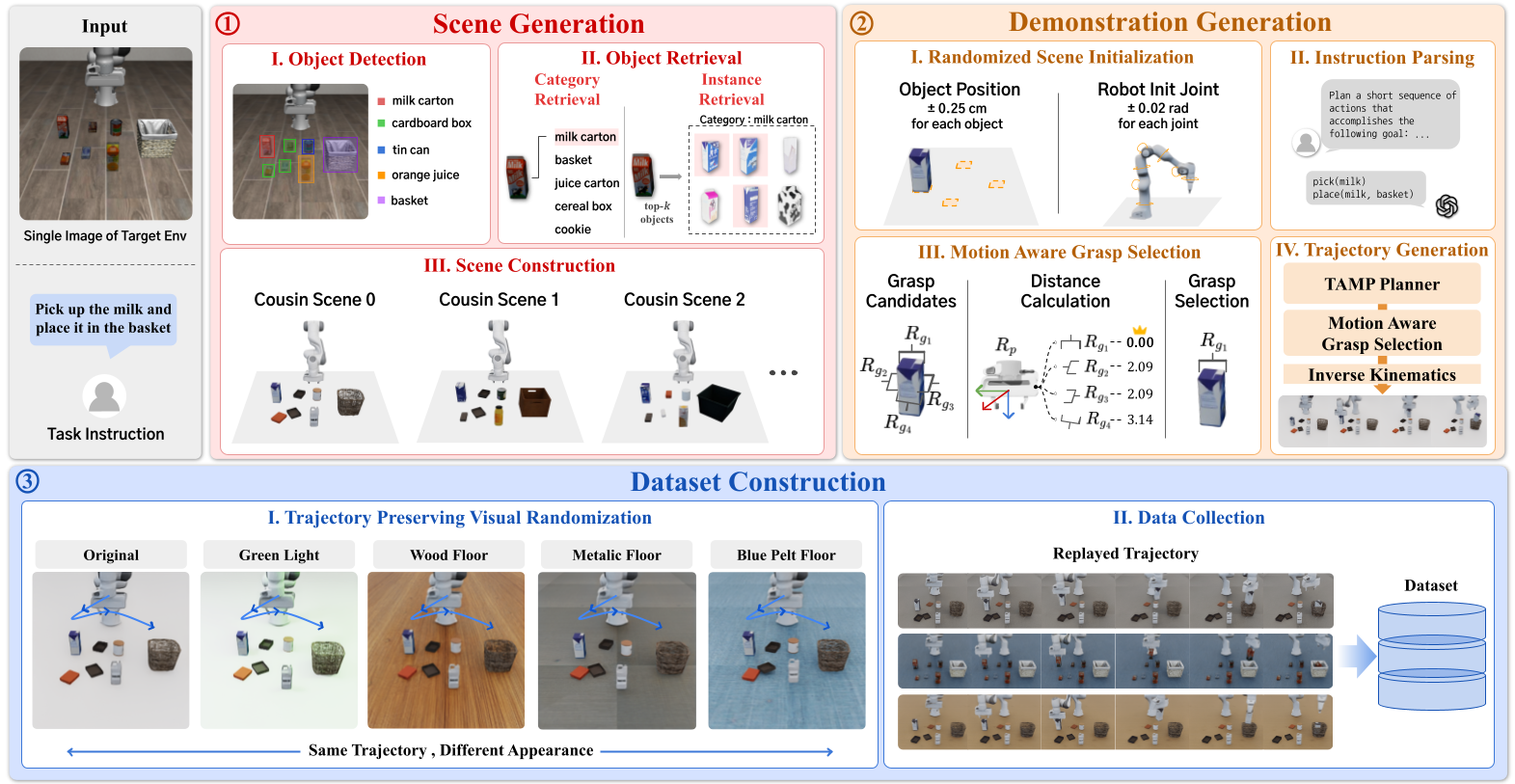}
    \caption{Overview of PRISM framework.}
    \vspace{-10pt}
    \label{fig:framework}
\end{figure*}


\section{Methodology}
\label{sec:method}
PRISM is an end-to-end pipeline that generates robotic datasets from a single image and a natural-language instruction.
The pipeline consists of three stages.
First, Section~\ref{subsec:method_section1} presents \textbf{Scene Generation}, which reconstructs a virtual scene and places task-relevant objects using VLM-based reasoning.
Second, Section~\ref{subsec:method_section2} describes \textbf{Demonstration Generation}, where the instruction is converted into a symbolic plan (PDDL) and synthesized into a collision-free trajectory via task-and-motion planning (TAMP).
Finally, Section~\ref{subsec:method_section3} introduces \textbf{Dataset Construction}, which applies domain randomization and success validation to build a diverse dataset for robust policy learning.

\subsection{Scene Generation}
\label{subsec:method_section1}
\vspace{-3pt}
Inspired by GAIA~\cite{gaia}, we construct a robot-interactable simulation scene that is structurally and semantically aligned with a user’s real-world environment.
Given an environment image and task instruction, PRISM extracts object-level information, retrieves aligned 3D assets, and builds a task-aware simulation scene for robot execution by integrating scene geometry, assets, and task context.

\textbf{Object Detection and Segmentation.}~\enskip
Given the input image $\mathbf{I}$, we extract instance masks, a depth map, and camera intrinsic parameters.
For instance mask extraction, we first infer a set of object names $\mathcal{O} = \{\textbf{o}_1, \dots, \textbf{o}_N\}$ present in the scene using a vision-language model (VLM).
Given these object names, we apply Grounded-SAM~\cite{grounded-sam} to obtain an instance-level segmentation mask for each object.
If RGB-D observations and camera intrinsics are available, we directly use the provided depth maps and intrinsics.
Otherwise, when only a single RGB image is given, we estimate the depth map and camera intrinsics using Depth Anything v2~\cite{depth_anything} and Perspective Fields~\cite{perspective-fields}, respectively.

\textbf{Object Retrieval.}~\enskip
We retrieve candidate assets $\mathcal{C}_i$ for each detected object $\textbf{o}_i$ from a large-scale 3D asset library~\cite{behavior} through a two-stage process.
First, at the category level, we identify the top-$n$ categories whose CLIP embeddings are closest to $\textbf{o}_i$.
Second, at the instance level within each category, we retrieve the top-$m$ nearest-neighbor assets by comparing the segmented image of $\textbf{o}_i$ with rendered images of 3D assets in the DINOv2~\cite{dinov2} embedding space.
We then prompt a VLM with the retrieved renderings to select the top-$k$ assets $(k \le m)$ most visually similar to $\textbf{o}_i$, yielding the set of candidate assets $\mathcal{C}_i = \{\textbf{a}_i^{(1)}, \dots, \textbf{a}_i^{(k)}\}$.
Applying this procedure to all detected objects produces $\{\mathcal{C}_i\}_{i=1}^{N}$, which are subsequently used in scene construction.

\textbf{Scene Construction.}~\enskip
Finally, we integrate the extracted scene information and the retrieved candidate sets $\{\mathcal{C}_i\}_{i=1}^{N}$ to construct a robot-interactable simulation environment. 
For each detected object $\textbf{o}_i$, we reconstruct its point cloud from the depth map and segmentation mask to estimate its location, and place a randomly sampled asset $\textbf{a}_i^{(j)}, j \in \{1, \dots, k\}$ at the estimated location.
Optionally, given the task instruction, a VLM can infer additional task-relevant objects that may be missing from the scene and place them at appropriate locations.
The environment can also be augmented with distractor objects, walls, and floor surfaces, increasing scene-level diversity while reflecting the structure and appearance of the user environment.

\vspace{-3pt}
\subsection{Demonstration Generation}
\label{subsec:method_section2}
\vspace{-3pt}
Inspired by VLM-TAMP~\cite{vlm-tamp}, this step converts a natural language task instruction into a sequence of actions using a VLM, and then uses task-and-motion planning to generate executable robot trajectories from the predicted action sequence.
First, geometric randomization is applied to diversify the initial scene configuration.
A VLM then decomposes the task instruction into an action sequence using the object-annotated scene image and detected object list.
Finally, task-and-motion planning generates physically feasible robot trajectories, including grasp poses and joint motion paths, based on the predicted action sequence.

\textbf{Randomized Scene Initialization.}~\enskip
\label{subsec:method_section2_stage1}
Prior to demonstration generation, we randomize the generated scene to prevent demonstrations from being specialized to a single configuration.
Specifically, object positions are slightly perturbed, the robot’s initial joint configuration is randomly shifted around a nominal pose, and optionally object scales are independently varied along each axis.
This enables PRISM to generate demonstrations under diverse object layouts, robot starting states, and shape variations even for the same task instruction.

\textbf{Instruction Parsing.}~\enskip
\label{subsec:method_section2_stage2}
To enable a robot to execute a task specified in natural language, the instruction must first be translated into a sequence of actions amenable to planning and execution.
We adopt a two-phase querying scheme to translate the natural-language instruction $\ell$ into a sequence of primitive actions.
In the first phase, we prompt a VLM with $\ell$, an environment image annotated with object bounding boxes, and the list of detected objects $\mathcal{O}$.
The VLM is then asked to produce a sequence of actions in natural language, $\hat{\pi}^{\mathrm{eng}}$.
In the second phase, the VLM translates $\hat{\pi}^{\mathrm{eng}}$ into a partially grounded action sequence $\hat{\pi} = \{\hat{a}_1, \dots, \hat{a}_K\}$, where each $\hat{a}_k$ is drawn from a predefined set of primitive actions $\mathcal{A}_{\mathrm{prim}}$ (e.g., $\texttt{pick}(\cdot)$, $\texttt{place}(\cdot, \cdot)$).

\textbf{Trajectory Generation.}~\enskip
\label{subsec:method_section2_stage3}
Given the plan skeleton $\hat{\pi}$ produced by the VLM, we employ a TAMP framework~\cite{piginet} to synthesize an executable robot trajectory $\tau$.
A TAMP problem is formulated as $\langle \mathcal{O}, \mathcal{I}, \hat{\pi} \rangle$, where $\mathcal{O}$ denotes the objects in the scene, $\mathcal{I}$ is the current state, and $\hat{\pi}$ is the plan skeleton to be refined.
For each partially grounded action $\hat{a}_k$, the planner searches for continuous parameters such as grasp poses, robot configurations, and collision-free motion paths, while jointly considering the robot's current state and the spatial arrangement of objects.

\textbf{Motion-Aware Grasp Selection.}~\enskip
\label{subsec:method_section2_stage4}
For manipulation actions, the TAMP planner must select a grasp pose before solving for the arm motion trajectory.
Instead of using a randomly sampled grasp, PRISM ranks candidate grasps according to their alignment with a set of canonical end-effector orientations $\mathcal{R}_c \subset \mathrm{SO}(3)$, defined relative to the robot base to correspond to natural front-facing approach directions.
For each candidate grasp with orientation $\mathbf{R}_g$ produced for the target object, we transform both $\mathbf{R}_g$ and the canonical orientations into the world frame and measure the deviation of $\mathbf{R}_g$ from a canonical orientation $\mathbf{R}_c \in \mathcal{R}_c$, which is the minimal rotation angle needed to align the two orientations.
Candidates with the smallest alignment error are prioritized and forwarded to the TAMP planner for inverse kinematics and collision-free motion planning.
This motion-aware grasp selection guides the planner toward smoother and more consistent arm motions, improving demonstration quality for downstream policy learning.

\vspace{-6pt}
\subsection{Dataset Construction}
\label{subsec:method_section3}
\vspace{-3pt}
Based on the executable trajectories generated in Section~\ref{subsec:method_section2}, we construct a dataset for policy learning by replaying these trajectories in simulation.
This stage aims to transform a limited set of executable trajectories into a visually diverse dataset that supports robust policy learning.
To this end, we augment the replay process with visual randomization, allowing each trajectory to generate multiple visually distinct observations while preserving identical underlying robot motions.

\textbf{Trajectory-Preserving Visual Randomization.}~\enskip
\label{subsec:method_sec3_stage1_2}
Conventional randomization approaches apply a different random configuration to each trajectory, which may cause the policy to memorize specific trajectory–visual condition pairings rather than learning to generalize across visual variations.
In contrast, our strategy keeps the action trajectory and all physical states fixed, and replays each demonstration multiple times while varying only visual factors such as lighting and background textures.
By exposing the policy to visually diverse observations that share identical task-relevant content, this scheme encourages it to focus on the state and action information essential to the task while remaining invariant to task-irrelevant visual factors.

\textbf{Data Collection.}~\enskip
\label{subsec:method_section3_stage2}
Using the visually randomized environments, we replay the executable trajectories generated during the demonstration generation stage to collect the dataset.
During each replay, we record the robot’s observations, executed actions, and task success signals at each timestep.
Leveraging the flexibility of simulation, we can place arbitrary sensors at desired locations, including rich sensory information in the dataset.
After the data collection process, we filter the dataset to retain only successful demonstrations, which are then used for policy learning.

\begin{figure}[t]
    \centering 
    \includegraphics[width=1.0\linewidth]{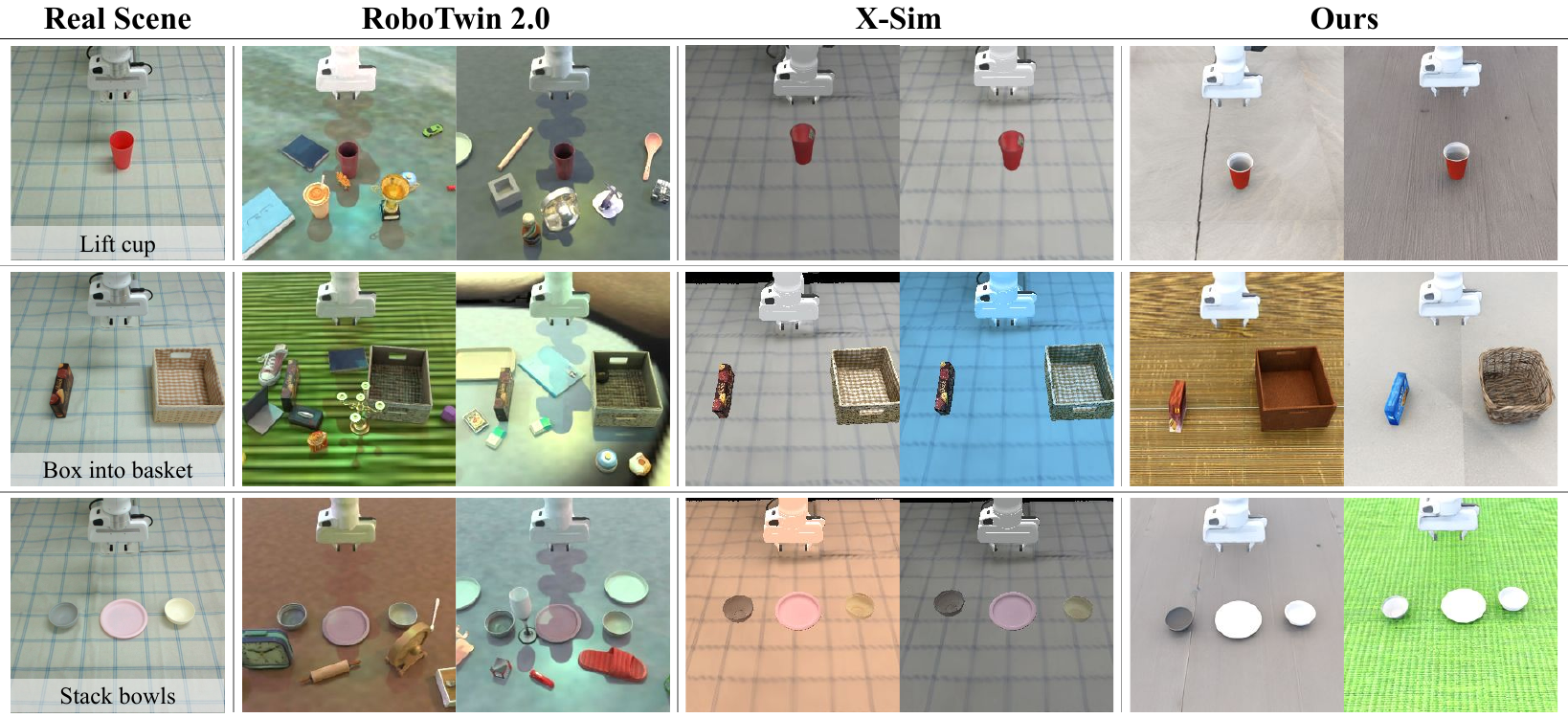}
    \caption{Qualitative results of the datasets generated by each method on real-to-sim-to-real tasks.}
    \label{fig:exp_r2s2r_qualitative_results}
    \vspace{-5pt}
\end{figure}

\begin{table*}[]
\resizebox{\textwidth}{!}{%
\begin{tabular}{lc|ccc|ccc}
\toprule 
                              &              & \multicolumn{3}{c|}{Put milk in basket} & \multicolumn{3}{c}{Put wine bottle on cabinet} \\
                              &              & In-Domain     & LIBERO        & LIBERO-Plus   & In-Domain      & LIBERO        & LIBERO-Plus    \\ \midrule
\multirow{3}{*}{$\pi_{0.5}$}  & RoboTwin 2.0 & 74.0          & 14.0          & 21.9          & 76.0           & 16.0          & 3.3            \\
                              & X-Sim        & \textbf{96.0} & 48.0          & 35.8          & 94.0           & 82.0          & \textbf{54.5}  \\
                              & Ours         & 72.0          & \textbf{98.0} & \textbf{67.6} & \textbf{98.0}  & \textbf{98.0} & 52.0           \\ \midrule
\multirow{3}{*}{DP}           & RoboTwin 2.0 & 84.0          & 2.0           & 33.7          & 78.0           & 34.0          & 27.2           \\
                              & X-Sim        & 84.0          & 80.0          & 2.8           & 40.0           & 44.0          & 0.6            \\
                              & Ours         & \textbf{95.0} & \textbf{94.0} & \textbf{35.6} & \textbf{100.0} & \textbf{56.0} & \textbf{28.8}  \\ \bottomrule
\end{tabular}
}
    \caption{\centering{Task success rate on the sim-to-sim experiment.}}
    \vspace{-5pt}
    \label{tab:sim2sim_result}
\end{table*}

\section{Experiments}
\label{sec:experiments}
We conducted experiments using PRISM to address the following research questions. 
\textbf{Q1)} Does the generated dataset effectively contribute to improving the robot's task success rate? 
\textbf{Q2)} Does the dataset enhance the generalization performance of the robotic policy model by incorporating environmental variations?
\textbf{Q3)} Is the generated dataset applicable to real-world environments?
\textbf{Q4)} How efficient is the PRISM pipeline in terms of data generation?

\subsection{Sim-to-Sim Experiment}
\label{subsec:experiment1}
To address Q1 and Q2, we trained policies on datasets generated by RoboTwin 2.0~\cite{robotwin2.0}, X-Sim~\cite{xsim}, and PRISM to measure task success rates across in-domain and out-of-domain simulation settings.
For each method, we generated 400 trajectories per task and used them to fine-tune $\pi_{0.5}$~\cite{pi05} and train Diffusion Policy~\cite{diffusionpolicy} from scratch.
The \textbf{In-Domain} setting evaluates each method in its native simulator, and the out-of-domain setting evaluates the same policies on two external benchmarks rendered in a different simulator, isolating cross-simulator transfer.
\textbf{LIBERO}~\cite{libero} serves as the base out-of-domain benchmark, and \textbf{LIBERO-Plus}~\cite{libero-plus} further evaluates policies over a substantially broader set of environments with variations in lighting, background, noise, and layout to rigorously stress generalization.
Additional details are provided in the supplementary material.

As shown in Table~\ref{tab:sim2sim_result}, policies trained on PRISM-generated datasets achieved strong performance across out-of-domain settings and outperformed baselines in most cases.
Moreover, in most settings PRISM exhibits the smallest performance drop from In-Domain to LIBERO.
We attribute this to PRISM's use of structurally and semantically analogous environments instead of exact digital twins.
This design choice increases diversity during data generation.
This trend holds for both $\pi_{0.5}$ fine-tuning and Diffusion Policy training.

\vspace{-3pt}
\subsection{Real-to-Sim-to-Real Experiment}
\label{subsec:experiment3}
\vspace{-3pt}

\begin{wrapfigure}{r}{0.60\linewidth}
    \vspace{-10pt}
    \centering
    \includegraphics[width=\linewidth]{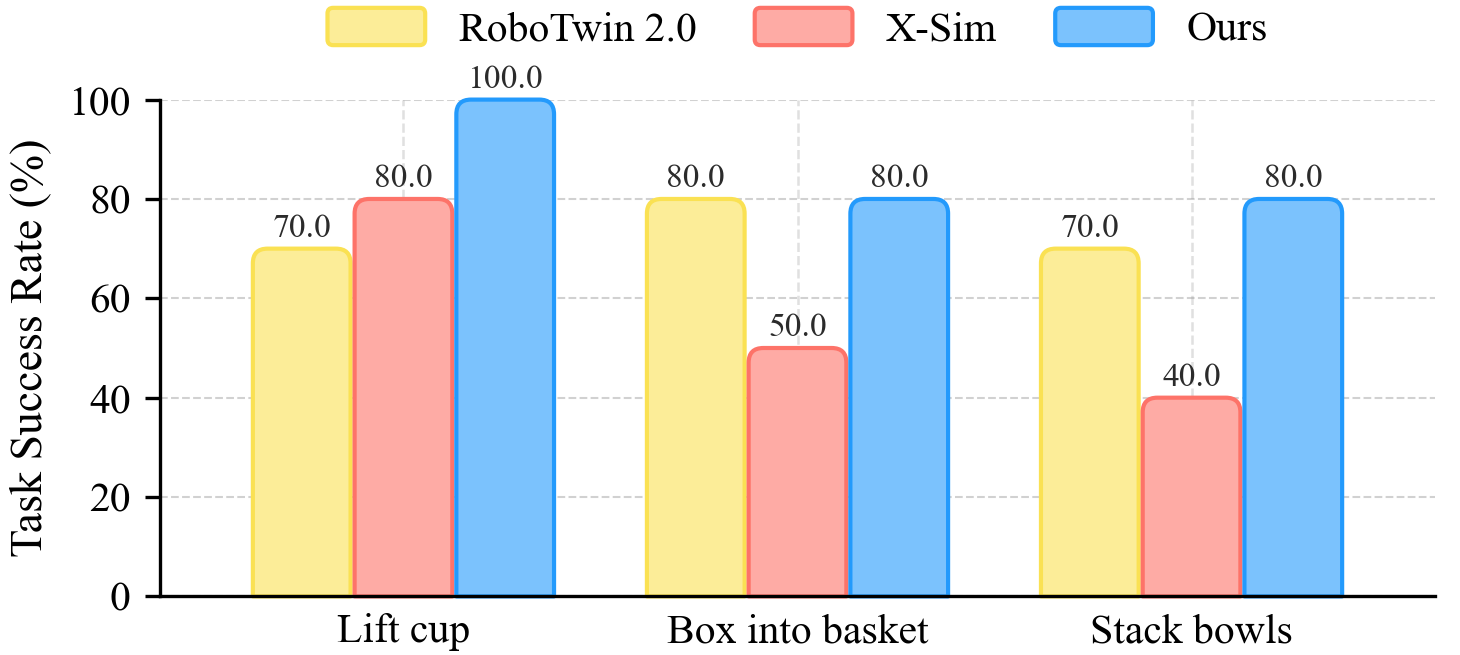}
    \caption{Task success rate of the real-to-sim-to-real experiments.}
    \label{fig:exp_r2s2r}
    \vspace{-10pt}
\end{wrapfigure}

\textbf{Quantitative Results.}~\enskip
To address Q3, we considered three real-world manipulation tasks and generated a dedicated dataset for each task.
Using these datasets, we fine-tuned $\pi_{0.5}$ and evaluated the resulting policies in real-world settings.
For each task, the trained policy was evaluated over 10 independent trials, and success rates were computed accordingly.
As shown in Figure~\ref{fig:exp_r2s2r}, policies trained on PRISM-generated datasets consistently outperformed those trained with baseline methods.
These results demonstrate that PRISM generates effective training data while preserving the structure of the target environments.

\textbf{Qualitative Results.}~\enskip
Figure~\ref{fig:exp_r2s2r_qualitative_results} visualizes representative samples from the datasets generated by each method in our real-to-sim-to-real experiments.
For the baselines, we generated a 3D object from each image, scaled it to match the real object's dimensions, and positioned it in the simulation according to measurements of each object's real-world location.
In contrast, PRISM requires no human intervention in determining object categories, sizes, or spatial relationships.
Despite this, as shown in the figure, the PRISM-generated datasets remain semantically aligned with the real scenes.

\vspace{-3pt}
\subsection{Effectiveness of Digital Cousin}
\label{subsec:experiment5}
\vspace{-3pt}

\begin{wraptable}{r}{0.48\textwidth} 
    \centering    
    \renewcommand{\arraystretch}{1.2} 
    \vspace{-5pt}
    
    \begin{tabular}{ccc}
    \toprule
                  & \makecell{Target \\ Environment} & \makecell{Variant \\ Environment} \\ \hline
    PRISM-Twin    & \textbf{100.0} & 30.0          \\ \hline
    PRISM-Cousin  & 80.0           & \textbf{80.0} \\ \bottomrule 
    \end{tabular}
    
    \caption{Success rate of policies trained on datasets from PRISM-Twin and PRISM-Cousin}
    \label{tab:ablation_dt_vs_dc}
    \vspace{-10pt}
\end{wraptable}
To assess the effect of instance-level diversity, we fine-tuned $\pi_{0.5}$ on two datasets generated by the PRISM pipeline. 
The first comes from a modified version that uses a digital twin scene of the target environment (PRISM-Twin), and the second comes from the unmodified version, which uses a digital cousin scene (PRISM-Cousin).
For the \textit{Box into basket} task, success rates were measured over 10 trials in the target environment and over 5 trials in each variant environment with different objects and background appearances.
As shown in Table~\ref{tab:ablation_dt_vs_dc}, PRISM-Twin achieves a higher success rate in the target environment but degrades sharply in the variant environments, where the objects and background are changed.
In contrast, PRISM-Cousin maintains a consistent success rate across both the target and variant environments.
This indicates that digital cousin scenes preserve enough fidelity while introducing enough diversity to avoid overfitting to a single scene instance.

\begin{wraptable}{r}{0.45\textwidth} 
    \centering
    \vspace{-5pt} 
    
    \renewcommand{\arraystretch}{1.2} 
    
    \begin{tabular}{ccc}
    \toprule
                & \makecell{Random \\ Grasp Sampling} & \makecell{Motion-Aware \\ Grasp Selection} \\ \hline
    $\pi_{0.5}$ & 56.0           & \textbf{98.0}   \\ \hline
    DP          & 52.0           & \textbf{56.0}   \\ \bottomrule 
    \end{tabular}
    \caption{Task success rate when motion-aware grasp selection is applied or not.}
    \label{tab:ablation_adaptive_grasp}
    \vspace{-10pt} 
\end{wraptable}

\vspace{-3pt}
\subsection{Ablation Study of Motion-Aware Grasp Selection}
\label{subsec:ablation_adaptive_grasp}
\vspace{-3pt}
To verify whether motion-aware grasp selection actually contributes to performance improvement, we fine-tuned and trained policy models on two datasets, one using random grasp sampling and the other using motion-aware grasp selection.
For the dataset without motion-aware grasp selection, we generated each demonstration trajectory by randomly sampling a grasp pose from the available grasp candidates.
As shown in Table~\ref{tab:ablation_adaptive_grasp}, applying our motion-aware grasp selection improves task success rate over the random sampling baseline for both $\pi_{0.5}$ and Diffusion Policy.
By selecting grasps aligned with canonical approach directions, it yields arm motions that are easier to imitate, and the gain across both policies shows that the benefit is not tied to a particular learning paradigm.

\begin{figure}[t]
    \centering
    
    \begin{minipage}[t]{0.49\linewidth}
        \centering
        \includegraphics[width=\linewidth]{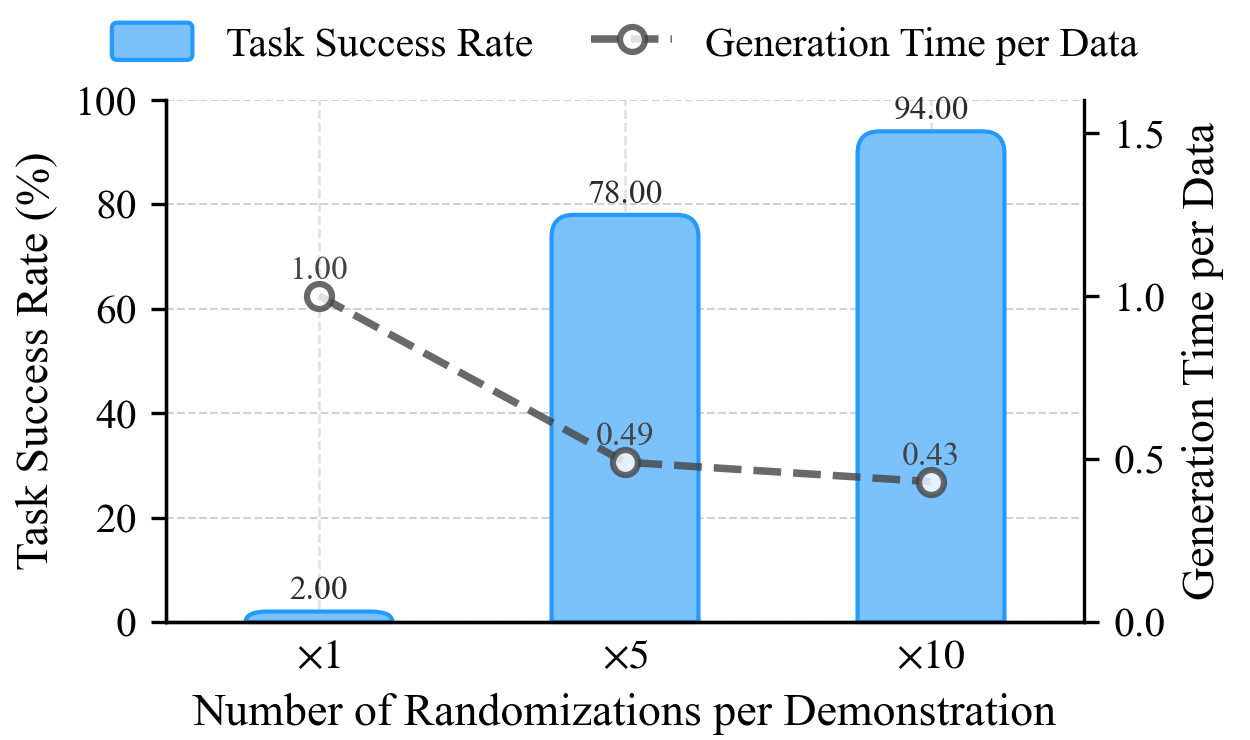}
        \caption{Ablation on the number of randomizations per demonstration.}
        \label{fig:ablation_tpvr}
    \end{minipage}
    \hfill
    \begin{minipage}[t]{0.49\linewidth}
        \centering
        \includegraphics[width=\linewidth]{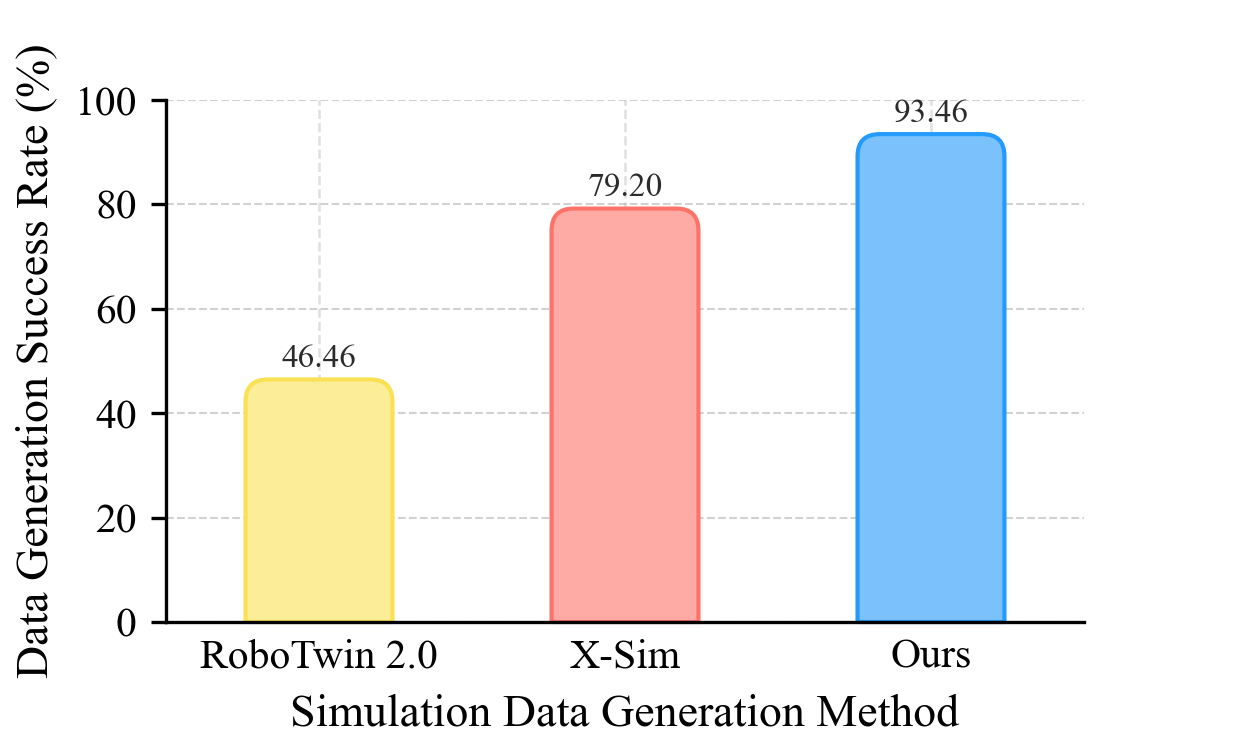}
        \caption{Task success rate of generated demonstrations.}
        \label{fig:data_gen_sr}
    \end{minipage}
    
    \vspace{-15pt}
\end{figure}

\vspace{-3pt}
\subsection{Effect of Trajectory-Preserving Visual Randomization}
\label{subsec:ablation_tpvr}
\vspace{-3pt}
We conducted an experiment to verify the effectiveness of applying visual randomization while preserving the trajectory during data collection. 
Using 40 demonstration trajectories, we evaluated this approach by varying the number of randomizations applied per trajectory across 1, 5, and 10. 
As shown in Figure~\ref{fig:ablation_tpvr}, applying multiple randomizations to a single trajectory leads to better performance. 
In addition, the figure plots the generation time per data sample, normalized to the case of one randomization per trajectory. 
Because PRISM is structurally designed to first generate a demonstration trajectory and then produce the dataset from it, using more than one randomization per trajectory reuses the same trajectory and saves the time needed to generate new demonstration trajectories, producing a high-quality dataset at a lower generation cost per sample.

\vspace{-3pt}
\subsection{Pipeline Efficiency}
\label{subsec:experiment4}
\vspace{-3pt}
To answer Q4, we evaluated the efficiency of each method in generating successful task demonstrations.
For each method, 400 successful dataset instances were collected, and the success rate was computed as the number of successful generations divided by the total number of generation attempts.
Unlike RoboTwin 2.0 and X-Sim, which require repeated trajectory generation or RL rollouts, PRISM generates a small set of successful demonstrations once and reuses them under visual randomization to construct the dataset.
We therefore measured the success rate at the demonstration generation stage, where successful demonstrations are reliably converted into dataset instances.
Further details are provided in the supplementary material.

\vspace{-3pt}
\section{Conclusion}
\label{sec:conclusion}
\vspace{-3pt}
We present PRISM, an end-to-end pipeline that generates personalized robotic datasets from a single image of a target environment and a natural-language instruction. 
By constructing digital cousin scenes, PRISM preserves the semantic and geometric structure of the user environment while introducing controlled instance-level diversity, and it synthesizes executable demonstrations without human teleoperation. 
Further, motion-aware grasp selection makes the generated demonstrations easier to imitate and trajectory-preserving visual randomization enables the pipeline to generate high-quality datasets efficiently. 
Extensive experiments show that policies trained on PRISM-generated datasets achieve a higher task success rate and stronger robustness to unseen variations.

\vspace{-3pt}
\section{Limitations}
\label{sec:limitations}
\vspace{-3pt}
While PRISM demonstrates effective personalized dataset generation, it has several limitations that point to directions for future work.
It is currently limited to the Franka robot arm and to rigid objects, leaving extensions to broader embodiments and to deformable objects such as cloth, ropes, or food items for future work.
Moreover, since PRISM reconstructs the scene from a single image, heavy occlusion can lead to inaccurate reconstruction and degraded demonstration generation.
We refer the reader to the supplementary material for more detailed analysis.


\clearpage


\bibliography{references}  


\clearpage

\appendix
\input{appendix}

\end{document}

%% file: appendix.tex
\section{Overview}
The Appendix contains the following :
\begin{itemize}
    \item \textbf{PRISM Implementation Details} (Appendix~\ref{sec:supp_prism_pipeline}): Provides detailed descriptions of the PRISM pipeline, including scene generation, demonstration trajectory generation, motion-aware grasp selection, dataset construction, randomization strategies, and the statistics of the generated datasets.

    \item \textbf{Baseline Pipelines} (Appendix~\ref{sec:supp_baseline_pipeline}): Summarizes the implementation details of the baseline methods used for comparison, including X-Sim and RoboTwin 2.0, along with a comparison across pipelines.

    \item \textbf{Policy Training Hyperparameters} (Appendix~\ref{sec:supp_implementation}): Presents policy training configurations and implementation details for Diffusion Policy and $\pi_{0.5}$.

    \item \textbf{Experimental Setup} (Appendix~\ref{sec:supp_task_description}): Describes the simulation and real-world robotic manipulation tasks used in our experiments, including task definitions and evaluation settings.

    \item \textbf{Extended Experimental Results \& Analysis} (Appendix~\ref{sec:supp_extended_results}): Provides additional quantitative and qualitative analyses, including data generation efficiency, LIBERO generalization results, real-world generalization, and a failure case taxonomy.

    \item \textbf{Example Prompts} (Appendix~\ref{sec:supp_generation_prompts}): Lists the prompts used in the PRISM pipeline, including the scene generation and demonstration generation prompts.
\end{itemize}


\section{PRISM Implementation Details}
\label{sec:supp_prism_pipeline}

\subsection{PRISM Pipeline Details}
\label{sec:supp_pipeline_details}

\paragraph{Overview.}
PRISM synthesizes imitation learning trajectories entirely in simulation by combining digital cousin scene generation with a vision-language task-and-motion planner.
Given a single RGB-D observation of the target workspace and a natural-language task description, the pipeline proceeds in three stages.
It first reconstructs a population of digital cousin scenes that preserve the topological structure of the observed workspace while varying in asset identity. 
It then synthesizes feasible manipulation trajectories on each scene through a VLM-guided TAMP solver augmented with a motion-aware grasp selector. 
Finally, it applies trajectory-preserving visual randomization to expand each successful trajectory into a diverse set of training samples. 
This section describes the implementation details of each stage that could not be included in the main text.

\paragraph{Input Acquisition.}
The input to PRISM is a single RGB-D image of the target workspace together with a task instruction.
The intrinsic parameters of the camera that captured the image are also provided. In our sim-to-sim experiments the RGB-D image was rendered directly from MuJoCo using the simulator's reference camera, and the intrinsics provided by MuJoCo~\cite{mujoco} were passed to PRISM without modification.
In our real-world experiments the image was acquired with an Intel RealSense L515 camera, and the device's factory-calibrated intrinsics were used as-is.
The same single image and the same intrinsics are used throughout the pipeline, from digital cousin reconstruction to dataset generation.

\paragraph{Digital Cousin Scene Generation.}
The RGB-D observation is passed to GAIA~\cite{gaia} to generate a population of digital cousin scenes that preserve the spatial layout of the target workspace while varying in object identity.
The prompts used to drive GAIA are reproduced verbatim in Section~\ref{sec:supp_scene_generation}. 
For every object in the workspace, GAIA retrieves 8 candidate assets from its asset database, ranked by visual and category similarity to the observation.
We then construct 4 final scenes by independently choosing one asset per object from the 8 candidates.
The same 4-scene budget is applied uniformly across every task in our experiments.

\paragraph{Demonstration Trajectory Generation.}
Demonstration trajectories are synthesized on each generated digital cousin scene using a VLM-TAMP~\cite{vlm-tamp} planner, which allows PRISM to produce executable, task-completing demonstrations without human teleoperation.
The task-planning prompts we provide to the VLM, including the system message and the structured task-description fields, are listed in full in Section~\ref{sec:supp_demo_generation}. 
Before any per-episode randomization is applied, the robot base and the observation camera are placed manually at fixed reference poses.
We placed them under the same protocol for every method, including all baseline pipelines, so that the spatial relationship between the workspace and the agent was comparable across methods.
The first input image to the VLM is rendered from this reference camera.
After the reference setup is fixed, the randomization parameters described in Section~\ref{sec:supp_randomization} are sampled and applied to the scene to instantiate the episode used for trajectory synthesis.

\begin{wrapfigure}{r}{0.45\textwidth}
    \centering
    \includegraphics[width=\linewidth]{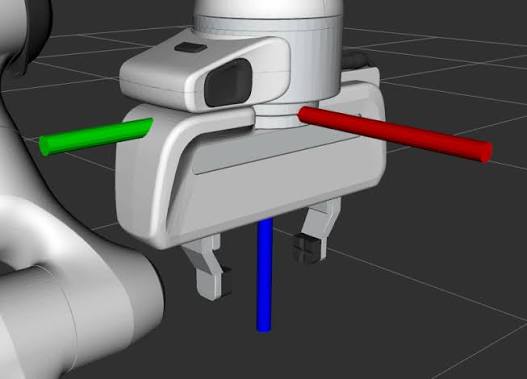}
    \caption{Canonical hand pose.}
    \label{fig:ablation_canonical_hand_pose}
    \vspace{-10pt}
\end{wrapfigure}
\paragraph{Motion-Aware Grasp Selection.}
For every object that the planner intends to grasp, PRISM enumerates candidate grasp poses by aligning the gripper to each face of the object's axis-aligned bounding box, which yields 6 candidate grasps per object by default.
For bowl objects, PRISM additionally examines 4 grasps aligned with the cardinal directions from above the bowl, so that the gripper can approach the rim from the top.

Rather than sampling a grasp at random from this candidate set, PRISM scores each candidate by its rotational distance from a canonical hand pose and selects the closest one.
The canonical pose is the hand pose of the robot in its initialized configuration, in which the gripper points downward toward the workspace and the hand's $x$-axis is aligned with the robot base's forward viewing direction. Given the canonical hand rotation $R_c \in \mathrm{SO}(3)$ and a candidate grasp rotation $R_g \in \mathrm{SO}(3)$, the rotational distance is the geodesic distance on $\mathrm{SO}(3)$:

\begin{equation*}
    d(R_g, R_c) = \arccos\!\left( \frac{\mathrm{tr}\!\left(R_c^{-1} R_g\right) - 1}{2} \right).
    \label{eq:grasp-geodesic}
\end{equation*}

The candidate with the smallest distance is selected, which corresponds to the grasp that requires the least wrist re-orientation from the canonical configuration and steers the TAMP planner toward smoother, more consistent arm motions.
A visualization of the canonical pose used in our experiments is provided in Figure~\ref{fig:ablation_canonical_hand_pose}.

\paragraph{Dataset Construction.}
For each of the 4 digital cousin scenes, PRISM synthesizes 20 candidate trajectories and retains the 10 shortest, since shorter trajectories tend to be more stable for downstream imitation learning and less prone to compounding errors during replay.
Each retained trajectory is then expanded by a factor of 10 through trajectory-preserving visual randomization, in which the kinematic trajectory of the robot and the object is held fixed while the workspace floor's background texture is varied using a texture pool drawn from RoboTwin 2.0's texture library.
Because each trajectory is reused rather than regenerated, this expansion adds visual diversity at a lower cost than synthesizing new demonstrations.
The expansion yields 100 trajectories per scene and 400 trajectories per task in total.

To keep the visual distribution comparable across tasks and across baseline pipelines, we sampled 400 background textures uniformly at random at the start of the experiment and reused the same 400 textures for every task.
Success or failure of each generated trajectory is determined by the task-specific success conditions defined in Section~\ref{sec:supp_task_description}, and only successful trajectories are saved into the final training dataset.

\subsection{Randomization Parameters}
\label{sec:supp_randomization}
\begin{table*}[t]
\centering
\resizebox{\linewidth}{!}{%
\begin{tabular}{ll}
\toprule
\textbf{Randomization Type} & \textbf{Description} \\
\midrule
Lighting Randomization &
Random variation of the scene lighting color \\

Texture Randomization &
Random variation of background textures and object surface appearances \\

Camera Pose Randomization &
Random perturbation of camera positions and orientations \\

Object Pose Randomization &
Random variation of object positions and orientations \\

Robot Initial State Randomization &
Random perturbation of initial robot joint configurations and end-effector poses \\
\bottomrule
\end{tabular}
}

\caption{Randomization parameters supported by PRISM.}
\label{tab:randomization_parameters}
\end{table*}

\begin{figure}[t]
    \centering
    \includegraphics[width=1.0\linewidth]{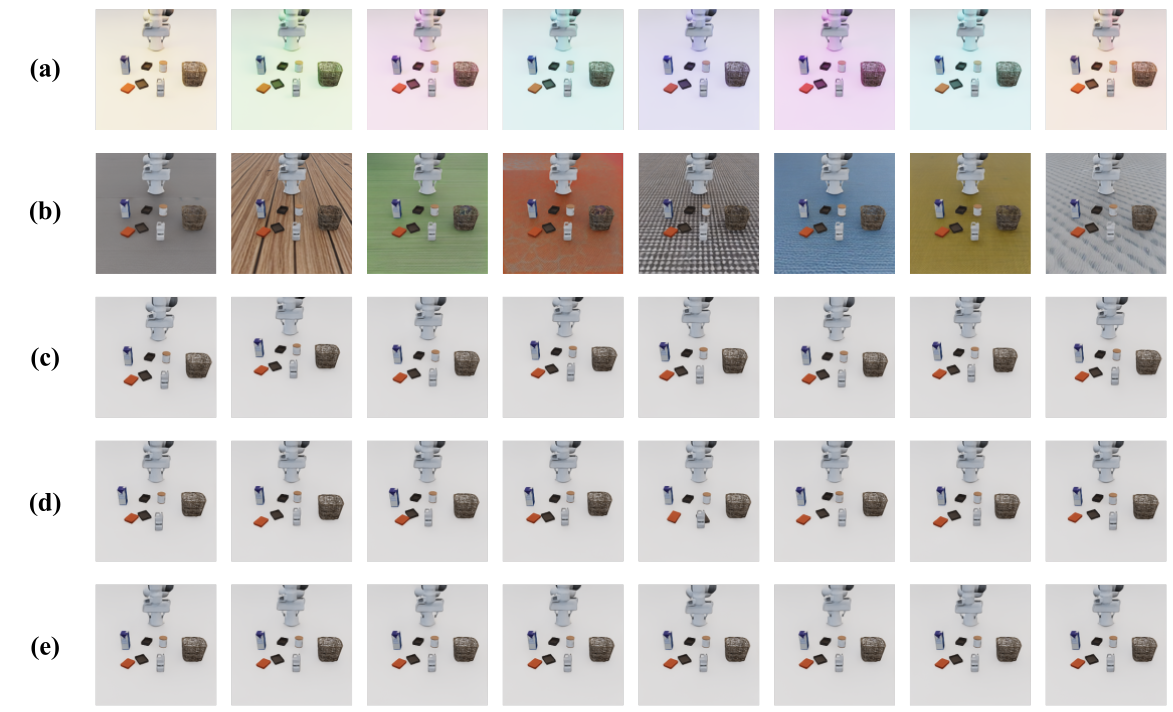}
    \caption{Examples of each randomization parameter.
(a) Lighting Randomization: varying the scene lighting color.
(b) Texture Randomization: substituting background and surface textures.
(c) Camera Pose Randomization: perturbing the camera position and orientation.
(d) Object Pose Randomization: randomizing the positions and orientations of task-relevant objects.
(e) Robot Initial State Randomization: randomizing the robot's initial joint configuration.}
    \label{fig:randomization_examples}
\end{figure}

To improve the robustness and generalization capability of manipulation policies, PRISM supports diverse randomization strategies during dataset generation.
The proposed randomization pipeline enables systematic variations in lighting conditions, background and object textures, object poses and orientations, camera poses, and robot initialization states. This section provides detailed descriptions of each randomization parameter in the proposed pipeline, along with visual examples of environments generated under each setting.

\paragraph{Lighting Randomization.}
PRISM randomizes the scene lighting to reduce overfitting to specific rendering conditions.
When this randomization is applied, the scene lighting color is randomly varied across episodes, which changes the overall color cast of the rendered scene and exposes the policy to diverse illumination appearances.
Other lighting attributes, such as light intensity, direction, shadow strength, and ambient illumination, can also be configured, but they are held fixed across episodes in our pipeline.

\paragraph{Texture Randomization.}
PRISM randomizes the textures of both background regions and scene objects to simulate diverse visual conditions observed in real-world environments.
Floor textures, wall appearances, and object surface textures are randomly modified to expose the policy to a wide range of visual conditions during training.
This form of randomization encourages the policy to focus on task-relevant visual features rather than overfitting to specific surface appearances or rendering artifacts present in a fixed simulation environment.

\paragraph{Camera Pose Randomization.}
Camera-related randomization is introduced by perturbing the camera pose around its reference configuration.
Both the camera position and orientation are randomly varied.
The position is shifted around its reference location, and the orientation is perturbed in yaw, pitch, and roll, with the viewing direction adjusted accordingly.
Such perturbations help improve robustness to viewpoint changes that may arise between simulation and real-world deployment.

\paragraph{Object Pose Randomization.}
To increase scene diversity, PRISM perturbs object poses at the beginning of each episode.
Object positions and orientations are randomly perturbed around predefined reference coordinate frames while maintaining physically valid placements and task-relevant semantic configurations.
Specifically, object positions are perturbed along the $x$ and $y$ axes, and object yaw orientations are randomly varied within predefined ranges.
This enables the generation of diverse scene layouts while preserving manipulation feasibility across episodes, and helps the policy generalize to unseen object configurations at test time.

\paragraph{Robot Initial State Randomization.}
Robot initialization noise can be introduced by perturbing the initial joint configuration.
The initial joint angles of the robot arm are randomized within a predefined range around reference values, and the end-effector pose is perturbed accordingly at the beginning of each episode.
This helps prevent the policy from overfitting to a single deterministic robot initialization state, encouraging the policy to generalize across diverse starting configurations.

Table~\ref{tab:randomization_parameters} summarizes the randomization parameters supported by PRISM, and Figure~\ref{fig:randomization_examples} illustrates visual examples of environments generated under each randomization setting.

\subsection{Generated Dataset Details}
\label{sec:supp_dataset_generation}

This section describes the structure of the demonstration datasets produced by PRISM, the information stored in each dataset, and the corresponding dataset statistics.

\begin{figure}[h]
\begin{center}
\begin{minipage}{0.60\linewidth}
\scriptsize
\begin{verbatim}
data/                                     (group, 400 episodes)
|-- demo_0 ... demo_399                    one episode each
|   |-- [num_samples]            int64    episode length T
|   |-- [sim_states]             JSON     full simulator state / replay
|   |-- actions          (T,7)   float64  EEF delta pose (6) + gripper (1)
|   |-- rewards          (T,)    int64    sparse, 1 on task success
|   |-- dones            (T,)    int64    episode-termination flag
|   |-- obs/
|       |-- third-person img (T,256,256,3) uint8    third-person RGB
|       |-- wrist img        (T,256,256,3) uint8    wrist-mounted RGB
|       |-- eef position     (T,3)  float32  end-effector position
|       |-- eef orientation  (T,4)  float32  end-effector orientation
|       |-- gripper position (T,2)  float32  gripper joint position
|       |-- gripper velocity (T,2)  float32  gripper joint velocity
|       |-- joint position   (T,7)  float32  arm joint angles
|       |-- joint cos        (T,7)  float32  cos of joint angles
|       |-- joint sin        (T,7)  float32  sin of joint angles
|       |-- joint velocity   (T,7)  float32  arm joint velocities
\end{verbatim}
\end{minipage}
\end{center}
\caption{Internal layout of a PRISM-generated dataset. Each dataset stores a set of successful demonstration episodes, and every episode contains per-step actions, rewards, termination flags, and a multimodal observation group with visual and proprioceptive signals. $T$ denotes the episode length.}
\label{fig:dataset_layout}
\end{figure}

\paragraph{Dataset Composition.}
Each generated dataset corresponds to a single manipulation task and consists of a collection of successful demonstration episodes.
Every episode pairs synchronized multimodal observations with the action executed at each time step.
The observations comprise RGB images and robot proprioceptive states.
For visual observations, each frame contains a third-person RGB image and a wrist-mounted RGB image, both rendered at a fixed resolution of $256 \times 256$.
The proprioceptive state provides a complete description of the robot configuration at each step, including the end-effector position and orientation, the gripper joint position and velocity, the arm joint angles, and the corresponding arm joint velocities.
The arm joint angles are additionally stored in a trigonometric encoding to provide a continuous representation that avoids the wrap-around discontinuity of raw angle values.
The action recorded at each step is a continuous end-effector control command consisting of relative translation, delta rotation, and a binary gripper open/close command.
Each step also records an episode-termination flag.
In addition, each episode stores the complete simulator state of all scene entities as a JSON record in the \texttt{sim\_states} field, enabling the exact environment configuration to be reconstructed for replay or re-rendering.
Figure~\ref{fig:dataset_layout} illustrates the internal layout of a generated dataset and the per-step quantities stored in each episode.

\begin{figure}[t]
\begin{center}
\begin{minipage}{0.70\linewidth}
\scriptsize
\begin{verbatim}
{
  "object_registry": {
    "milk_carton": {
      "category": "carton_of_milk",
      "scale": [0.677, 0.500, 0.643],
      "pos":   [-0.132, -0.239, 0.070],
      "ori":   [-0.000, 0.000, 0.702, 0.712]
    },
    "wicker_basket": { "category": "wicker_basket", "pos": [...], "ori": [...] },
    "tin":           { "category": "tin_can",       "pos": [...], "ori": [...] },
    "...":           "remaining grocery objects",
    "robot":         { "joint_pos": [...], "root_link": { "pos": [...] } }
  },
  "camera_info": {
    "third_person": { "image_size": [256, 256], "pos": [...], "ori": [...] },
    "wrist":        { "image_size": [256, 256], "pos": [...], "ori": [...],
                      "mounted_on": "robot_end_effector" }
  },
  "randomization_info": {
    "floor_texture": "background_texture/440.png"
  }
}
\end{verbatim}
\end{minipage}
\end{center}
\caption{Representative excerpt of the per-episode simulation state. Each object entry stores its semantic category, scale, and 6-DoF pose (position and orientation quaternion). The camera configuration records the third-person and wrist camera poses, while the randomization metadata records the sampled visual variation. Numeric values are rounded and repeated entries are abbreviated for readability.}
\label{fig:sim_state}
\end{figure}

\paragraph{Scene Configuration.}
The scene layout is fully described by the per-episode simulation state stored with each demonstration, which records the pose of every object together with the robot configuration, so that the exact initial arrangement of an episode can be recovered.
For the representative \textit{Put milk in basket} task, the scene is a tabletop grocery setting containing seven manipulable objects: the target milk carton, the wicker basket that serves as the goal container, and five additional grocery items, namely a tin, two cardboard food boxes, a milk-carton package, and an orange juice carton.
At the beginning of each episode, the planar position of every object is randomized within an approximately $5\,\text{cm}$ region around its reference placement, and its yaw is randomized over the full in-plane range, while physically valid and task-feasible layouts are preserved.
The robot base pose is held fixed across episodes, and only a small initial joint perturbation is applied to the arm.
In addition to the object and robot states, the simulation state records the configuration of both the third-person and wrist-mounted cameras, including their poses and intrinsics.
The third-person camera provides an external viewpoint and is the only camera subject to optional viewpoint randomization, whereas the wrist camera is rigidly attached to the robot end-effector and therefore moves together with the arm rather than being randomized independently.
Recording these camera configurations makes the viewpoint from which each observation is rendered part of the recoverable scene description.
Because all objects, rather than only the target, are repositioned at the start of each episode, the resulting layouts exhibit substantial spatial diversity and require the policy to localize and act upon the correct object from its visual appearance instead of relying on a fixed, memorized location.
Figure~\ref{fig:sim_state} shows a representative excerpt of the simulation state recorded for a single episode, including the per-object pose, the semantic category and scale of each object, the camera configuration, and the visual randomization applied to the scene.



\begin{figure}[t]
\centering
\newlength{\scenerowh}\newlength{\sceneha}\newlength{\scenehb}
\settoheight{\sceneha}{\includegraphics[width=0.4\linewidth]{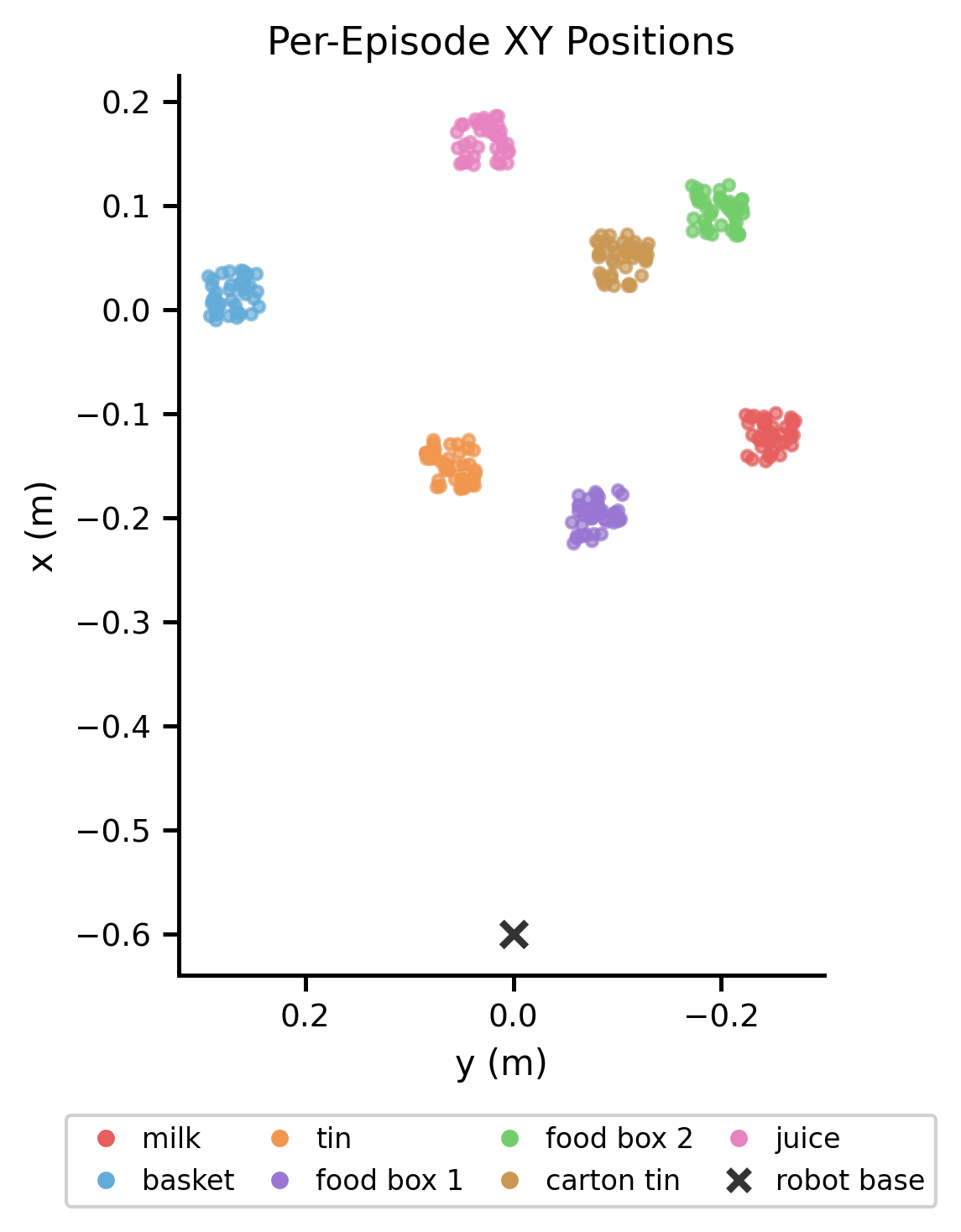}}
\settoheight{\scenehb}{\includegraphics[width=0.52\linewidth]{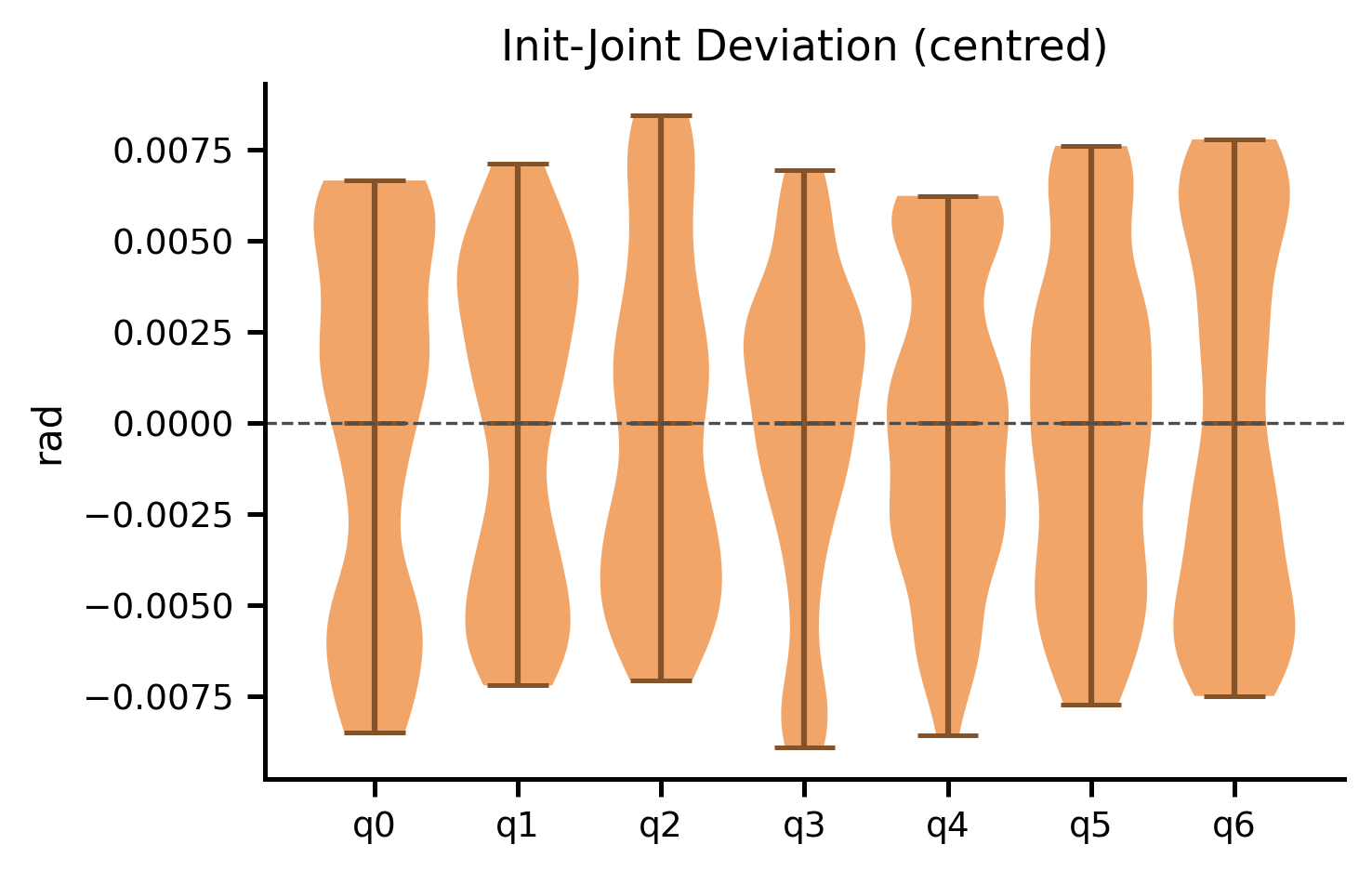}}
\setlength{\scenerowh}{\sceneha}\ifdim\scenehb>\scenerowh\setlength{\scenerowh}{\scenehb}\fi
\subfloat[Per-episode object positions]{\begin{minipage}[b][\scenerowh][c]{0.4\linewidth}\centering\includegraphics[width=\linewidth]{scene_stat_xy_positions_v2.png}\end{minipage}}
\hfill
\subfloat[Init-joint deviation]{\begin{minipage}[b][\scenerowh][c]{0.52\linewidth}\centering\includegraphics[width=\linewidth]{scene_stat_init_joint_deviation_v2.png}\end{minipage}}
\caption{Scene-level statistics of a representative PRISM-generated dataset for the \textit{Put milk in basket} task. \textbf{(a)} Per-episode planar positions of all scene objects in the world frame, with the fixed robot base marked by a red cross, showing the spatial spread induced by object pose randomization. \textbf{(b)} Centered per-joint deviation of the robot initialization across episodes.}
\label{fig:dataset_scene_stats}
\end{figure}

\paragraph{Scene Statistics.}
Table~\ref{tab:supp_scene_statistics} summarizes the scene-level statistics of a representative generated dataset for the \textit{Put milk in basket} task.
The dataset is built from 4 procedurally generated scene variants, each contributing 10 retained trajectories, yielding 40 distinct spatial configurations, each of which is rendered under 10 independent visual randomizations, for a total of 400 demonstrations.
Each scene contains 7 manipulable objects, comprising the target, the goal container, and 5 additional grocery objects, all of which are repositioned per episode within an approximately $5\,\text{cm}$ planar region, while the robot base remains fixed and only a small per-joint initialization noise is applied.
Figure~\ref{fig:dataset_scene_stats} visualizes these scene-level statistics: the per-episode planar positions of all scene objects in the world frame, and the centered per-joint deviation of the robot initialization.

\begin{table*}[t]
\centering
\begin{tabular}{ll}
\toprule
\textbf{Property} & \textbf{Value} \\
\midrule
Scene variants & 4 \\
Spatial configurations & 40 \\
Visual randomizations per configuration & 10 \\
Manipulable objects per scene & 7 (1 target + 1 goal + 5 others) \\
Object planar position range & $\sim 5\,\text{cm}$ \\
Object yaw randomization & full in-plane \\
Robot base pose & fixed \\
Robot initial joint noise (per joint) & $\sim 0.015\,\text{rad}$ \\
Third-person camera & fixed external viewpoint \\
Wrist camera & rigidly mounted on end-effector \\
\bottomrule
\end{tabular}
\caption{Scene-level statistics of a representative PRISM-generated dataset (\textit{Put milk in basket}).}
\label{tab:supp_scene_statistics}
\end{table*}

\begin{table*}[t]
\centering
\begin{tabular}{ll}
\toprule
\textbf{Property} & \textbf{Value} \\
\midrule
\multicolumn{2}{l}{\textit{Scale}} \\
Number of demonstrations & 400 \\
Success rate of demonstrations & 100\% \\
Total time steps & 50{,}748 \\
Episode length (mean / std / min / max) & 127 / 12 / 108 / 152 \\
\midrule
\multicolumn{2}{l}{\textit{Motion}} \\
End-effector path length (mean) & $\sim 1.11\,\text{m}$ \\
Gripper state transitions per episode & 2 \\
Per-step translation magnitude & $\leq 0.02\,\text{m}$ \\
Per-step rotation magnitude & $\leq 0.08\,\text{rad}$ \\
\bottomrule
\end{tabular}
\caption{Trajectory-level statistics of a representative PRISM-generated dataset (\textit{Put milk in basket}).}
\label{tab:supp_trajectory_statistics}
\vspace{-15pt}
\end{table*}
\begin{figure}[h]
\centering
\subfloat[Episode length]{\includegraphics[width=0.45\linewidth]{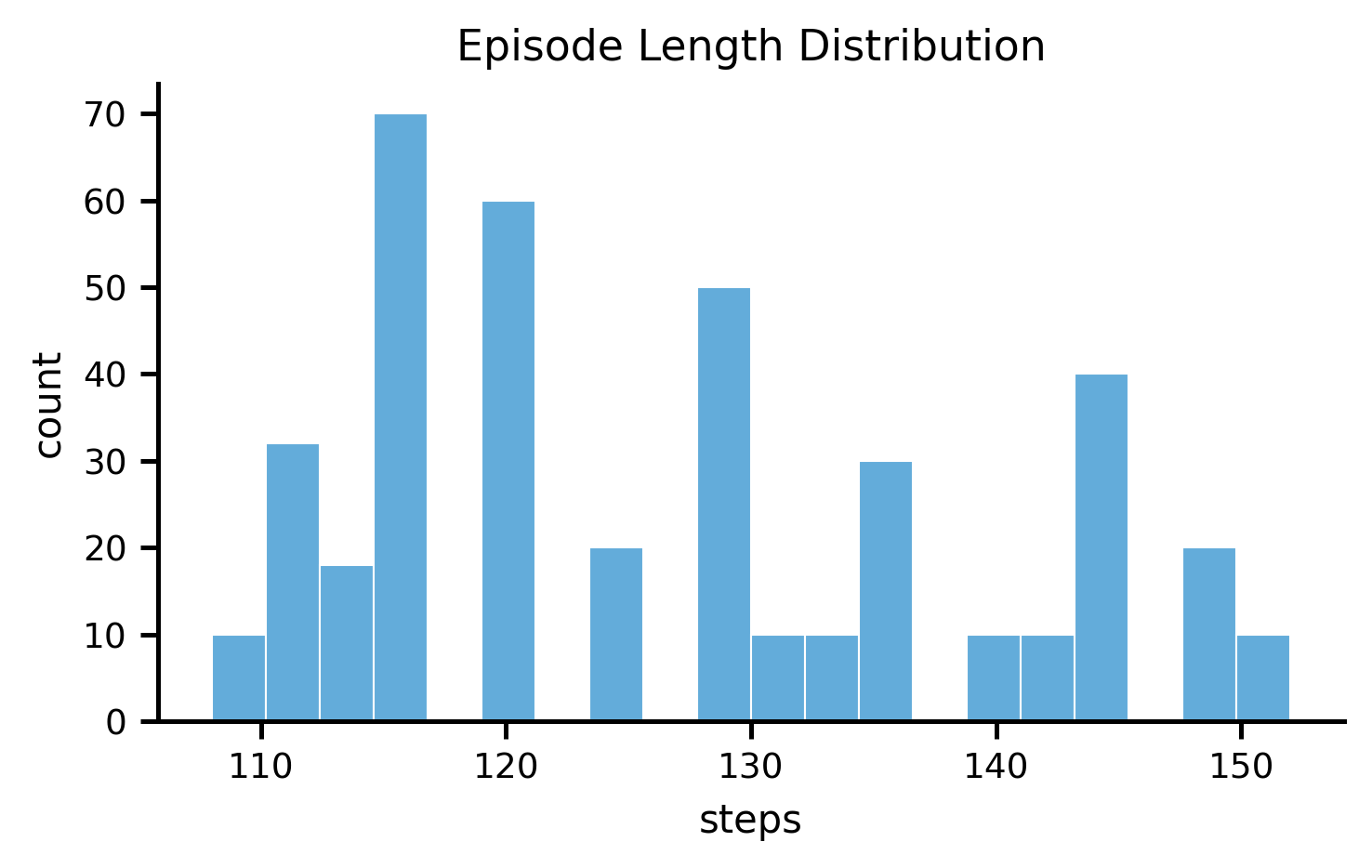}}
\hfill
\subfloat[Action range]{\includegraphics[width=0.45\linewidth]{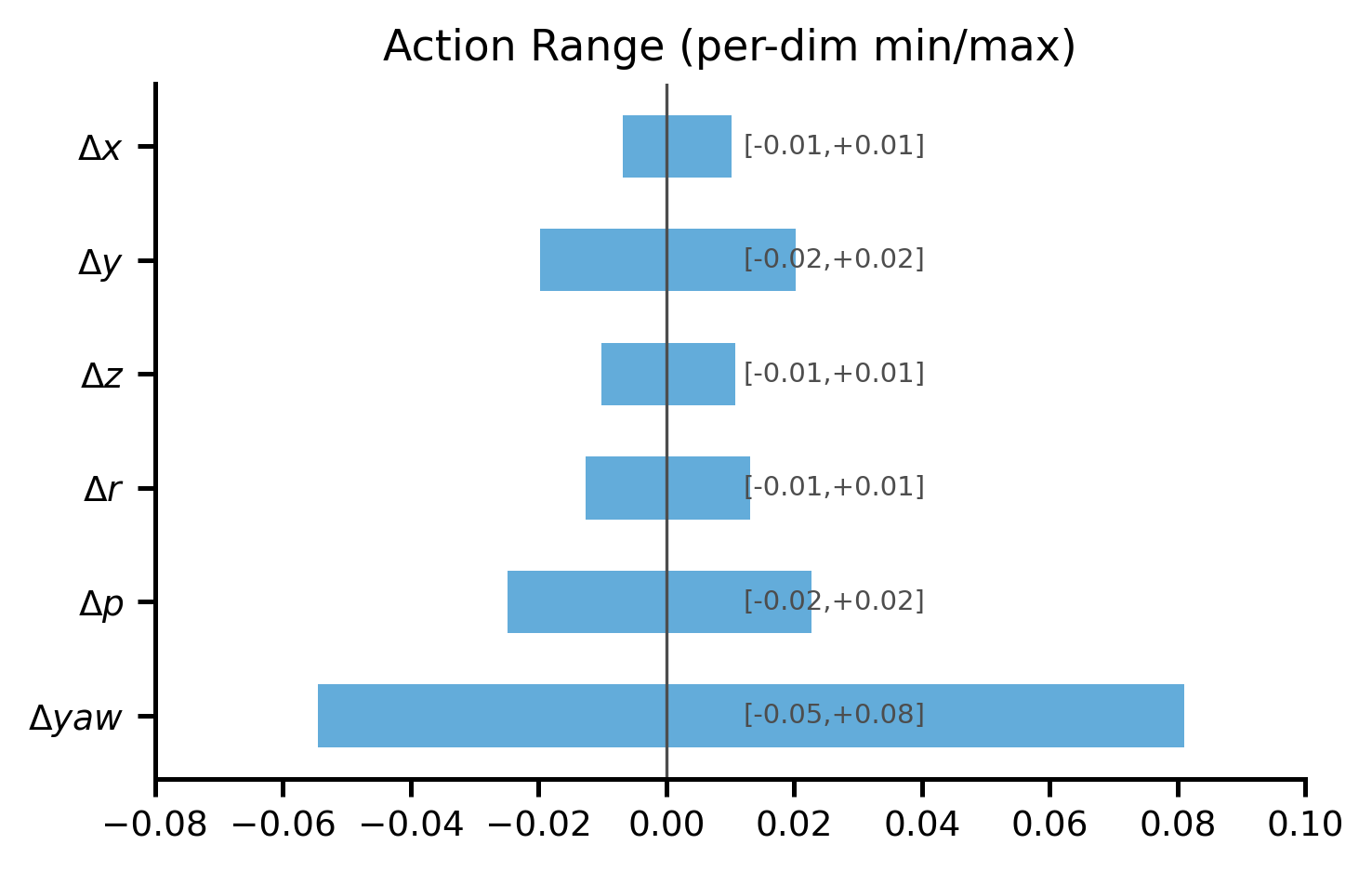}}
\\
\subfloat[EEF trajectories]{\includegraphics[width=0.55\linewidth]{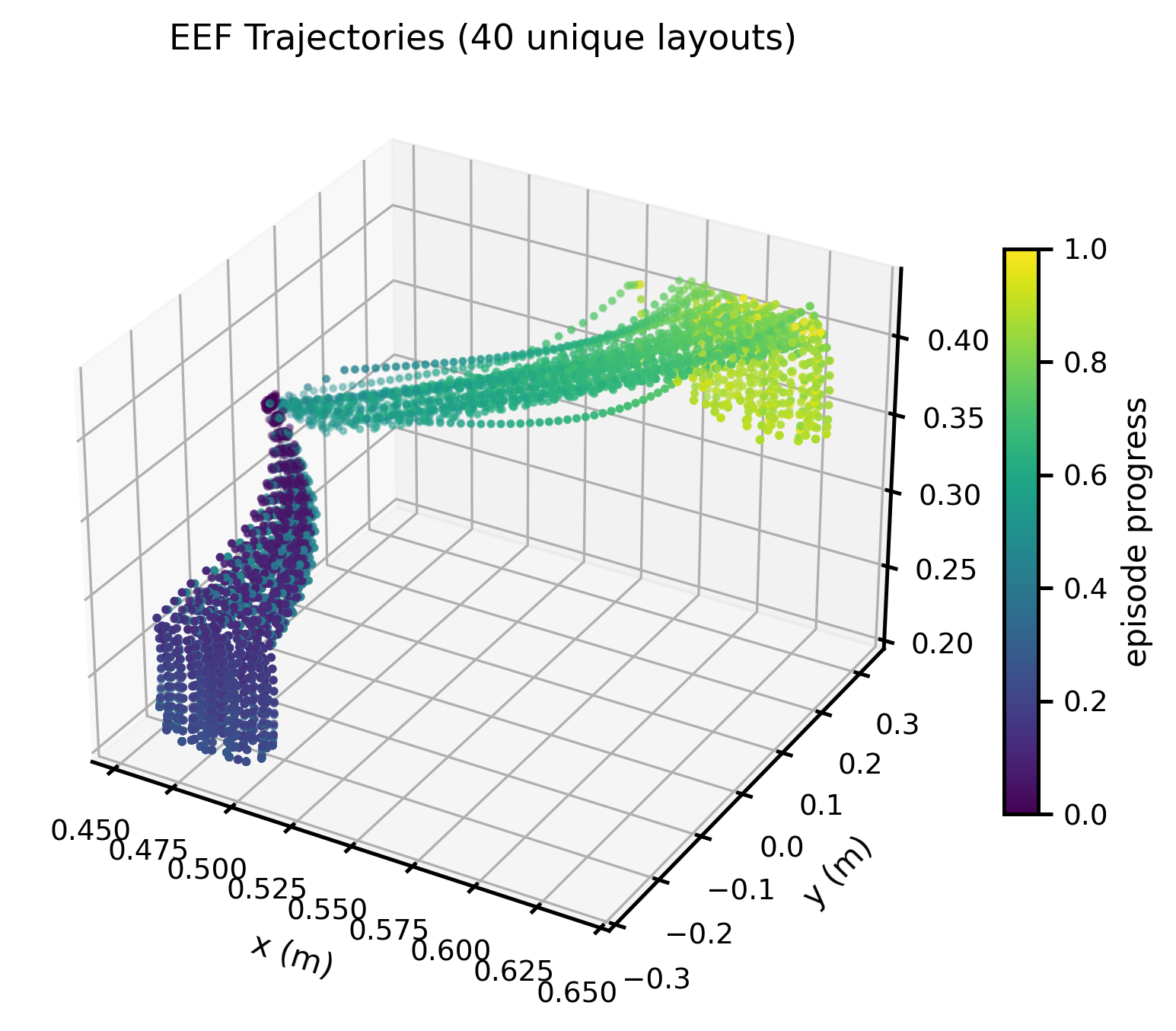}}
\caption{Trajectory-level statistics of a representative PRISM-generated dataset (\textit{Put milk in basket}). \textbf{(a)} Distribution of episode lengths. \textbf{(b)} Per-dimension minimum/maximum range of the end-effector action increments (the binary gripper command is excluded). \textbf{(c)} End-effector trajectories across the distinct spatial layouts, colored by episode progress.}
\label{fig:dataset_trajectory_stats}
\vspace{-20pt}
\end{figure}

\paragraph{Trajectory Statistics.}
Table~\ref{tab:supp_trajectory_statistics} summarizes the trajectory-level statistics of the same dataset.
The dataset contains 400 successful demonstrations spanning a total of 50{,}748 time steps, with an average episode length of approximately 127 steps (ranging from 108 to 152).
On average, each demonstration traces an end-effector path of about $1.11\,\text{m}$ and contains a single grasp-and-release cycle, reflected by two gripper state transitions per episode.
The per-step end-effector commands are small and smooth, with translation increments bounded by about $0.02\,\text{m}$ and rotation increments by about $0.08\,\text{rad}$, consistent with closed-loop control.
All demonstrations satisfy the task success predicate, yielding a fully successful demonstration set.
Figure~\ref{fig:dataset_trajectory_stats} visualizes these trajectory-level statistics, including the distribution of episode lengths, the per-dimension range of the end-effector action increments (excluding the binary gripper command), and the spatial spread of end-effector trajectories across layouts.


\section{Baseline Pipelines}
\label{sec:supp_baseline_pipeline}

\subsection{X-Sim Pipeline Details}
\label{sec:supp_xsim}

X-Sim~\cite{xsim} is a pipeline that synthesizes imitation learning trajectories in simulation from a single human demonstration video. The pipeline first reconstructs the target object and the workspace as a digital twin, extracts a 6-DoF object trajectory from the demonstration video, trains a manipulation policy via reinforcement learning using the extracted trajectory as the reference for its reward, and finally rolls out the trained policy in the simulator to synthesize trajectories for behavior cloning.

\paragraph{Real-to-Sim Reconstruction.}
This section describes how we reconstructed the digital twin for each task.
For tasks with a real physical setup (\textit{Box into basket}, \textit{Stack bowls}, \textit{Lift cup}), 
the source images were captured directly in the physical environment.
For LIBERO benchmark tasks (\textit{Put milk in basket}, \textit{Put wine bottle on cabinet}), 
the corresponding LIBERO simulator was used as the visual source.

Object meshes were obtained using the Polycam~\cite{polycam} smartphone application only for the objects that the robot directly interacts with during the task. For each such object, we captured approximately 100 photographs from a 360° viewing arc, after which Polycam automatically generated a mesh. The mesh was then imported into Blender~\cite{blender}, where we rescaled it to match the real-world physical dimensions, set its center of mass, and exported the final result in the GLB format.

\begin{figure}[h]
    \centering

    \begin{minipage}[b]{0.30\linewidth}
        \centering
        \includegraphics[width=\linewidth, height=3cm, keepaspectratio]{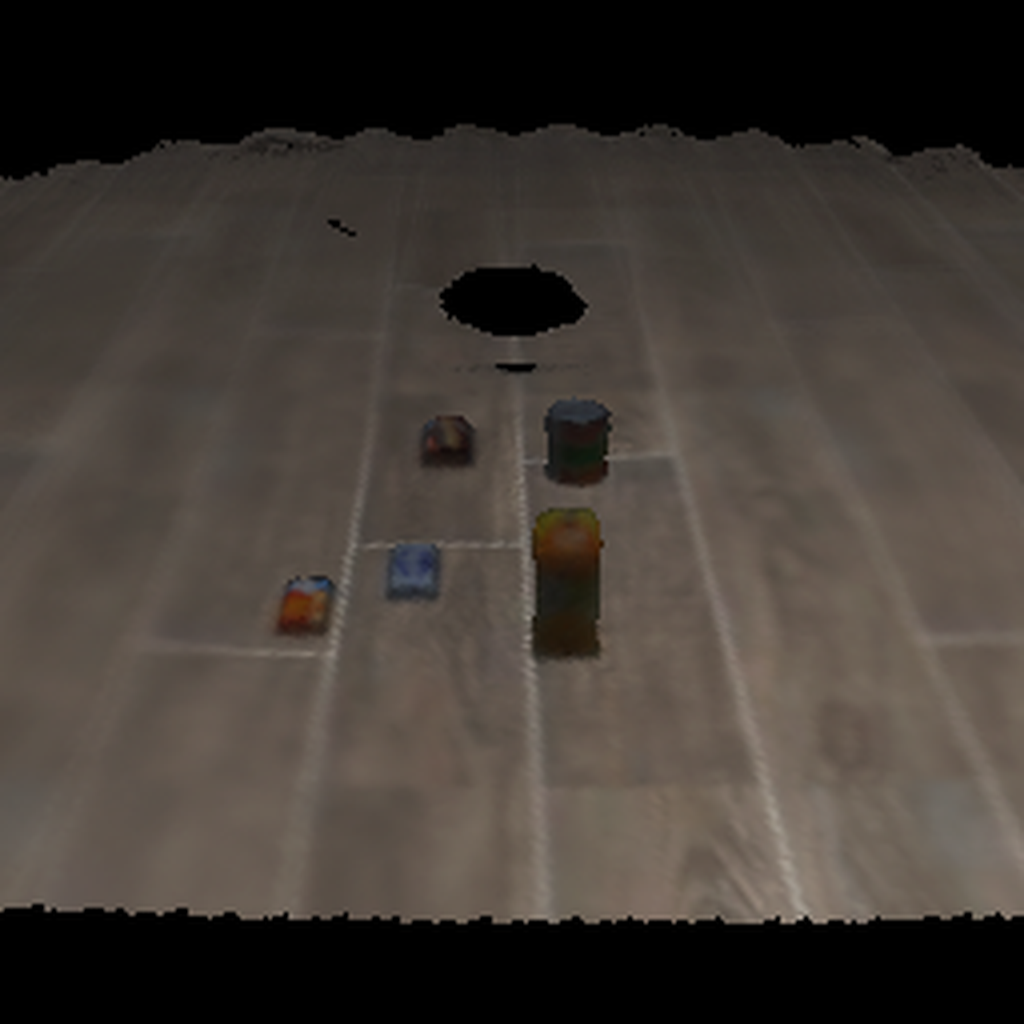}\\
        \vspace{0.3em}
        (a) X-Sim Scene
    \end{minipage}
    \hspace{0.04\linewidth}
    \begin{minipage}[b]{0.34\linewidth}
        \centering
        \begin{minipage}[b]{0.46\linewidth}
            \centering
            \includegraphics[width=\linewidth, height=2.5cm, keepaspectratio]{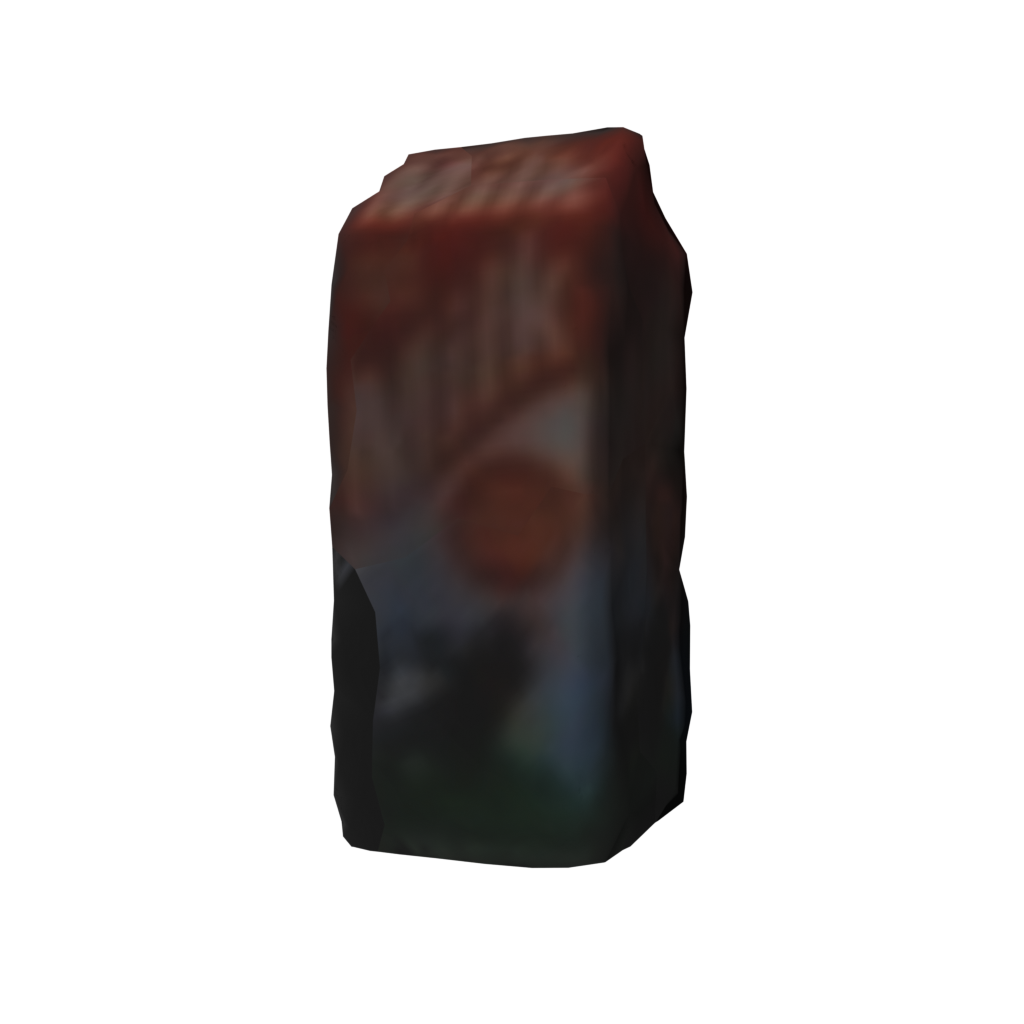}\\
            \vspace{0.3em}
            {\small Milk}
        \end{minipage}
        \hfill
        \begin{minipage}[b]{0.46\linewidth}
            \centering
            \includegraphics[width=\linewidth, height=2.5cm, keepaspectratio]{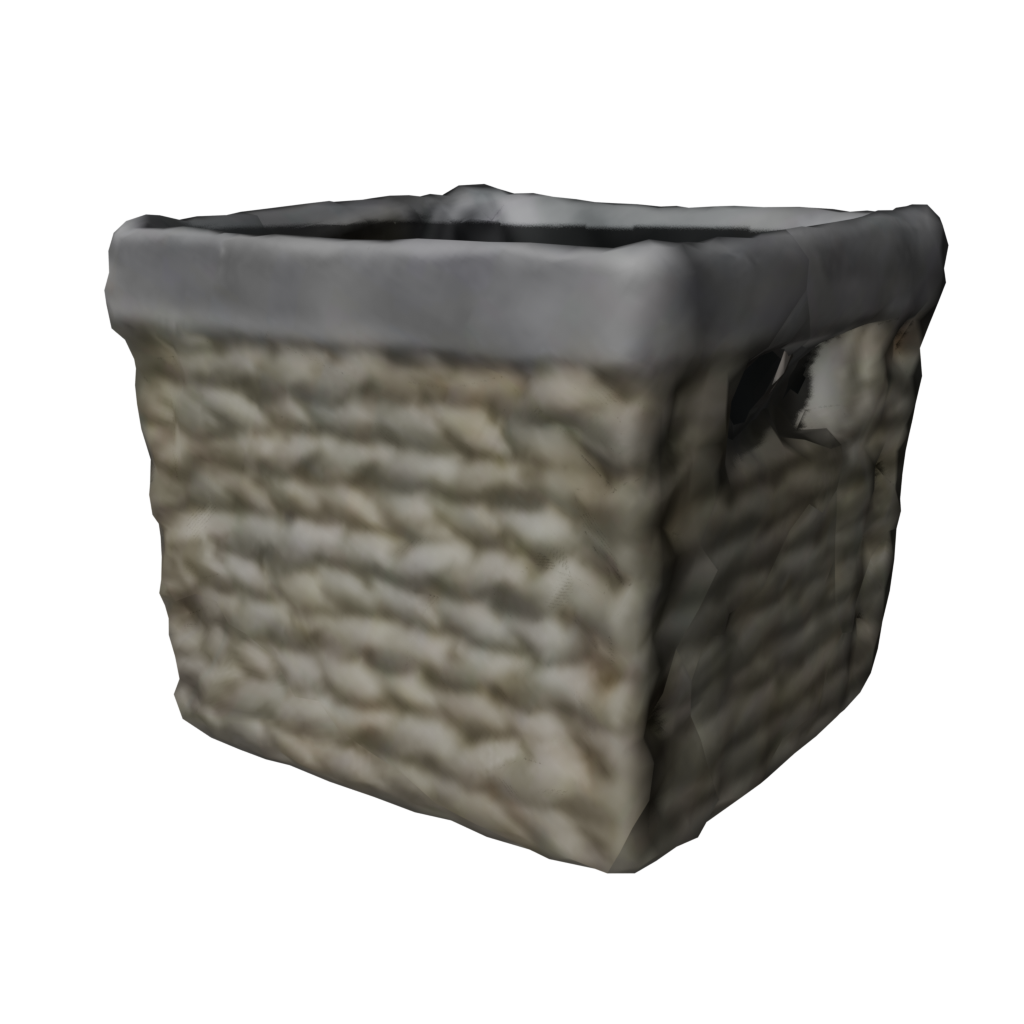}\\
            \vspace{0.3em}
            {\small Basket}
        \end{minipage}\\
        \vspace{0.3em}
        (b) Reconstructed Objects
    \end{minipage}

    \caption{X-Sim assets for the \textit{Put milk in basket} task: \textbf{(a)} the reconstructed tabletop scene from 2D Gaussian Splatting (the robot is hidden for visualization), and \textbf{(b)} the two interactive objects (milk and basket) reconstructed 
with Polycam.}
    \label{fig:xsim_scene_objects}
\end{figure}

Workspace scenes were reconstructed via 2D Gaussian Splatting~\cite{gaussiansplatting2d}. We recorded a video of the real workspace for roughly one minute from a diverse range of viewpoints, converted it into frame images, and used these as input to train a 2DGS model. The resulting scene mesh contained all non-interactive elements—the workspace surface, the background, and any stationary scene objects—onto which the Polycam-reconstructed interactive objects were subsequently placed. Once training had converged, the splat field was baked into a mesh, which we then cleaned up in Blender by removing extraneous points and rescaling it to match the real-world dimensions before aligning it to the simulator's world coordinate frame. Figure~\ref{fig:xsim_scene_objects} shows the reconstructed scene and objects used in the \textit{Put milk in basket} task.

\paragraph{Trajectory Extraction.}
The object trajectory that anchors the reward signal is obtained by applying FoundationPose~\cite{foundationpose} frame-by-frame to a single RGB demonstration video, yielding a sequence of 6-DoF object poses. At the beginning of each training episode, 30 waypoints are uniformly subsampled from this sequence and used as the reference trajectory for the reinforcement learning agent. To avoid lateral collisions during the approach phase, an additional hovering waypoint is inserted $10\,\text{cm}$ above the first grasp waypoint.

\paragraph{Reinforcement Learning.}
Each task environment consists of a Panda 7-DoF arm equipped with a 
parallel-jaw gripper, the reconstructed workspace mesh, a dynamic 
body for the target object, and a kinematic target container. The 
environment is equipped with an external camera that observes the 
workspace from an oblique angle and a wrist-mounted egocentric 
camera, with the external camera's intrinsics and extrinsics matched 
to those used in the real-world demonstration. Both cameras render 
at a resolution of $256 \times 256$.

The observation is a low-dimensional state vector composed of joint 
positions and velocities, gripper width, end-effector and object 
poses, the current target waypoint, and the relative offsets from 
the end-effector to the object and from the object to the target 
waypoint. The action is a 7-dimensional continuous vector consisting 
of a 3-DoF translational displacement of the end-effector, a 3-DoF 
rotational displacement, and a scalar gripper command (RGB images 
are rendered only during the dataset generation stage).

The reward function is a dense composite of multiple terms that 
jointly guide the policy through the canonical phases of the task:
\begin{equation*}
r = r_\text{reach} + r_\text{track} + r_\text{static}
    + r_\text{tilt} + r_\text{grip} + r_\text{success}.
\end{equation*}
Each term is defined as follows.
\begin{equation*}
\begin{array}{l@{\;}c@{\;}l}
\lhs{r_\text{reach}}   & = & 
  \begin{cases}
  \bigl(1 - \tanh(k_d \, d_\text{obj})\bigr)
  + \bigl(1 - \tanh(k_o \, \epsilon_o)\bigr) & \text{if not grasped} \\
  2.0 & \text{otherwise}
  \end{cases} \\[8pt]
\lhs{r_\text{track}}   & = & 
  2\bigl(1 - \tanh(\alpha_i \, d_i)\bigr)
  \cdot \mathbf{1}_{\text{hover} \,\lor\, \text{grasped}}
  + 2 \cdot \mathbf{1}_\text{advance} \\[8pt]
\lhs{r_\text{static}}  & = & 
  \bigl(1 - \tanh(k_v \, \|\dot{q}\|)\bigr)
  \cdot \mathbf{1}_\text{at\_goal}
  \cdot \mathbf{1}_\text{last\_wp} \\[8pt]
\lhs{r_\text{tilt}}    & = & 
  -0.5 \cdot \max(0,\ \Delta\theta - \theta_\text{max}) \\[8pt]
\lhs{r_\text{grip}}    & = & 
  \begin{cases}
  +0.3\ (\text{open}) \,/\, -0.1\ (\text{closed})
     & \text{approach phase } (i \leq 2) \\
  +1.2\ (\text{grasped}) \,/\, -0.2\ (\text{otherwise})
     & \text{grasp phase } (i > 2)
  \end{cases} \\[8pt]
\lhs{r_\text{success}} & = & 7.0 \cdot \mathbf{1}_\text{success}
\end{array}
\end{equation*}
The variables are defined as
\begin{equation*}
\begin{array}{l@{\;}c@{\;}l}
\lhs{d_\text{obj}} & = & \|p_\text{ee} - p_\text{obj}\| \\[3pt]
\lhs{\epsilon_o}   & = & 
  \text{alignment error between gripper $z$-axis and workspace normal} \\[3pt]
\lhs{d_i}          & = & 
  \begin{cases}
  \|p_\text{ee}  - p_{\text{wp},0}\| & \text{if } i = 0 \\
  \|p_\text{obj} - p_{\text{wp},i}\| & \text{if } i \geq 1
  \end{cases} \\[3pt]
\lhs{\alpha_i}     & = & 1 / \|p_{\text{wp},i} - p_{\text{wp},i-1}\| \\[3pt]
\lhs{\Delta\theta} & = & 
  \text{rotation deviation from the held object's initial orientation}
\end{array}
\end{equation*}
Here $i \in \{0, \ldots, N\}$ indexes the current waypoint 
($i=0$ corresponds to the hovering pose and $i=N$ to the final 
placement), and $\alpha_i$ is the inverse spacing between adjacent 
waypoints, which keeps the tracking gradient smooth across 
non-uniform segments. The indicator functions are 
$\mathbf{1}_\text{hover}$ ($i=0$), 
$\mathbf{1}_\text{grasped}$ (object is currently held), 
$\mathbf{1}_\text{advance}$ (single-step bonus when the agent 
first reaches the next waypoint), 
$\mathbf{1}_\text{at\_goal}$ ($\|p_\text{obj}-p_{\text{wp},N}\| < 5\,\text{cm}$), 
$\mathbf{1}_\text{last\_wp}$ ($i=N$), and 
$\mathbf{1}_\text{success}$ (object grasped, within $5\,\text{cm}$ of the 
final waypoint, and the robot static for 20 consecutive frames). 
We use the constants $k_d = k_o = k_v = 5.0$ and 
$\theta_\text{max} = 0.2\,\text{rad}$.

The policy is optimized with PPO~\cite{ppo} (clipped surrogate + finite-horizon 
GAE) using $1{,}024$ parallel environments, a maximum rollout length of 
$200$, a random seed of $42$, and a maximum of $5 \times 10^{7}$ 
environment steps, following the setup of the original X-Sim paper; training is terminated early once the episode return converges.
Full hyperparameters are listed in Table~\ref{tab:supp_xsim_ppo_hp}.

\begin{table*}[!ht]
\centering
\small
\setlength{\tabcolsep}{5pt}
\begin{tabular}{ll}
\toprule
\textbf{Hyperparameter} & \textbf{Value} \\
\midrule
Learning rate & $3 \times 10^{-4}$ \\
Discount factor ($\gamma$) & 0.8 \\
GAE parameter ($\lambda$) & 0.9 \\
Clipping parameter ($\epsilon$) & 0.2 \\
Value function coefficient & 0.5 \\
Entropy coefficient & 0.0 \\
Target KL divergence & 0.1 \\
Maximum gradient norm & 0.5 \\
Mini-batch size & 6{,}400 \\
Number of parallel environments & 1{,}024 \\
Actor network & MLP (state dim $\rightarrow$ 256 $\rightarrow$ 256 $\rightarrow$ 256 $\rightarrow$ action dim) \\
Critic network & MLP (state dim $\rightarrow$ 256 $\rightarrow$ 256 $\rightarrow$ 256 $\rightarrow$ 1) \\
Activation function & Tanh \\
Optimizer & Adam \\
Adam epsilon & $1 \times 10^{-5}$ \\
\bottomrule
\end{tabular}
\caption{PPO hyperparameters for X-Sim.}
\label{tab:supp_xsim_ppo_hp}
\vspace{-10pt}
\end{table*}

\paragraph{Data Generation.}
Once reinforcement learning has converged, the trained policy is rolled out in deterministic mode for up to 200 steps to collect imitation learning trajectories. At the start of each episode, per-joint uniform noise is added to the robot's initial configuration, and at every step the RGB images from both cameras are stored together with the robot and object states. The recorded action is the actual step-wise $\Delta$pose of the wrist link, stored in the same delta-pose format as PRISM datasets to maintain consistency for baseline comparison.

A trajectory is accepted as successful only when the task-specific success condition is held for 20 consecutive frames. Successful trajectories are saved into the training set, while failed trajectories are stored separately. Each file contains the sequence of observations and actions together with metadata such as asset meshes, camera parameters, lighting, initial poses, and the waypoint sequence.

\begin{figure}[h]
    \centering

    \begin{minipage}[b]{0.22\linewidth}
        \centering
        \includegraphics[width=\linewidth]{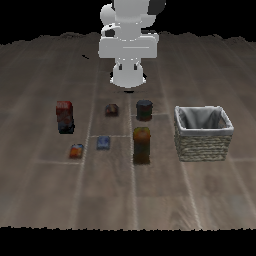}\\
        \vspace{0.3em}
        (a) Clean Scene
    \end{minipage}
    \hspace{0.04\linewidth}
    \begin{minipage}[b]{0.70\linewidth}
        \centering
        \begin{minipage}[b]{0.31\linewidth}
            \centering
            \includegraphics[width=\linewidth]{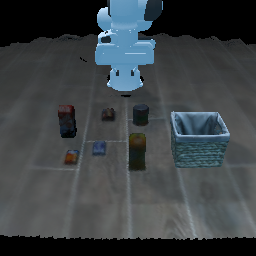}
        \end{minipage}
        \hfill
        \begin{minipage}[b]{0.31\linewidth}
            \centering
            \includegraphics[width=\linewidth]{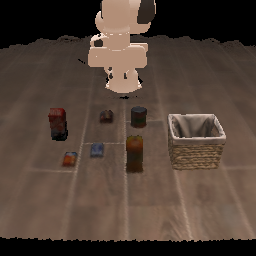}
        \end{minipage}
        \hfill
        \begin{minipage}[b]{0.31\linewidth}
            \centering
            \includegraphics[width=\linewidth]{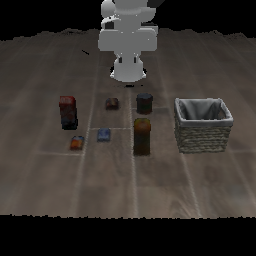}
        \end{minipage}\\
        \vspace{0.3em}
        (b) Randomized scenes
    \end{minipage}

    \caption{Visualization of clean and randomized X-Sim scenes.}
    \label{fig:xsim_clean_random_scenes}
    \vspace{-10pt}
\end{figure}

\paragraph{Domain Randomization.}
At the beginning of every episode, X-Sim applies four kinds of randomization: lighting, external camera pose, initial position of the target object, and initial robot joint configuration. Lighting is sampled uniformly from a library of four presets, each specifying the ambient illumination together with main and fill light parameters. The external camera is perturbed in position around its reference location, while its look-at target and up vector are kept fixed. Only the target object that the robot directly manipulates is perturbed in its initial horizontal position around the demonstration's starting location, and all non-target scene objects remain at their reference locations. The robot receives independent uniform noise on every joint to prevent overfitting to a single deterministic starting pose. The camera intrinsics, the wrist camera, asset shapes and textures, non-target scene objects, and the geometry of the workspace are held fixed across episodes to preserve the consistency of the digital twin itself. Figure~\ref{fig:xsim_clean_random_scenes} shows representative clean and randomized scenes, and the distribution and range of each randomization are listed in Table~\ref{tab:supp_xsim_rand_params}.

\vspace{-5pt}
\begin{table*}[!ht]
\centering
\begin{tabular}{llll}
\toprule
\textbf{Component} & \textbf{Parameter} & \textbf{Distribution} & \textbf{Range} \\
\midrule
Lighting        & Preset (ambient, main, fill)    & Categorical & 4 presets \\
External camera & Position offset along each axis & Uniform     & $\pm 3\,\text{cm}$ \\
Target object          & Horizontal position offset      & Uniform     & $\pm 2.5\,\text{cm}$ \\
Robot           & Per-joint initialization offset & Uniform     & $\pm 0.02\,\text{rad}$ \\
\bottomrule
\end{tabular}
\caption{Randomization parameters applied at episode initialization in X-Sim.}
\label{tab:supp_xsim_rand_params}
\vspace{-10pt}
\end{table*}

\paragraph{Dataset Statistics.}

\begin{table*}[!ht]
\centering

\begin{tabular}{lccccc}
\toprule
\textbf{Task} & \textbf{RL training} & \textbf{Successes} & 
\textbf{Attempts} & \textbf{Success rate} & \textbf{Collection} \\
\midrule
Box into basket                  & $\sim$1\,h\,4\,min  & 400 & 401 & 99.75\% & 20\,min  \\
Stack bowls            & $\sim$1\,h\,16\,min & 400 & 439 & 91.11\% & 43\,min  \\
Lift cup                    & $\sim$1\,h\,40\,min & 400 & 401 & 99.75\% & 27\,min      \\
Put milk in basket          & $\sim$17\,min       & 400 & 620 & 64.51\% & 90\,min  \\
Put wine bottle on cabinet  & $\sim$22\,min       & 400 & 976 & 40.98\% & 129\,min \\
\bottomrule
\end{tabular}
\caption{Per-task reinforcement learning training time, dataset 
         generation success rate, and collection time for X-Sim.}
\label{tab:supp_xsim_dataset_stats}
\end{table*}

Table~\ref{tab:supp_xsim_dataset_stats} reports the per-task RL training time, 
dataset generation success rate, and collection time. On the three tasks reconstructed directly from real-world observations (\textit{Box into basket}, \textit{Stack bowls}, \textit{Lift cup}), the trained policy yielded generation success rates above $90\%$. The LIBERO benchmark tasks required longer episodes to avoid the abrupt, large per-step motions that short rollouts induce; this longer horizon made the trajectories more prone to compounding errors and thus lowered the generation success rate on \textit{Put milk in basket} and \textit{Put wine bottle on cabinet}.

\subsection{RoboTwin 2.0 Pipeline Details}
\label{sec:supp_robotwin}

The RoboTwin 2.0~\cite{robotwin2.0} baseline pipeline synthesizes expert demonstrations for each task by combining simulation-based scene generation with automated demonstration collection under structured domain randomization. The pipeline proceeds in three stages: scene generation, domain randomization, and demonstration generation. A total of 400 successful trajectories are collected per task.

\begin{figure}[h]
    \centering

    \newcommand{\objectpair}[3]{
        \begin{minipage}[t]{0.22\linewidth}
            \centering
            {\small\textbf{#1}}\\[-0.1em]
            {\hspace{0.4em} \scriptsize LIBERO \hspace{1.7em} RoboTwin~2.0}\\[0.3em]
            \makebox[0.48\linewidth][c]{\includegraphics[height=1.4cm,width=0.46\linewidth,keepaspectratio]{#2}
            
            }
            \hfill
            \makebox[0.48\linewidth][c]
            {\includegraphics[height=1.4cm,width=0.46\linewidth,keepaspectratio]{#3}
            }
        \end{minipage}
    }

    \objectpair{Milk}{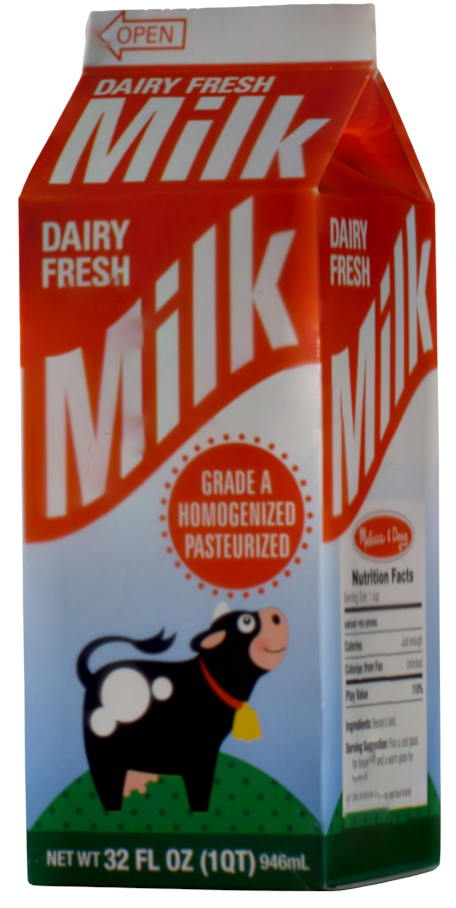}{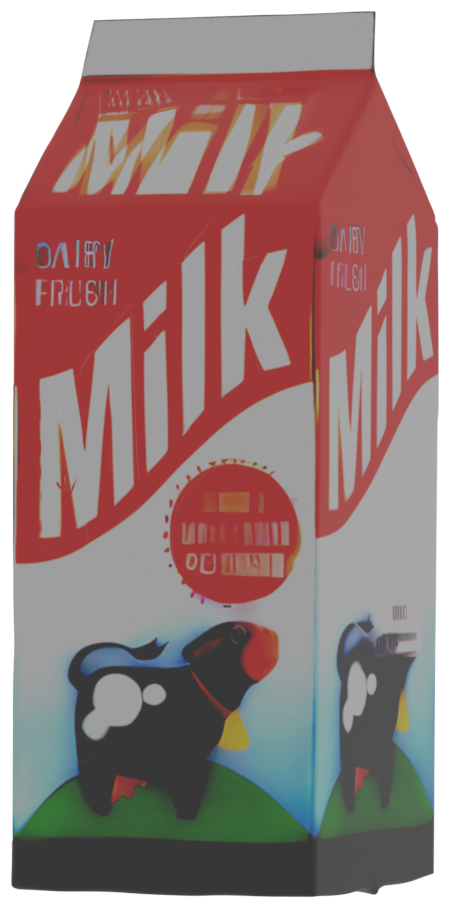}
    \hfill
    \objectpair{Orange juice}{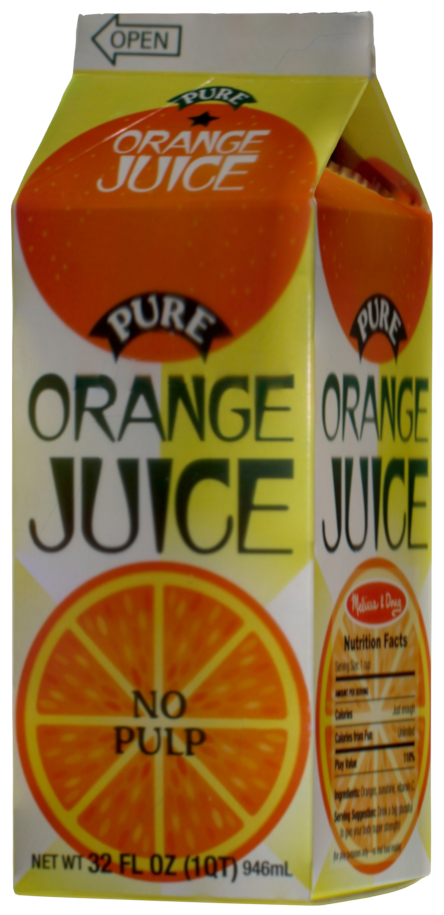}{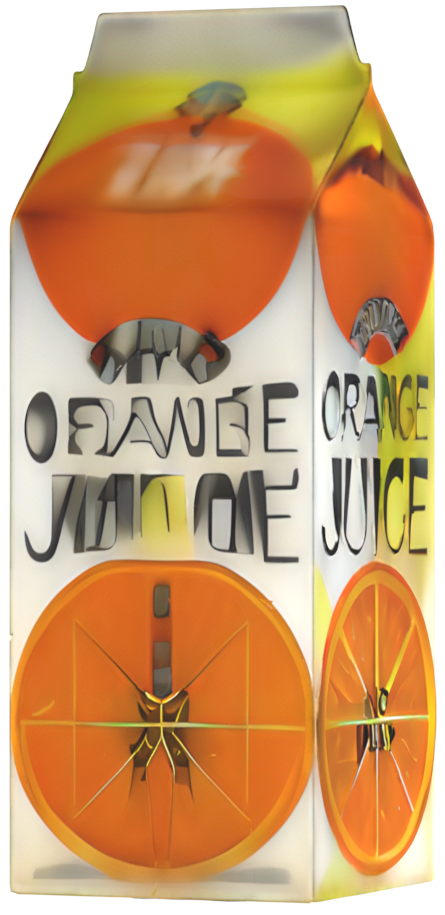}
    \hfill
    \objectpair{Butter}{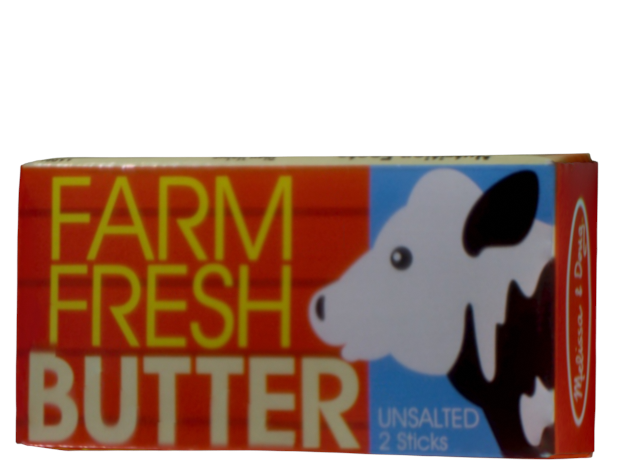}{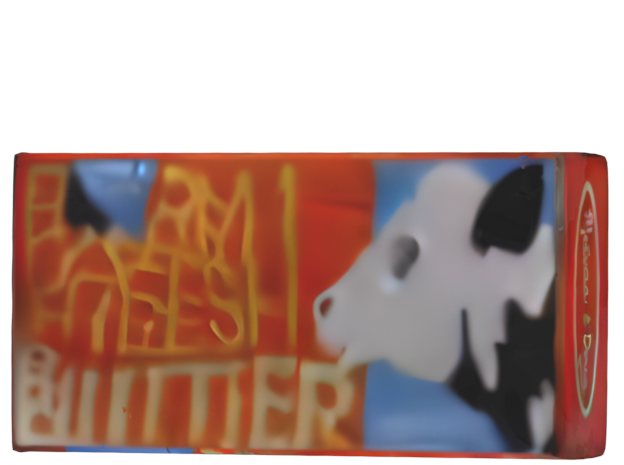}
    \hfill
    \objectpair{Tomato sauce}{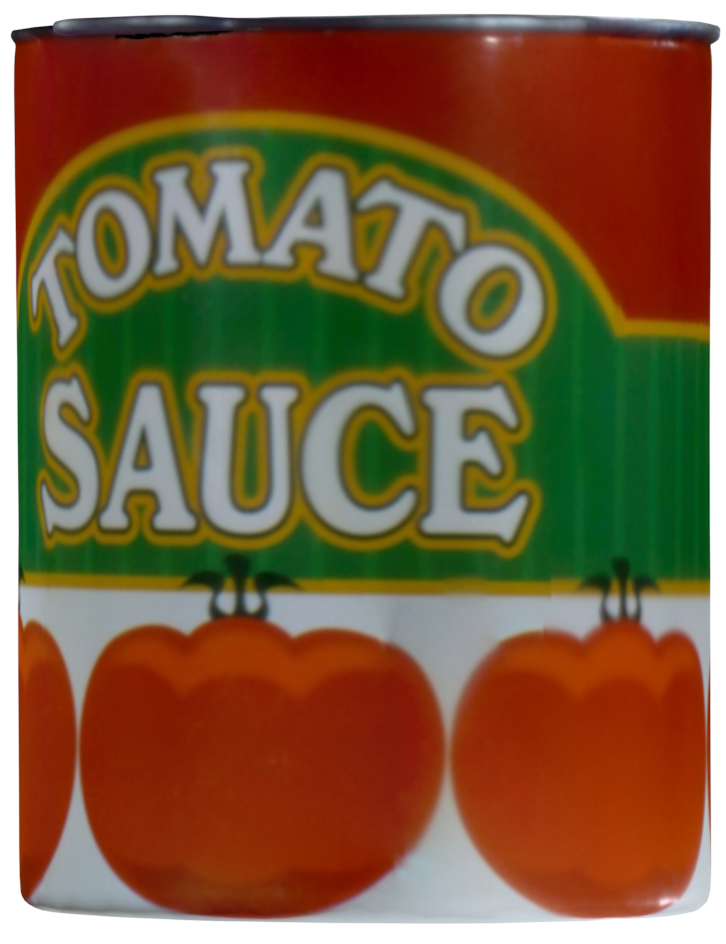}{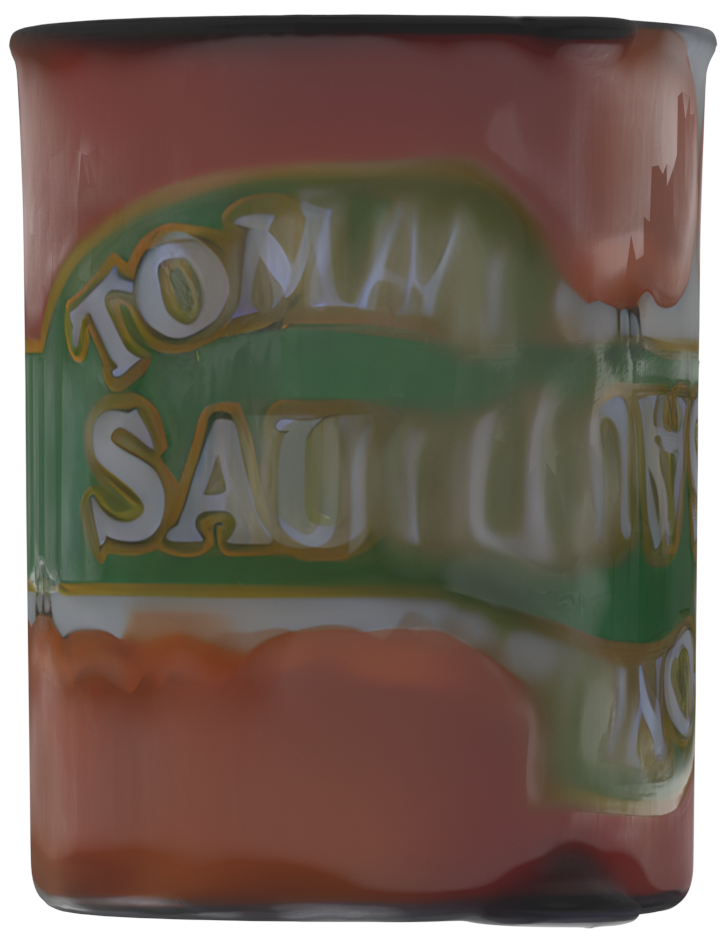}

    \vspace{1.2em}

    \objectpair{Basket}{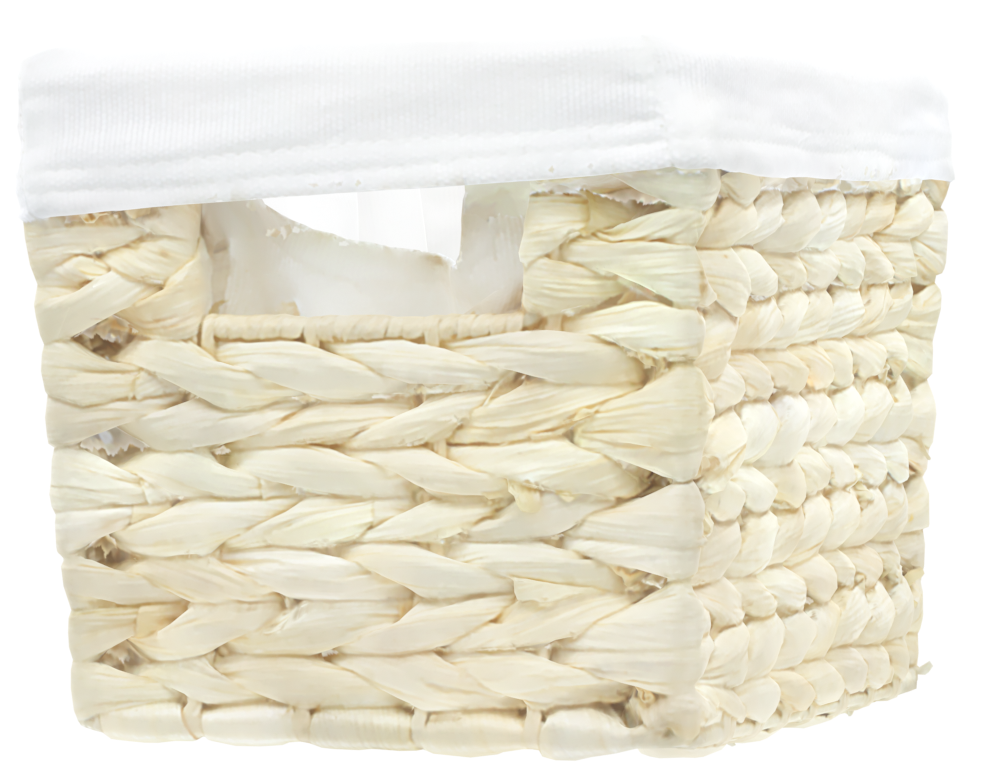}{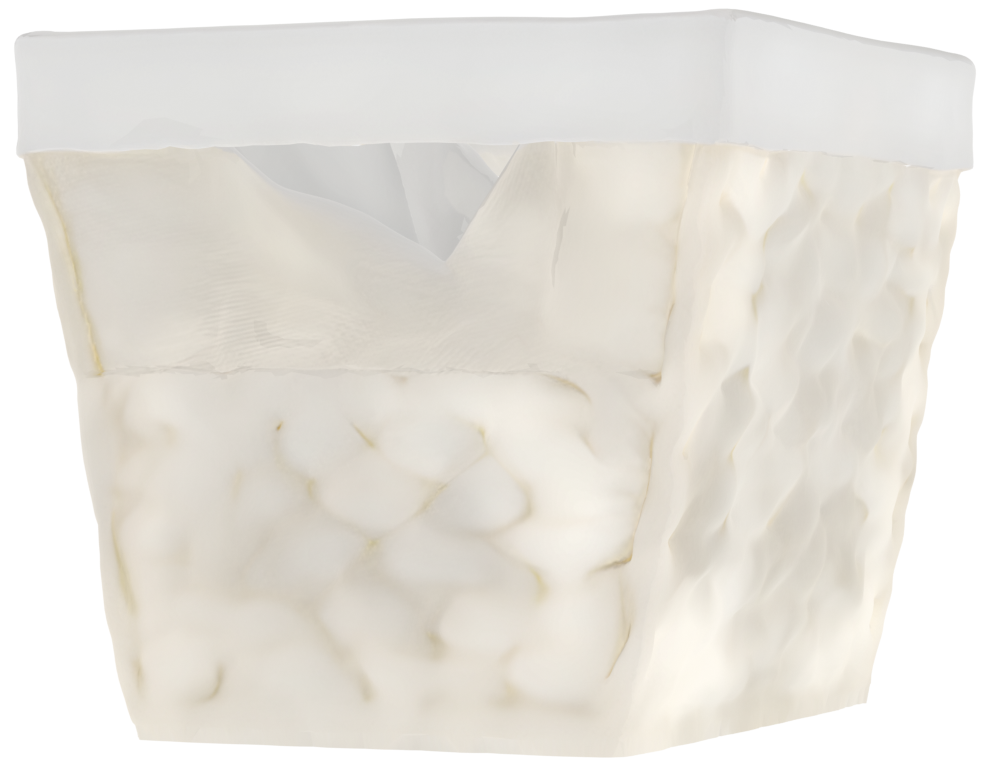}
    \hspace{0.06\linewidth}
    \objectpair{Cream cheese}{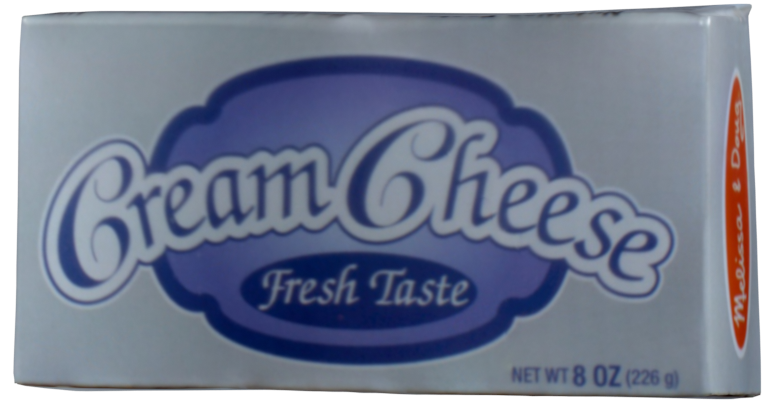}{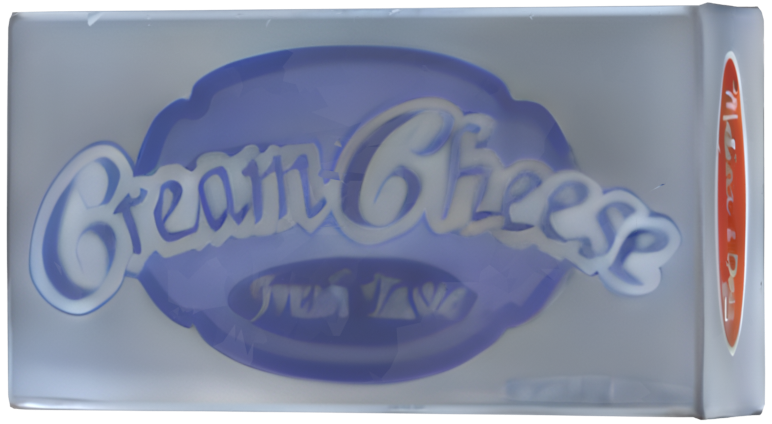}
    \hspace{0.06\linewidth}
    \objectpair{Chocolate pudding}{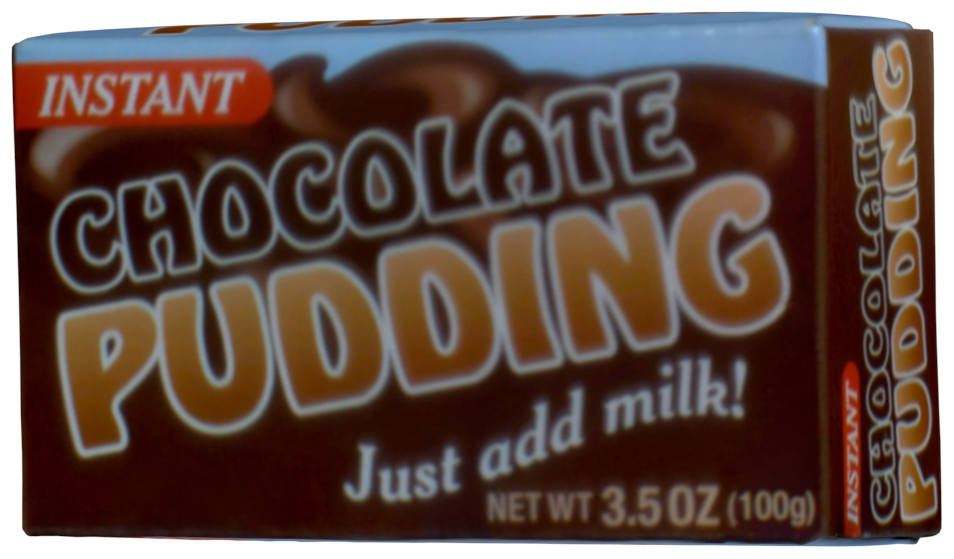}{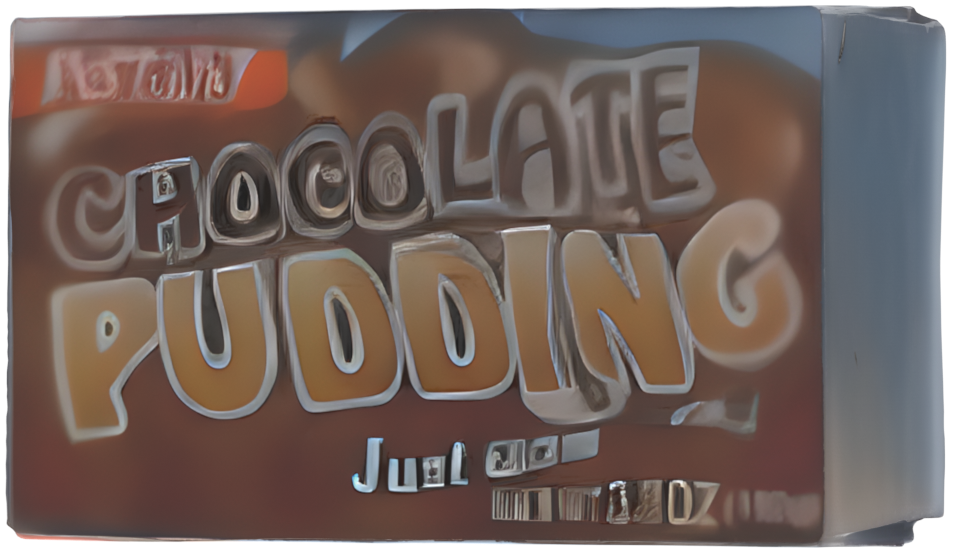}

    \caption{Comparison of LIBERO objects and their reconstructed RoboTwin 2.0 counterparts used in the \textit{Put milk in basket} task scene generation.}
    \label{fig:libero_robotwin_object_comparison}
\end{figure}

\paragraph{Scene Generation.}
To construct diverse simulation scenes, RoboTwin 2.0 generates simulation-ready object assets through Rodin-based 3D reconstruction and subsequent mesh processing. To adapt LIBERO objects for use in RoboTwin 2.0 simulation scenes, we reconstructed each object with the Rodin 3D platform~\cite{rodin} and converted it into a RoboTwin 2.0-compatible asset. For each object, the LIBERO mesh was first imported into Blender and rendered from 32 viewpoints obtained by combining four elevation angles ($-30^\circ, 0^\circ, 30^\circ, 45^\circ$) with eight uniformly spaced azimuth angles ($0^\circ, 45^\circ, 90^\circ, 135^\circ, 180^\circ, 225^\circ, 270^\circ, 315^\circ$). Five views that best captured the object's distinctive geometry and surface appearance were selected from this set and submitted to Rodin to generate a 3D mesh. Because Rodin does not recover metric scale, the reconstructed mesh was rescaled to match the LIBERO asset's bounding box dimensions along all three axes. For the milk carton in the \textit{Put milk in basket} task, Rodin reconstruction alone produced insufficient visual fidelity; the LIBERO asset's surface appearance was therefore transferred onto the Rodin geometry using Blender's texture baking functionality, overlaying the original texture onto the reconstructed mesh. Examples of the resulting RoboTwin 2.0-compatible assets for the \textit{Put milk in basket} scene are shown in Figure~\ref{fig:libero_robotwin_object_comparison}.

For each object, the initial positions are drawn uniformly from the coordinate range defined in the corresponding LIBERO BDDL file, preserving the spatial distribution of the original benchmark.

\begin{figure}[h]
    \centering

    \begin{minipage}[b]{0.22\linewidth}
        \centering
        \includegraphics[width=\linewidth]{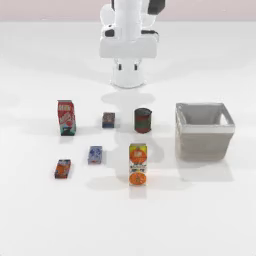}\\
        \vspace{0.3em}
        (a) Clean scene
    \end{minipage}
    \hspace{0.04\linewidth}
    \begin{minipage}[b]{0.70\linewidth}
        \centering
        \begin{minipage}[b]{0.31\linewidth}
            \centering
            \includegraphics[width=\linewidth]{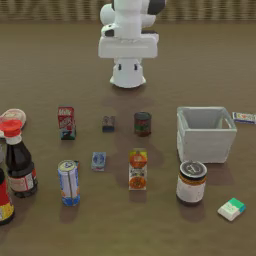}
        \end{minipage}
        \hfill
        \begin{minipage}[b]{0.31\linewidth}
            \centering
            \includegraphics[width=\linewidth]{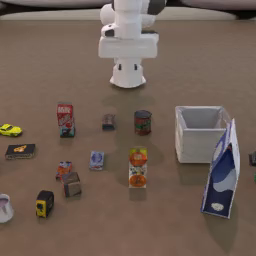}
        \end{minipage}
        \hfill
        \begin{minipage}[b]{0.31\linewidth}
            \centering
            \includegraphics[width=\linewidth]{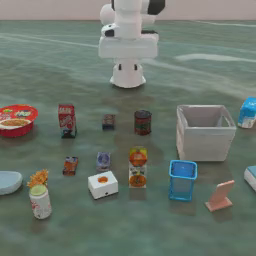}
        \end{minipage}\\
        \vspace{0.3em}
        (b) Randomized scenes
    \end{minipage}

    \caption{Examples of clean and randomized RoboTwin 2.0 scenes.}
    \label{fig:robo_clean_random_scenes}
\end{figure}

\paragraph{Domain Randomization.}
We follow the domain randomization scheme proposed in RoboTwin 2.0, applying four of the five randomization axes: background textures, lighting conditions, scene clutter, and table height. Randomization of language instruction is excluded. Background randomization varies the textures of the tabletop surface and wall. Lighting randomization perturbs light color, type, intensity, and position. Scene clutter places task-irrelevant distractor objects on the tabletop via collision-aware sampling. Table height is uniformly perturbed by up to $3\,\text{cm}$ from the reference configuration. Figure~\ref{fig:robo_clean_random_scenes} compares the clean RoboTwin 2.0 scene with three randomized variants produced by the selected randomization axes.

\paragraph{Demonstration Generation.}
RoboTwin 2.0 provides task execution code through its automated MLLM-based code generation pipeline. However, the generated task programs for our benchmark tasks exhibited insufficient success rates in our setting. We therefore used the RoboTwin 2.0 simulation and skill-library framework, but replaced the automatically generated task programs with manually implemented task-specific execution programs to ensure reliable demonstration collection. Each program encodes the manipulation logic as a sequence of high-level API calls from the RoboTwin 2.0 skill library, including grasp, place, and displacement primitives. Demonstrations were collected by rolling out these programs in the simulator until 400 successful trajectories were accumulated for each task.

\begin{figure}[h] \centering \newlength{\adaptrowh}\newlength{\adaptha}\newlength{\adapthb}\newlength{\adapthc} \settoheight{\adaptha}{\includegraphics[width=0.29\linewidth]{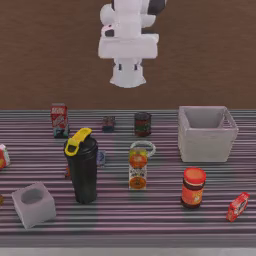}} \settoheight{\adapthb}{\includegraphics[width=0.29\linewidth]{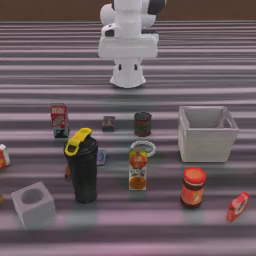}} \settoheight{\adapthc}{\includegraphics[width=0.29\linewidth]{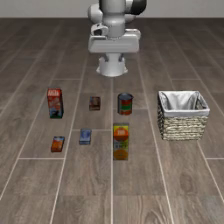}} \setlength{\adaptrowh}{\adaptha} \ifdim\adapthb>\adaptrowh\setlength{\adaptrowh}{\adapthb}\fi \ifdim\adapthc>\adaptrowh\setlength{\adaptrowh}{\adapthc}\fi \subfloat[Original scene]{ \begin{minipage}[b][\adaptrowh][c]{0.29\linewidth} \centering \includegraphics[width=\linewidth]{robo_table_origin.png} \end{minipage}} \hfill \subfloat[Adapted RoboTwin 2.0 scene]{ \begin{minipage}[b][\adaptrowh][c]{0.29\linewidth} \centering \includegraphics[width=\linewidth]{robo_table_adapted.png} \end{minipage}} \hfill \subfloat[LIBERO scene]{ \begin{minipage}[b][\adaptrowh][c]{0.29\linewidth} \centering \includegraphics[width=\linewidth]{robo_libero_eval.png} \end{minipage}} \caption{Scene-level adaptation for the \textit{Put milk in basket} task. \textbf{(a)} The original RoboTwin 2.0 tabletop scene. \textbf{(b)} The adapted scene with adjusted table dimensions. \textbf{(c)} The corresponding LIBERO evaluation scene.} \label{fig:robo_table_adaptation} \end{figure}

\paragraph{Sim-to-LIBERO Adaptation.}
Although RoboTwin 2.0 provides a tabletop simulation setup, several LIBERO tasks require scene-level adjustments to better match the corresponding LIBERO layouts. For both \textit{Put milk in basket} and \textit{Put wine bottle on cabinet}, the table dimensions (width and length) were adjusted so that the boundary between the table edge and the back wall aligned with the corresponding LIBERO scene layout, reducing the visual domain gap between the RoboTwin 2.0 training environment and the LIBERO evaluation environment. In the \textit{Put milk in basket} task, this adjustment additionally caused the scene to visually approximate the original floor-level configuration, in which the scene is initialized on the floor rather than on a tabletop surface. Figure~\ref{fig:robo_table_adaptation} visualizes this scene-level adaptation by comparing the original RoboTwin 2.0 scene, the adjusted RoboTwin 2.0 scene, and the corresponding LIBERO scene in the \textit{Put milk in basket} task.

\paragraph{Dataset Statistics.}
Table~\ref{tab:robotwin_stats} reports the success rates and collection times per task. Success rates varied considerably across tasks, ranging from 13.28\% for \textit{Stack bowls} to 100\% for \textit{Put wine bottle on cabinet}. The low success rate on \textit{Stack bowls} reflected the tight geometric constraints of the task, where precise bowl alignment is required for success. 
Collection time grew as the success rate decreased, with \textit{Stack bowls} requiring 139 minutes to accumulate 400 trajectories while \textit{Put wine bottle on cabinet} completed the same count in 13 minutes.

\begin{table*}[h]
\centering
\begin{tabular}{lcccc}
\toprule
Task & Success & Attempts & Success Rate & Time \\
\midrule
Box into basket              & 400 & 1302 & 30.72\% & 43\,min \\
Stack bowls         & 400 & 3011 & 13.28\% & 139\,min \\
Lift cup                & 400 & 1094 & 36.56\% & 26\,min \\
Put milk in basket      & 400 &  773 & 51.75\% & 35\,min \\
Put wine bottle on cabinet & 400 & 400 & 100\% & 13\,min \\
\bottomrule
\end{tabular}
\caption{Dataset generation success rates and collection times.}
\label{tab:robotwin_stats}
\end{table*}

\subsection{Pipeline Comparison}
\label{sec:supp_pipeline_comparison}

We compare PRISM, X-Sim, and RoboTwin 2.0 across three pipeline stages: scene generation, demonstration generation, and dataset construction. Although all three methods share this common structure, they differ substantially in how each stage is implemented.

\paragraph{Scene Generation.}
The three methods reflect a fundamental trade-off between target alignment and instance diversity.
X-Sim reconstructs the target workspace as a digital twin by combining Polycam-based object mesh capture and 2D Gaussian Splatting for the workspace, so the resulting scene is geometrically faithful to the deployment environment but fixed to a single instance.
RoboTwin 2.0 constructs scenes by importing task-specific objects into a generic tabletop layout that is not explicitly conditioned on the target scene.
It introduces diversity through domain randomization and instance-level asset variation, but because the layout is generic rather than reconstructed from a target image, the resulting scenes are not closely aligned with the geometry or spatial layout of the target environment.
PRISM generates a population of digital cousin scenes from a single RGB-D image of the target environment: GAIA retrieves multiple candidate assets per object from a large asset database and samples a distinct combination for each scene, so the resulting scenes share the spatial topology of the target workspace while varying in object identity.
This allows PRISM to be the only method that simultaneously preserves semantic and geometric alignment with the target environment and introduces controlled instance-level diversity.

\paragraph{Demonstration Generation.}
X-Sim extracts a reference trajectory from a single human demonstration video via FoundationPose and trains a state-based RL policy from scratch using a dense waypoint-tracking reward, then rolls out the converged policy to collect trajectories.
This approach requires per-task RL training, which incurs substantial compute overhead and limits scalability.
RoboTwin 2.0 uses a hand-crafted task program composed of skill-library API calls to drive the robot in simulation, with no learning involved; the program is replayed repeatedly under randomized conditions until 400 successful trajectories are collected.
The success rate is strongly task-dependent, ranging from 13.28\% on \textit{Stack bowls} to 100\% on \textit{Put wine bottle on cabinet}.
PRISM generates demonstrations through a VLM-TAMP pipeline that translates a natural-language task instruction into a symbolic action sequence and solves for a motion plan in each digital cousin scene.
Because the VLM-TAMP planner is task-agnostic and requires no per-task RL training, the demonstration generation stage is both scalable and consistent across tasks.

\paragraph{Dataset Construction.}
X-Sim and RoboTwin 2.0 both expand their datasets by repeating the trajectory generation procedure under domain randomization.
Consequently, each dataset sample requires an independent simulation rollout, and the total dataset size is directly proportional to the number of successful rollout attempts.
PRISM decouples dataset construction from trajectory generation by applying trajectory-preserving visual randomization: once a successful demonstration trajectory is collected, it is expanded into multiple training samples by varying visual appearance while holding the robot and object kinematics fixed.
This reuse strategy eliminates the need for repeated trajectory synthesis and reduces the per-sample generation cost significantly, as demonstrated in Section~4.5 of the main paper.

Table~\ref{tab:pipeline_comparison} summarizes the key design choices of each method.

\begin{table*}[h]
\centering
\renewcommand{\arraystretch}{1.3}
\resizebox{0.85\textwidth}{!}{%

\begin{tabular}{lccc}
\toprule
\textbf{Property} & \textbf{RoboTwin 2.0} & \textbf{X-Sim} & \textbf{PRISM (Ours)} \\
\midrule
\multicolumn{4}{l}{\textit{Scene Generation}} \\
\midrule
Single image as input              & $\times$   & $\times$   & \checkmark \\
Automatic scene alignment          & $\times$   & \checkmark & \checkmark \\
Instance-level scene diversity     & \checkmark & $\times$   & \checkmark \\
\midrule
\multicolumn{4}{l}{\textit{Demonstration Generation}} \\
\midrule
No human demonstration needed      & \checkmark & $\times$   & \checkmark \\
No per-task RL training needed     & \checkmark & $\times$   & \checkmark \\
Task-agnostic planning             & $\times$   & $\times$   & \checkmark \\
\midrule
\multicolumn{4}{l}{\textit{Dataset Construction}} \\
\midrule
Trajectory reuse (no re-rollout)      & $\times$   & $\times$   & \checkmark \\
Visual randomization                  & \checkmark & \checkmark & \checkmark \\
\bottomrule
\end{tabular}
}
\caption{Comparison of dataset generation pipeline design choices across RoboTwin 2.0, X-Sim, and PRISM.}
\label{tab:pipeline_comparison}
\end{table*}


\section{Policy Training Hyperparameters}
\label{sec:supp_implementation}

\subsection{Diffusion Policy Training Details}
\label{sec:supp_dp_training}

We followed the Diffusion Policy implementation from OpenVLA-OFT~\cite{openvla-oft} with the following configuration.
We trained a UNet-based Diffusion Policy~\cite{diffusionpolicy} using the DDIM scheduler~\cite{DDIM} for efficient inference.
The visual observations were encoded via a ResNet-18 backbone with a SpatialSoftmax pooling layer (32 keypoints, feature dimension 64), followed by color jitter and random crop augmentation during training.
Low-dimensional proprioceptive states consist of end-effector position, end-effector quaternion, and gripper joint position.
The observation encoder's BatchNorm layers were replaced with GroupNorm to ensure compatibility with Exponential Moving Average (EMA) training.

The noise prediction network is a 1D Conditional UNet with FiLM conditioning~\cite{FiLM}, where diffusion timestep embeddings and observation features are concatenated and injected at each residual block.
The UNet channel dimensions are $[256, 512, 1024]$ with a kernel size of 5 and 8 groups for GroupNorm.
The diffusion step embedding dimension is set to 256.
Actions were normalized to $[-1, 1]$ using min-max normalization prior to training.
Detailed training hyperparameters are summarized in Table~\ref{tab:supp_dp_hyperparams}.

\begin{table*}[h]
\centering
\begin{tabular}{lc}
\toprule
\textbf{Hyperparameter} & \textbf{Value} \\
\midrule
\multicolumn{2}{l}{\textit{Hardware}} \\
GPUs & 1 $\times$ NVIDIA RTX 3090 \\
Precision & float32 \\
\midrule
\multicolumn{2}{l}{\textit{Input}} \\
RGB image (\texttt{agentview\_image}) & $224 \times 224$ \\
EEF position (\texttt{robot0\_eef\_pos}) & 3 \\
EEF quaternion (\texttt{robot0\_eef\_quat}) & 4 \\
Gripper joint position (\texttt{robot0\_gripper\_qpos}) & 2 \\
\midrule
\multicolumn{2}{l}{\textit{Training}} \\
Batch size & 128 \\
Number of epochs & 1000 \\
Steps per epoch & 100 \\
Learning rate & $1 \times 10^{-4}$ \\
Weight decay (L2) & 0.0 \\
Optimizer & Adam \\
EMA power & 0.75 \\
\midrule
\multicolumn{2}{l}{\textit{Horizon}} \\
Observation horizon ($T_o$) & 2 \\
Action horizon ($T_a$) & 3 \\
Prediction horizon ($T_p$) & 8 \\
Frame stack & 2 \\
Sequence length & 7 \\
\midrule
\multicolumn{2}{l}{\textit{Diffusion (DDIM)}} \\
Training timesteps & 100 \\
Inference timesteps & 10 \\
Noise schedule & \texttt{squaredcos\_cap\_v2} \\
Prediction type & $\epsilon$-prediction \\
Clip sample & True \\
\midrule
\multicolumn{2}{l}{\textit{Visual Encoder}} \\
Backbone & ResNet-18 \\
Pooling & SpatialSoftmax (32 keypoints) \\
Feature dimension & 64 \\
Color jitter (brightness/contrast/saturation/hue) & 0.3 \\
Crop size & $202 \times 202$ \\
\midrule
\multicolumn{2}{l}{\textit{UNet Architecture}} \\
Channel dimensions & [256, 512, 1024] \\
Kernel size & 5 \\
GroupNorm groups & 8 \\
Diffusion step embedding dim & 256 \\
\bottomrule
\end{tabular}
\caption{Diffusion Policy training hyperparameters.}
\label{tab:supp_dp_hyperparams}
\end{table*}

\subsection{$\pi_{0.5}$ Training Details}
\label{sec:supp_pi05_training}
We built on $\pi_{0.5}$~\cite{pi05}, a vision-language-action model that couples a PaliGemma backbone, consisting of a SigLIP~\cite{siglip} image encoder and a Gemma-2B language model, with a lightweight Gemma-300M action expert trained using a flow-matching objective.
Starting from the public $\pi_{0.5}$-base checkpoint, we fine-tuned the model using LoRA adapters~\cite{lora} on both the backbone (rank 16) and the action expert (rank 32).
The adapters were applied to the attention and feed-forward(FFN) modules, while all pretrained weights remained frozen.
The policy receives a $224 \times 224$ third-person RGB image, a wrist-mounted camera image, an 8-D proprioceptive state containing the end-effector pose and gripper state, and a natural-language task instruction. 
It predicts a chunk of 10 future actions in a single forward pass, of which the first 8 are executed in an open-loop manner at test time.
During training, third-person observations were augmented using random cropping (scale 0.95), resizing, and random rotations within $[-5^\circ, +5^\circ]$.
In addition, all camera views were augmented using color jitter with brightness, contrast, and saturation factors of 0.3, 0.4, and 0.5, respectively.
We fine-tuned for 30{,}000 gradient steps with a batch size of 32, using the AdamW optimizer with gradient-norm clipping of 1.0 and a cosine learning-rate schedule that warms up over 10{,}000 steps to a peak of $5\times10^{-5}$.
All training hyperparameters are summarized in Table~\ref{tab:supp_pi05_hyperparams}.

\begin{table*}[h]
\centering
\begin{tabular}{lc}
\toprule
\textbf{Hyperparameter} & \textbf{Value} \\
\midrule

\multicolumn{2}{l}{\textit{Hardware}} \\
GPUs & 1 $\times$ NVIDIA RTX PRO 6000 Blackwell \\
Precision & bfloat16 \\

\midrule
\multicolumn{2}{l}{\textit{Model}} \\
Base model & $\pi_{0.5}$ \\
Vision-language backbone & PaliGemma (SigLIP encoder + Gemma-2B) \\
Action expert & Gemma-300M (flow-matching) \\
Adaptation & LoRA (backbone + action expert, base weights frozen) \\
LoRA target modules & attention + FFN \\
Backbone LoRA rank/alpha & 16 / 16 \\
Action expert LoRA rank/alpha & 32 / 32 \\

\midrule
\multicolumn{2}{l}{\textit{Training}} \\
Training steps & 30{,}000 \\
Batch size & 32 \\
Optimizer & AdamW (gradient clip 1.0) \\
Learning rate & $5\times10^{-5}$ (cosine schedule, 10k warmup) \\
EMA & disabled \\

\midrule
\multicolumn{2}{l}{\textit{Inputs}} \\
Input images (LIBERO) &
1 third-person camera image, 1 wrist-mounted camera image \\
Input images (Real) &
1 third-person camera image \\
Input image size & $224 \times 224$ \\
Proprioceptive state &
8-D: end-effector pose (3 position + 3 orientation) + 2 gripper \\
Max token length & 200 \\

\midrule
\multicolumn{2}{l}{\textit{Action Prediction}} \\
Action chunk size &
10 steps (predict 10, execute 8 open-loop at test time) \\
Action dimension & 32 \\
Action parameterization & delta (gripper absolute) \\

\midrule
\multicolumn{2}{l}{\textit{Image Augmentation}} \\
Third-person camera &
random\_crop(scale=0.95)+resize, rotate=$[-5^\circ, +5^\circ]$ \\
All cameras &
color\_jitter(brightness=0.3, contrast=0.4, saturation=0.5) \\

\bottomrule
\end{tabular}
\caption{$\pi_{0.5}$ training hyperparameters.}
\label{tab:supp_pi05_hyperparams}
\end{table*}

\section{Experimental Setup}
\label{sec:supp_task_description}

This section provides additional details regarding the robotic manipulation tasks used throughout our experiments.
We evaluated PRISM in both simulation and real-world environments to analyze dataset generation quality, policy learning capability, and generalization performance under diverse scene variations.

\begin{figure}[t]
    \centering
    \textbf{Libero Experiment Tasks}\\[0.5em]
    \begin{minipage}[b]{0.3\linewidth}
        \centering
        \includegraphics[width=\linewidth]{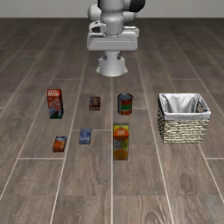}\\
        Libero Object
    \end{minipage}
    \hspace{0.05\linewidth}
    \begin{minipage}[b]{0.3\linewidth}
        \centering
        \includegraphics[width=\linewidth]{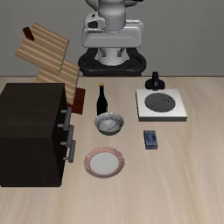}\\
        Libero Goal
    \end{minipage}
    
    \vspace{1em}
    
    \textbf{Real-World Experiment Tasks}\\[0.5em]
    \begin{minipage}[b]{0.3\linewidth}
        \centering
        \includegraphics[width=\linewidth]{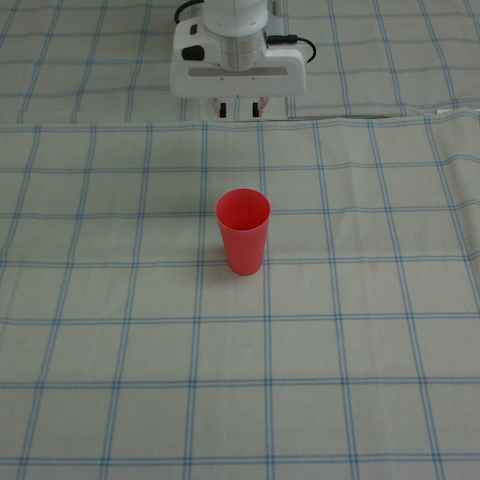}\\
        Lift cup
    \end{minipage}
    \hfill
    \begin{minipage}[b]{0.3\linewidth}
        \centering
        \includegraphics[width=\linewidth]{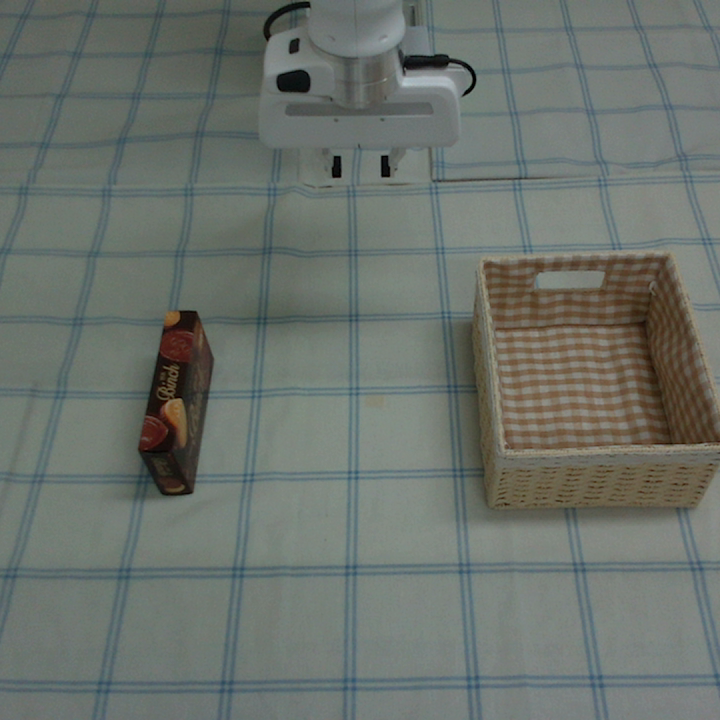}\\
        Box into basket
    \end{minipage}
    \hfill
    \begin{minipage}[b]{0.3\linewidth}
        \centering
        \includegraphics[width=\linewidth]{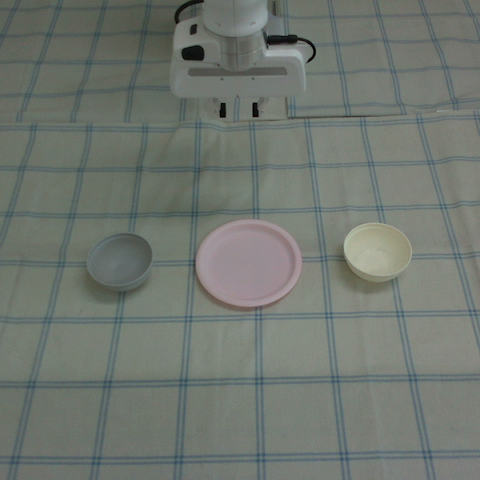}\\
        Stack bowls
    \end{minipage}
    
    \caption{Experiment tasks used in our evaluation.}
    \label{fig:libero_realworld_tasks}
\end{figure}

\subsection{Simulation Experiment}
\label{sec:supp_sim_tasks}

\paragraph{Evaluation Settings.}
In-Domain evaluates each policy in its native simulator using the same environment generation pipeline employed during dataset creation. 
Specifically, for each method, we randomly sampled 50 environments from the 400 generated environments and evaluated the trained policy within the corresponding simulator (Isaac Sim~\cite{isaacsim} for PRISM and Sapien~\cite{sapien} for X-Sim and RoboTwin 2.0).
LIBERO~\cite{libero} evaluates the same policies on a standardized manipulation benchmark in MuJoCo~\cite{mujoco}. 
LIBERO-Plus~\cite{libero-plus} further introduces variations in object layout, camera viewpoint, robot initial state, lighting, background texture, sensor noise, and language instructions to evaluate robustness under distribution shifts.
Quantitative results for all evaluation settings are reported in Table~1 of the main paper.

\paragraph{Task Details.}
For the simulation experiments, we selected one task from each of the Object and Goal task categories provided in LIBERO.
The selected tasks are summarized in Table~\ref{tab:libero_tasks}.

\begin{table*}[h]
    \centering
    \resizebox{\linewidth}{!}{
    \begin{tabular}{l l l l}
        \toprule
        \textbf{Suite} & \textbf{Task} & \textbf{Language Instruction} & \textbf{Predicate Condition} \\
        \midrule
        LIBERO-Object
        & Put milk in basket
        & ``Put the milk into the basket.''
        & $\texttt{In}(\texttt{milk}, \texttt{basket})$ \\

        LIBERO-Goal
        & Put wine bottle on cabinet
        & ``Put the wine bottle on top of the cabinet.''
        & $\texttt{On}(\texttt{wine\_bottle}, \texttt{cabinet})$ \\
        \bottomrule
    \end{tabular}
    }
    \caption{Simulation tasks used in the experiments.}
    \label{tab:libero_tasks}
\end{table*}

Figure~\ref{fig:libero_realworld_tasks} shows the simulation environments used for each task.


\paragraph{Observation.}
The observation space consists of RGB visual observations and robot proprioceptive states.
For visual observations, Diffusion Policy~\cite{dp} receives a single third-person RGB image, whereas $\pi_{0.5}$~\cite{pi05} additionally utilizes wrist-mounted camera observations to provide egocentric visual information.
All RGB observations are resized to a fixed input resolution before being passed to the policy network.

The proprioceptive state includes the end-effector position represented relative to the robot base frame and the gripper joint positions.
For orientation representation, Diffusion Policy uses quaternion-based rotations, while $\pi_{0.5}$ uses Euler angle representations.

\paragraph{Action.}
The action space is defined as continuous end-effector control commands executed in the simulator.
Specifically, each action consists of relative end-effector translation $(\Delta x, \Delta y, \Delta z)$, delta rotation represented as axis-angle increments, and a binary gripper open/close command.
The predicted actions are executed in a closed-loop manner at every environment step.


\subsection{Real-world Experiment}
\label{sec:supp_real_tasks}

\paragraph{Real-world Experimental Setup.}
The real-world experiments were conducted using a tabletop manipulation setup consisting of a robotic arm and an external RGB camera.
The external camera captures the overall task environment, providing third-person visual observations during manipulation. 
Unlike the simulation setup, no wrist-mounted camera was used in the real-world experiments.

The experimental environments contained various household objects including cups, bowls, baskets, and boxes.
During evaluation, the overall scene configuration and lighting conditions were kept largely consistent, and only the object placement was varied across episodes.
Specifically, target objects were placed at predefined yellow marker points by a human operator, so that the variation across evaluation episodes arose naturally from the human placement error around these reference points.
This setup allowed us to assess real-world robustness under realistic positional perturbations.

\begin{figure}[t]
    \centering
    \includegraphics[width=0.8\linewidth]{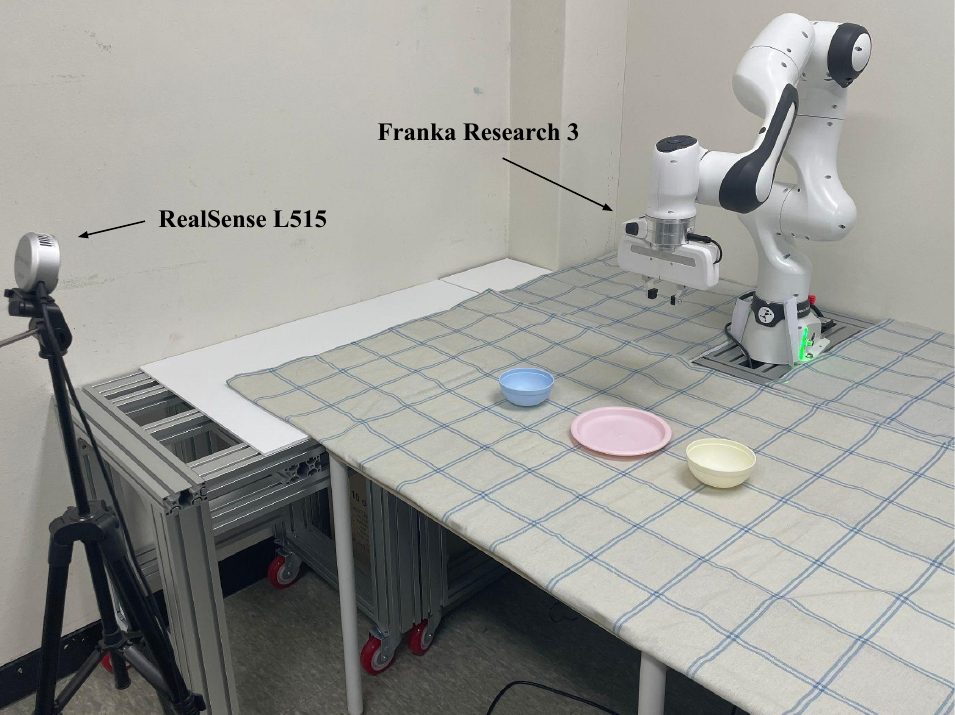}
    \caption{Real-world environments used in the experiments.}
    \label{fig:real_world_setup}
\end{figure}

{Figure~\ref{fig:real_world_setup} shows the real-world environment setup used in the experiments.}

\paragraph{Task Details.}
To evaluate real-world policy performance, we conducted experiments on three representative real-world manipulation tasks.
The selected tasks are summarized in Table~\ref{tab:real_tasks}.

\begin{table*}[h]
    \centering
    \resizebox{\linewidth}{!}{
    \begin{tabular}{l l l}
        \toprule
        \textbf{Task} & \textbf{Language Instruction} & \textbf{Success Condition} \\
        \midrule
        Lift cup
        & ``Lift the cup.''
        & The cup is lifted more than $10\,\text{cm}$ above the table surface. \\

        Box into basket
        & ``Put the box into the basket.''
        & The box is placed inside the basket. \\

        Stack bowls
        & ``Stack the scattered bowls onto the plate.''
        & Two bowls are stacked on the plate. \\
        \bottomrule
    \end{tabular}
    }
    \caption{Real-world tasks used in the experiments.}
    \label{tab:real_tasks}
\end{table*}

Figure~\ref{fig:libero_realworld_tasks} shows the real-world environments used for each task.


\paragraph{Observation.}
The observation space consists of RGB visual observations and robot proprioceptive states.
Unlike the simulation environment, all policies in the real-world experiments used only a single third-person RGB camera observation as visual input.
All RGB observations are resized to a fixed input resolution before being passed to the policy network.

The proprioceptive state includes the end-effector position represented relative to the robot base frame and the gripper joint positions.
For orientation representation, Diffusion Policy uses quaternion-based rotations, while $\pi_{0.5}$ uses Euler angle representations.

\paragraph{Action.}
The action space consists of continuous end-effector control commands executed on the real robot.
Specifically, each action consists of relative end-effector translation $(\Delta x, \Delta y, \Delta z)$, delta rotation represented using Euler angle increments, and a binary gripper open/close command.
The predicted actions are executed in a closed-loop manner at every control step.

\subsection{Effectiveness of Digital Cousin: Experimental Setup}
\label{subsec:supp_dt_vs_dc}

This subsection provides the detailed experimental setup for the digital twin versus digital cousin comparison reported in Section~4.3 (Table~2) of the main paper.

\paragraph{Compared pipelines.}
To isolate the contribution of instance-level diversity, we compared two variants of the PRISM pipeline on the \textit{Box into basket} task.
\textbf{PRISM-Twin} is a modified version of the pipeline that reconstructs a single digital twin scene that closely mirrors the target environment, so the generated dataset contains no instance-level variation across scenes.
\textbf{PRISM-Cousin} is the unmodified pipeline, which reconstructs digital cousin scenes by retrieving visually and geometrically similar assets, yielding instance-level diversity in objects and layout across scenes.
Both pipelines used identical settings for everything except scene construction, and both produced datasets of the same size, so any performance difference could be attributed to the twin-versus-cousin scene design rather than to dataset scale.
We fine-tuned $\pi_{0.5}$ separately on each dataset and evaluated the resulting policies in the real world.

\begin{figure}[t]
    \centering
    \includegraphics[width=0.85\linewidth]{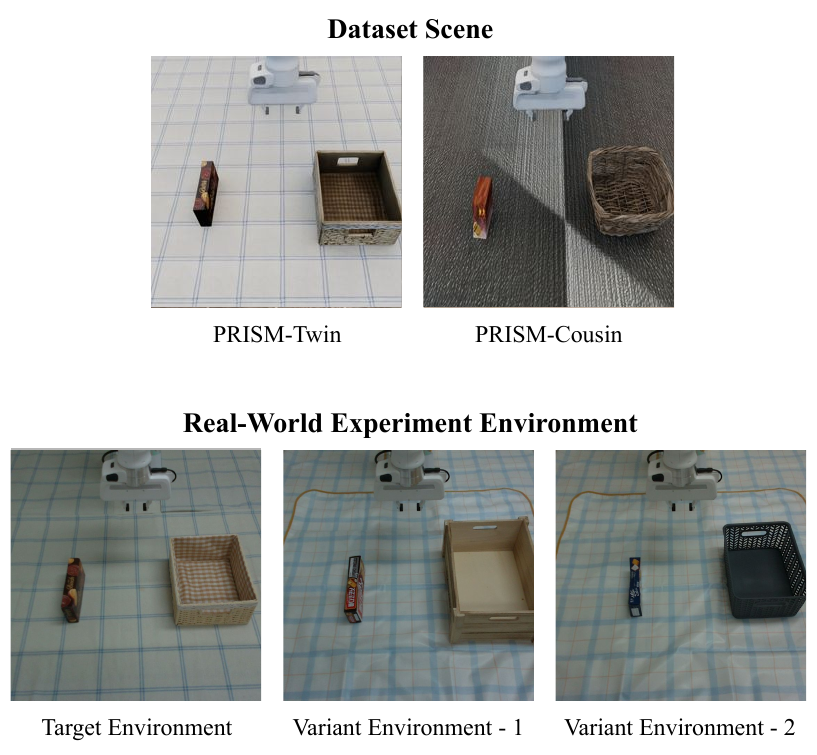}
    \caption{Experimental setup for the digital twin versus digital cousin comparison. \textbf{Top (Dataset Scene):} representative scenes from which the two datasets are generated; PRISM-Twin uses a single digital twin scene of the target environment, whereas PRISM-Cousin uses a diverse digital cousin scene with different objects and background. \textbf{Bottom (Real-World Experiment Environment):} the three real-world environments used for evaluation; the target environment matches the reconstructed scene, while the two variant environments change the objects and background appearance.}
    \label{fig:dt_vs_dc_setup}
\end{figure}

\paragraph{Datasets.}
The top row of Figure~\ref{fig:dt_vs_dc_setup} shows the scenes from which the two datasets are generated.
The PRISM-Twin dataset is generated entirely from the single digital twin scene, so its samples share a nearly identical appearance and object configuration.
The PRISM-Cousin dataset is generated from digital cousin scenes and therefore exhibits diversity in object instances and background appearance across demonstrations.

\paragraph{Evaluation environments.}
We evaluated both policies in three real-world environments, shown in the bottom row of Figure~\ref{fig:dt_vs_dc_setup}.
The \emph{target environment} matches the scene that the digital twin was reconstructed from.
The two \emph{variant environments} replace the objects and background appearance with previously unseen instances to probe generalization beyond the reconstructed scene.

\paragraph{Evaluation protocol.}
For the target environment we ran $10$ trials.
For each of the two variant environments we ran $5$ trials, giving $10$ trials in the variant setting in total.
A trial was counted as a success when the box was placed inside the basket.
The success rates obtained under this protocol are reported in Table~2 of the main paper.

\section{Extended Experimental Results \& Analysis}
\label{sec:supp_extended_results}

\subsection{Data Generation Efficiency}
\label{sec:supp_generation_efficiency}

Because PRISM generates all demonstrations directly in simulation, the dataset is produced without any human teleoperation, and the only real cost is the computation time of the pipeline. We report this time separately for each task, following the procedure described in Section~\ref{sec:supp_prism_pipeline}. PRISM produces a dataset in three phases: (i)~\emph{scene generation}, which reconstructs $4$ digital cousin scenes from a single RGB-D image of the workspace; (ii)~\emph{demonstration generation}, in which a VLM-TAMP solver synthesizes task-completing trajectories on each scene; and (iii)~\emph{dataset construction}, which expands each trajectory into many demonstrations by replaying it under randomized appearance. For every task we keep $10$ trajectories per scene and replay each one $10$ times, producing $400$ demonstrations per task. Scene generation produces all $4$ scenes in a single run, whereas demonstration generation and dataset construction process the $4$ scenes in parallel (one process per scene); all reported times are actual elapsed times. We describe each phase in turn, using the \textit{Put milk in basket} task as a running example, and Table~\ref{tab:supp_generation_summary} at the end of this subsection summarizes the per-task cost across all tasks.

\paragraph{Phase 1: Scene generation.}
From a single RGB-D image of the workspace, PRISM reconstructs $4$ digital cousin scenes in three internal steps: (1)~\emph{real-world extraction}, which detects and segments the objects in the input image and recovers their geometry; (2)~\emph{digital cousin matching}, which retrieves, for each detected object, the most similar assets from a CAD database; and (3)~\emph{scene composition}, which places the retrieved assets into $4$ randomized scene layouts. All $4$ scenes are produced in a single run. Table~\ref{tab:supp_scene_generation} reports the time of each step for every task. The cost was dominated by digital cousin matching and ranged from a few minutes for the real-world tasks to about $1.2$ hours for the wine task, where many candidate assets must be compared.

\begin{table*}[h]
\centering
\resizebox{0.9\textwidth}{!}{%
\begin{tabular}{l c c c c}
\toprule
\textbf{Task} & \textbf{Step 1: Extraction} & \textbf{Step 2: Cousin matching} & \textbf{Step 3: Composition} & \textbf{Total} \\
\midrule
\multicolumn{5}{l}{\textit{Simulation tasks}} \\
Put milk in basket          & $\sim 3$ min   & $\sim 15$ min   & $\sim 3$ min   & $\sim 21$ min \\
Put wine bottle on cabinet  & $\sim 3.6$ min & $\sim 66$ min   & $\sim 4.1$ min & $\sim 74$ min \\
\midrule
\multicolumn{5}{l}{\textit{Real-world tasks}} \\
Lift cup            & $\sim 0.4$ min & $\sim 1.8$ min & $\sim 2.2$ min & $\sim 4.4$ min \\
Box into basket     & $\sim 0.4$ min & $\sim 3.1$ min & $\sim 2.4$ min & $\sim 5.9$ min \\
Stack bowls         & $\sim 0.6$ min & $\sim 5.6$ min & $\sim 3.0$ min & $\sim 9.2$ min \\
\bottomrule
\end{tabular}%
}
\caption{Scene-generation time per task, broken into the three internal steps (real-world extraction, digital cousin matching, scene composition). All $4$ digital cousin scenes are produced in a single run, so these times are not multiplied by the number of scenes.}
\label{tab:supp_scene_generation}
\end{table*}

\paragraph{Phase 2: Demonstration generation.}
For each scene, the VLM-TAMP solver attempts to generate trajectories and keeps the $10$ shortest task-completing ones, yielding $40$ kept trajectories across the $4$ scenes. Each attempt took about $90$ seconds on average. For the milk task, every attempt succeeded ($100\%$ success rate); for harder tasks some attempts failed to find a feasible plan, so the success rate was lower, while at least $10$ trajectories per scene still succeeded and were kept. Because the $4$ scenes ran at the same time, this phase finished in about $30$ minutes. Table~\ref{tab:supp_demo_generation} reports the per-task success rate and time.

\begin{table*}[h]
\centering
\resizebox{0.9\textwidth}{!}{%
\begin{tabular}{l c c c}
\toprule
\textbf{Task} & \textbf{Success rate} & \textbf{Avg.\ time / attempt} & \textbf{Total time} \\
\midrule
\multicolumn{4}{l}{\textit{Simulation tasks}} \\
Put milk in basket          & $100\%$  & $\sim 90$ s  & $\sim 30$ min \\
Put wine bottle on cabinet  & $92.3\%$ & $\sim 190$ s & $\sim 63$ min \\
\midrule
\multicolumn{4}{l}{\textit{Real-world tasks}} \\
Lift cup            & $100\%$  & $\sim 74$ s  & $\sim 25$ min \\
Box into basket     & $97.5\%$ & $\sim 133$ s & $\sim 49$ min \\
Stack bowls         & $77.5\%$ & $\sim 182$ s & $\sim 73$ min \\
\bottomrule
\end{tabular}%
}
\caption{Demonstration-generation time per task. For each task, the VLM-TAMP solver runs on the $4$ scenes in parallel and keeps the $10$ shortest task-completing trajectories per scene ($40$ in total). ``Success rate'' is the fraction of attempts that produce a task-completing trajectory, and ``Total time'' is the parallel elapsed time (the slowest of the $4$ processes).}
\label{tab:supp_demo_generation}
\end{table*}

\paragraph{Phase 3: Dataset construction.}
Each of the $40$ kept trajectories is then replayed $10$ times while only the visual appearance (e.g., background texture) is changed; the robot motion itself stays the same. This produces $100$ demonstrations per scene and $400$ demonstrations in total. Each replay took about $107$ seconds on average, and all $400$ replays succeeded (no failed demonstrations). Since this phase reuses trajectories that were already produced in the demonstration-generation phase, no new planning is needed. With the $4$ scenes running at the same time, this phase took about $3$ hours. Table~\ref{tab:supp_dataset_construction} reports the per-task time.

\begin{table*}[h]
\centering
\resizebox{0.9\textwidth}{!}{%
\begin{tabular}{l c c c}
\toprule
\textbf{Task} & \textbf{Demonstrations} & \textbf{Avg.\ time / demo} & \textbf{Total time} \\
\midrule
\multicolumn{4}{l}{\textit{Simulation tasks}} \\
Put milk in basket          & $400$ & $\sim 107$ s & $\sim 3.1$ h \\
Put wine bottle on cabinet  & $400$ & $\sim 69$ s  & $\sim 2.2$ h \\
\midrule
\multicolumn{4}{l}{\textit{Real-world tasks}} \\
Lift cup            & $400$ & $\sim 42$ s  & $\sim 1.2$ h \\
Box into basket     & $400$ & $\sim 114$ s & $\sim 3.3$ h \\
Stack bowls         & $400$ & $\sim 180$ s & $\sim 5.3$ h \\
\bottomrule
\end{tabular}%
}
\caption{Dataset-construction time per task. Each of the $40$ kept trajectories is replayed $10$ times under randomized appearance, yielding $400$ demonstrations; all replays succeed ($100\%$, no failed demonstrations). The $4$ scenes are processed in parallel ($100$ demonstrations each), and ``Total time'' is the parallel elapsed time (the slowest of the $4$ processes).}
\label{tab:supp_dataset_construction}
\end{table*}

\paragraph{Total time.}
Summing the three phases, generating the full set of $400$ demonstrations for the milk task took about $3.9$ hours ($21$ minutes of scene generation, $30$ minutes of demonstration generation, and $3$ hours of dataset construction), which was roughly $35$ seconds per demonstration on average. Across all tasks the total ranged from about $1.7$ hours (\textit{Lift cup}) to $6.6$ hours (\textit{Stack bowls}), as summarized in Table~\ref{tab:supp_generation_summary}. The whole process runs automatically and needs no human supervision, so the total time can be further reduced simply by using more parallel processes.

\begin{table*}[h]
\centering
\resizebox{0.99\textwidth}{!}{%
\begin{tabular}{l c c c c c}
\toprule
\textbf{Task} & \textbf{Success rate} & \textbf{Scene gen.\ (min)} & \textbf{Demo.\ gen.\ (h)} & \textbf{Dataset constr.\ (h)} & \textbf{Total (h)} \\
\midrule
\multicolumn{6}{l}{\textit{Simulation tasks}} \\
Put milk in basket          & $100\%$  & $\sim 21$ & $\sim 0.5$ & $\sim 3.1$ & $\sim 3.9$ \\
Put wine bottle on cabinet  & $92.3\%$ & $\sim 74$ & $\sim 1.1$ & $\sim 2.2$ & $\sim 4.5$ \\
\midrule
\multicolumn{6}{l}{\textit{Real-world tasks}} \\
Lift cup            & $100\%$  & $\sim 4$ & $\sim 0.4$ & $\sim 1.2$ & $\sim 1.7$ \\
Box into basket     & $97.5\%$ & $\sim 6$ & $\sim 0.8$ & $\sim 3.3$ & $\sim 4.2$ \\
Stack bowls         & $77.5\%$ & $\sim 9$ & $\sim 1.2$ & $\sim 5.3$ & $\sim 6.6$ \\
\midrule
\textbf{All $5$ tasks} & $\mathbf{93.5\%}$ & $\boldsymbol{\sim 114}$ & $\boldsymbol{\sim 4.0}$ & $\boldsymbol{\sim 15.1}$ & $\boldsymbol{\sim 21}$ \\
\bottomrule
\end{tabular}%
}
\caption{Per-task data generation cost across the three phases of the PRISM pipeline; each task yields $400$ demonstrations. ``Success rate'' is the demonstration-generation (VLM-TAMP) success rate, i.e.\ the fraction of attempts that produce a task-completing trajectory. Scene generation produces all $4$ scenes in a single run, while demonstration generation and dataset construction process the $4$ scenes in parallel. ``Total'' sums the three phases. The bottom row aggregates over all five tasks; when the tasks are generated concurrently on separate machines, the wall-clock is bounded by the slowest task ($\sim\!6.6$ hours). The entire process is automatic and requires no human teleoperation.}
\label{tab:supp_generation_summary}
\vspace{-10pt}
\end{table*}

\paragraph{End-to-end cost.}
Overall, PRISM turns a single RGB-D image into $400$ task demonstrations in roughly $2$ to $7$ hours of fully autonomous computation per task (Table~\ref{tab:supp_generation_summary}), with no human teleoperation at any stage. The cost is dominated by dataset construction, which simply replays already-solved trajectories and is therefore embarrassingly parallel; since every phase scales across scenes and processes, the end-to-end time can be reduced further given additional compute, making large-scale dataset generation practical. Aggregated over all five tasks (bottom row of Table~\ref{tab:supp_generation_summary}), PRISM produced $2{,}000$ demonstrations at an overall success rate of $93.5\%$, for a combined $\sim\!21$ hours of compute that dropped to about $6.6$ hours when the tasks were generated concurrently on separate machines.

\vspace{-5pt}
\subsection{LIBERO Generalization}
\label{sec:supp_libero_generalization}
\vspace{-5pt}

This section provides detailed results for the LIBERO-Plus generalization experiments reported in Table 1 of the main paper. We evaluated the generalization capability of our method using a subset of the perturbation settings from LIBERO-Plus~\cite{libero-plus}, a systematic robustness benchmark for Vision-Language-Action models. LIBERO-Plus extends the LIBERO benchmark~\cite{libero} by introducing controlled perturbations across seven dimensions including object layout, camera viewpoints, robot initial states, language instructions, lighting conditions, background textures, and sensor noise to expose generalization weaknesses that are otherwise hidden under standard evaluation protocols.

\begin{figure}[h]
\centering
\includegraphics[width=1.0\textwidth]{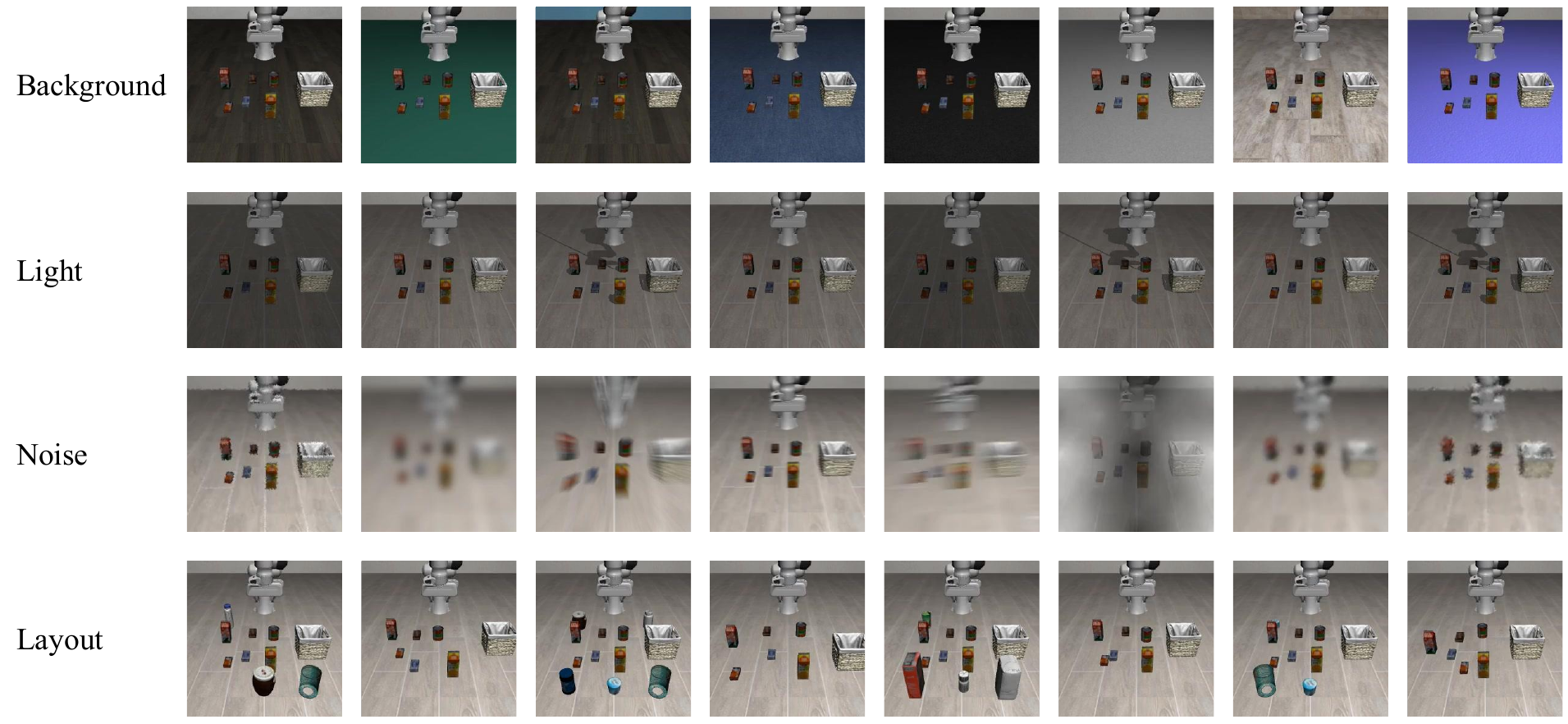}
\caption{Visualization of the test scenes for the \textit{Put milk in basket} task used in the LIBERO-Plus generalization experiments. We show representative perturbations across the four perturbation dimensions: \textbf{Light} (variations in illumination intensity, direction, and color), \textbf{Background} (changes in table and scene textures), \textbf{Noise} (photometric distortions such as motion blur, Gaussian blur, and fog), and \textbf{Layout} (addition of confounding objects and displacement of target objects). These perturbations expose generalization weaknesses that remain hidden under the standard evaluation protocol.}
\label{fig:supp_libero_plus_test_scenes}
\end{figure}

In our experiments, we focused on four perturbation dimensions, and each policy was evaluated on two tasks: \textit{Put milk in basket} and \textit{Put wine bottle on cabinet}, across three data settings: RoboTwin 2.0, X-Sim, and Ours (PRISM-generated data).
\begin{itemize}[itemsep=4pt, topsep=0pt, parsep=0pt, partopsep=0pt]
    \item \textbf{Light}: variations in illumination intensity, direction, and color.
    \item \textbf{Background}: changes in table and scene textures.
    \item \textbf{Noise}: photometric distortions such as motion blur, Gaussian blur, and fog.
    \item \textbf{Layout}: addition of confounding objects and displacement of target objects.
\end{itemize}
Figure~\ref{fig:supp_libero_plus_test_scenes} illustrates representative test scenes for the \textit{Put milk in basket} task across the four perturbation dimensions, and Table~\ref{tab:supp_libero_generalization} reports the success rates under each perturbation dimension, along with the overall average.

\begin{table*}[h]
\centering
\resizebox{0.8\textwidth}{!}{%
\begin{tabular}{llccccc}
\toprule
\textbf{Model} & \textbf{Data} & \textbf{Light} & \textbf{Background} & \textbf{Noise} & \textbf{Layout} & \textbf{Total} \\
\midrule
\multicolumn{7}{l}{\textit{Put milk in basket}} \\
\midrule
\multirow{3}{*}{$\pi_{0.5}$}
    & RoboTwin 2.0&  24.00 & 21.88 & 12.82 & 28.89 & 21.90 \\
    & X-Sim    &  68.00 & 43.80 &  2.60 & 28.90 & 35.83 \\
    & Ours     &  96.00 & 90.60 & 28.20 & 55.60 & 67.60 \\
    \midrule
    \multirow{3}{*}{\shortstack[l]{Diffusion\\Policy}}
    & RoboTwin 2.0&   84.00 & 50.00 &  0.00 &  0.00 &  33.50 \\
    & X-Sim    &   8.00 &  3.12 &  0.00 &  0.00 &  2.78 \\
    & Ours     &  92.00 & 40.62 &  5.13 &  4.44 & 35.55 \\
    \midrule
    \multicolumn{7}{l}{\textit{Put wine bottle on cabinet}} \\
    \midrule
    \multirow{3}{*}{$\pi_{0.5}$}
    & RoboTwin 2.0&   3.57 &  0.00 &  2.63 &  6.80 &  3.25 \\
    & X-Sim    &  50.00 & 34.15 & 78.95 & 54.54 & 54.41 \\
    & Ours     &  85.70 & 61.00 &  0.00 & 61.40 & 52.03 \\
    \midrule
    \multirow{3}{*}{\shortstack[l]{Diffusion\\Policy}}
    & RoboTwin 2.0&  25.00 &  4.90 & 13.16 & 65.91 &  27.24 \\
    & X-Sim    &   0.00 &  0.00 &  0.00 &  2.30 &  0.58 \\
    & Ours     & 17.90  & 53.70 & 23.70 & 40.90 & 34.05$^{*}$ \\
\bottomrule
\end{tabular}%
}
\vspace{2pt}
\begin{minipage}{0.8\textwidth}
\footnotesize{$^*$ This value differs from the main paper (28.8) due to a calculation error. The value reported here is correct.}
\end{minipage}
\caption{LIBERO-Plus generalization results across perturbation dimensions. Success rates (\%) are reported for each data setting and policy. RoboTwin 2.0 and X-Sim denote existing data sources; Ours denotes PRISM-generated data.}
\label{tab:supp_libero_generalization}
\end{table*}

\subsection{Generalization in Real-world}
\label{sec:supp_real_generalization}


\begin{wraptable}{r}{0.5\textwidth}
    \centering
    \renewcommand{\arraystretch}{1.2}
    \begin{tabular}{lccc}
    \toprule
          & Lift cup & Box into basket & Stack bowls \\ \hline
    PRISM & 100.0 & 80.0 & 50.0 \\ \bottomrule
    \end{tabular}
    \caption{Task success rate of PRISM in variant environments across all three real-world tasks.}
    \label{tab:supp_real_generalization}
    \vspace{-5pt}
\end{wraptable}

While Section~4.1 of the main paper evaluates generalization in simulation through LIBERO-Plus, and Section~4.3 of the main paper provides a limited generalization evaluation in the real world restricted to the \textit{Box into basket} task, a comprehensive assessment of real-world generalization across all tasks has not been presented.
To address this, we extended the variant environment evaluation to the remaining two real-world tasks, \textit{Lift cup} and \textit{Stack bowls}, following the same protocol as Section~4.3 of the main paper.
For each task, we fine-tuned $\pi_{0.5}$ on PRISM-generated datasets and evaluated the resulting policy over 10 trials in a variant environment with different objects and background appearances from those seen during training.
As shown in Table~\ref{tab:supp_real_generalization}, PRISM maintained consistent performance across all three real-world tasks under variant conditions, demonstrating that the instance-level diversity introduced by digital cousin scenes supports generalization not only in simulation but also across a broad range of real-world manipulation tasks.

\subsection{Failure Case Taxonomy}
\label{sec:supp_failure_case}

\begin{figure}[h]\centering
\subfloat[Misaligned object 1]{\includegraphics[width=0.32\linewidth]{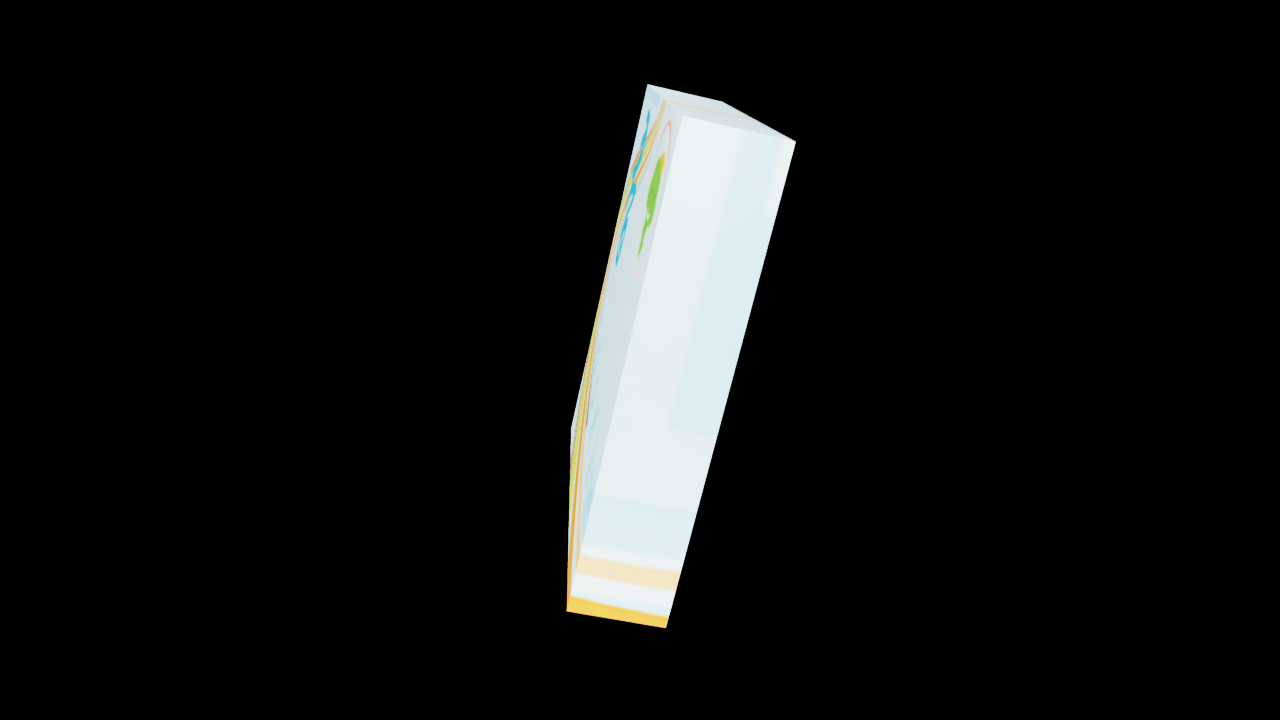}}
\hfill
\subfloat[Scene with the misaligned object]{\includegraphics[width=0.32\linewidth]{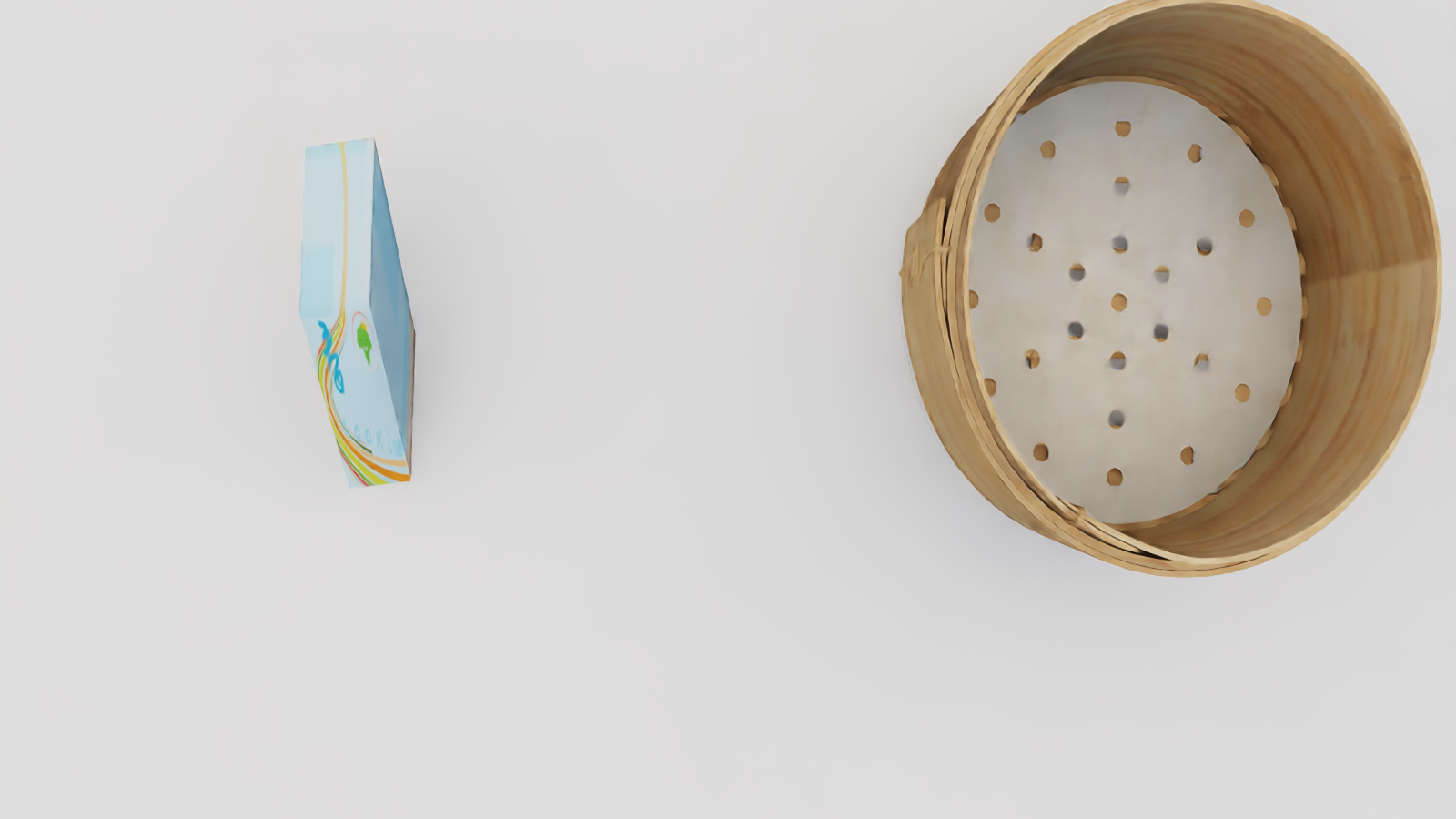}}
\hfill
\subfloat[Scene with the misaligned object]{\includegraphics[width=0.32\linewidth]{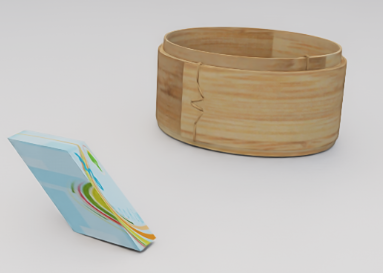}}
\\
\subfloat[Misaligned object 2]{\includegraphics[width=0.32\linewidth]{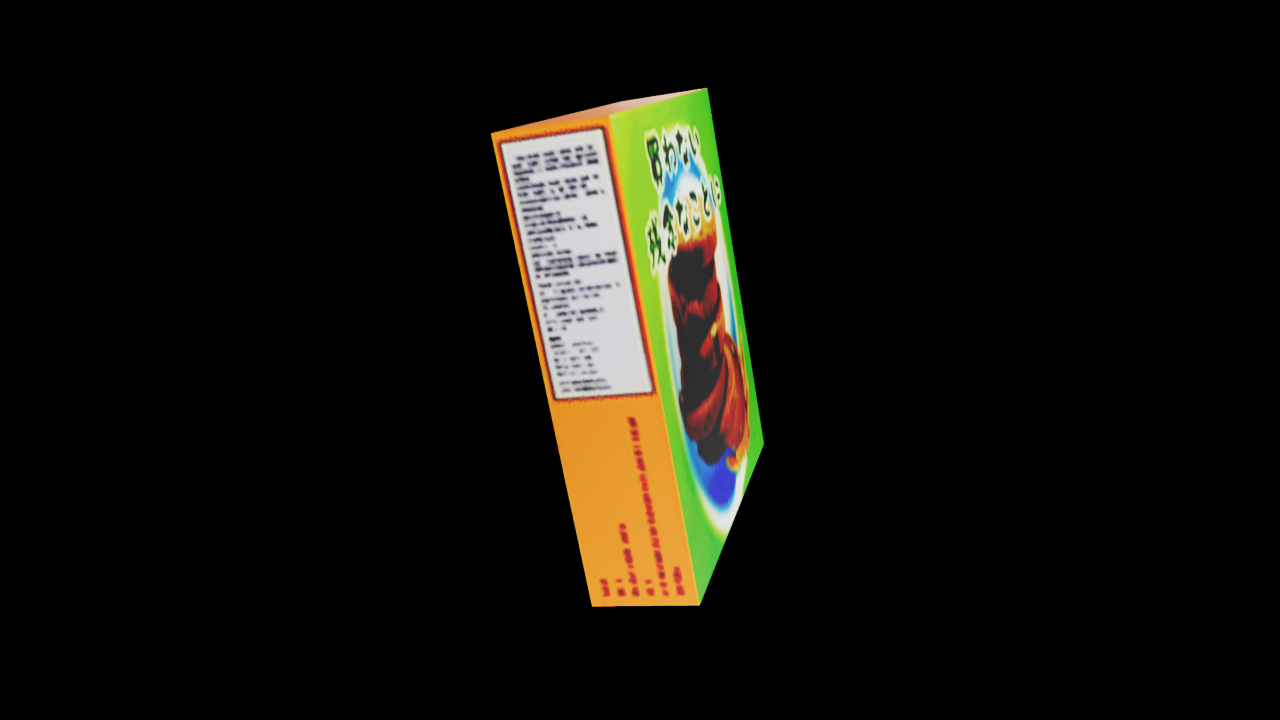}}
\hfill
\subfloat[Scene with the misaligned object2]{\includegraphics[width=0.32\linewidth]{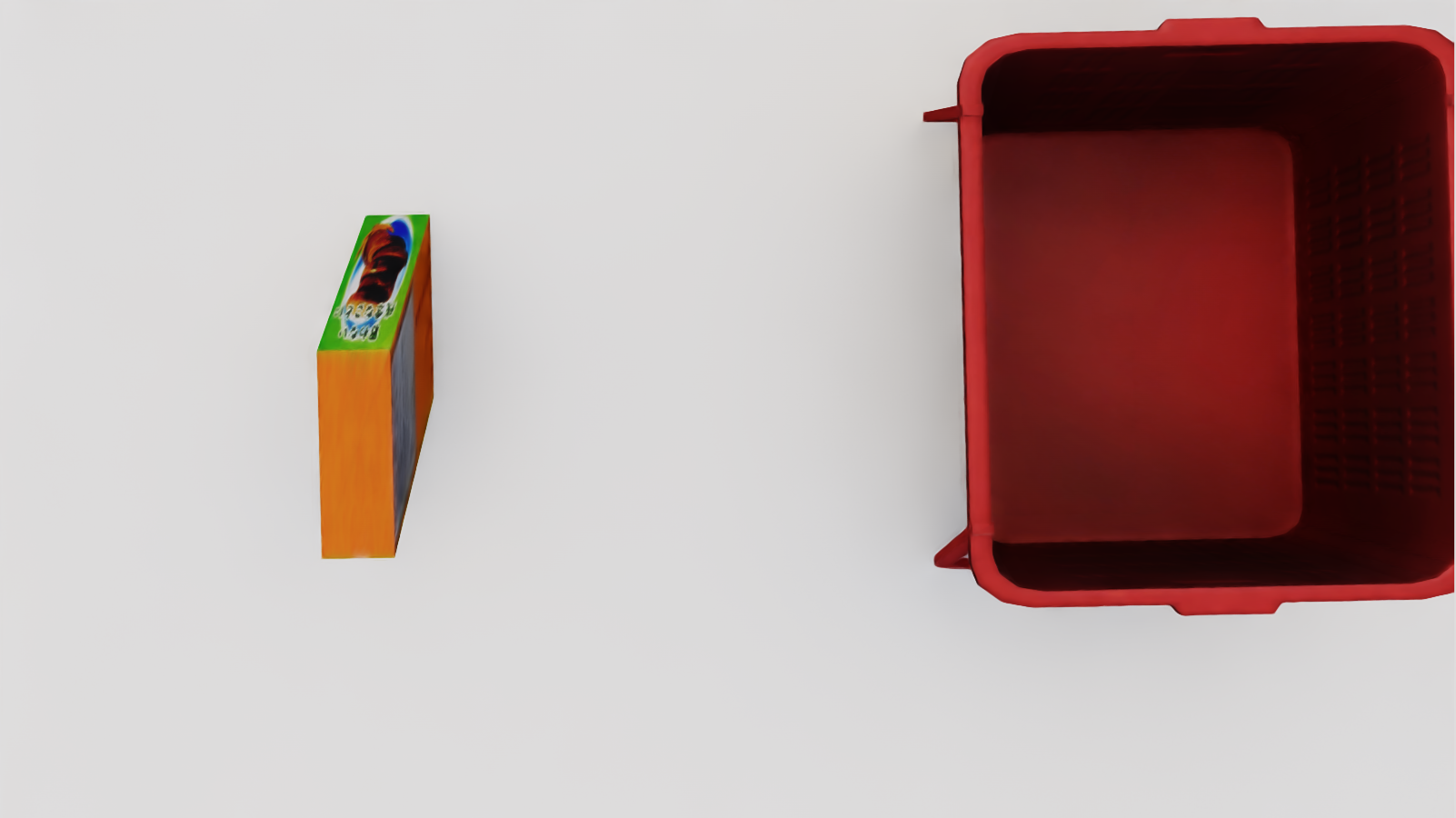}}
\hfill
\subfloat[Scene with the misaligned object2]{\includegraphics[width=0.32\linewidth]{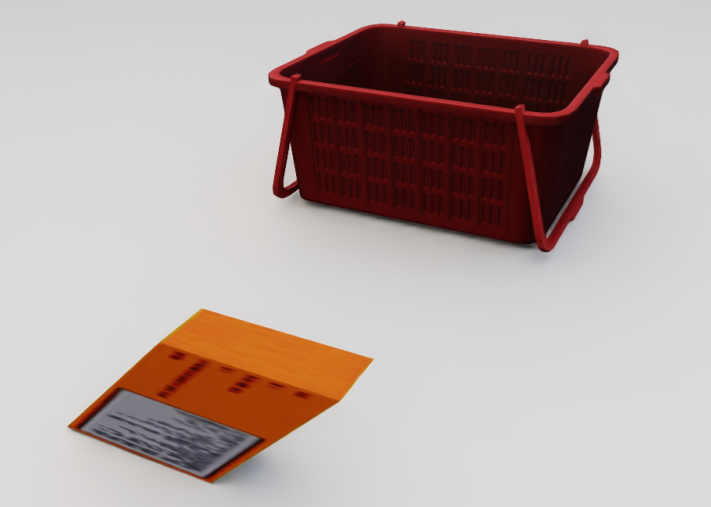}}
\caption{\textbf{(a)} Misaligned object 1. \textbf{(b)}, \textbf{(c)} Simulation scene with misaligned object. \textbf{(d)} Misaligned object 2. \textbf{(e)},\textbf{(f)} Simulation scene with misaligned object2.}
\label{fig:failure_misaligned_object}
\end{figure}

\paragraph{Scene Generation.}
When generating a scene from an image, if occlusion occurs on an object, the occluded portion renders the point cloud of that object inaccurate, which can in turn place the object in an inappropriate state when it is rescaled.
Additionally, if the captured image does not provide a view sufficient to obtain the 3D bounding box, for example when only one face of the object is visible, the point cloud of the object likewise cannot be properly reconstructed, so the object fails to be placed correctly in the scene.
Moreover, when an asset in the asset library is itself not well initialized to an upright orientation, it is imported as-is and may therefore be spawned into the simulation in a misaligned state, as illustrated in Figure~\ref{fig:failure_misaligned_object}.

\paragraph{Demonstration Generation.}
When generating a trajectory for a task, the VLM must convert the task instruction into an action sequence.
Because the bounding boxes of objects are drawn on the camera image and the action sequence is constructed based on them, an occluded object, particularly an object directly relevant to the task, can lead to an incorrect action sequence, causing the demonstration trajectory generation to fail.

\section{Example Prompts}
\label{sec:supp_generation_prompts}

PRISM queries a vision-language model (VLM) in two phases, described below in pipeline order: digital cousin \emph{scene generation} (Section~\ref{sec:supp_scene_generation}) and \emph{demonstration generation} (Section~\ref{sec:supp_demo_generation}).
In all prompts, fields enclosed in \texttt{\{\,\}} are instantiated per object and per scene; all other text is fixed, and each query is decoded at a low temperature ($0$-$0.1$) for reproducibility.

\subsection{Scene Generation Prompt}
\label{sec:supp_scene_generation}

Scene generation uses the GAIA digital cousin pipeline, which turns a single RGB-D image into a simulated scene in three steps.
\textbf{Step~1 (Real-World Extraction)} parses the image into a scene graph of objects, their articulation, and how each is supported (floor vs.\ wall).
\textbf{Step~2 (Digital Cousin Matching)} selects, per object, the simulator asset and orientation that best match the observation.
\textbf{Step~3 (Simulated Scene Generation)} composes the assets into a physically consistent scene; it is fully procedural and issues no VLM query, so the prompts below cover only Steps~1 and~2.
Besides the instruction text reproduced here, each call also receives the relevant images (the scene, the target's bounding box and mask, and the candidate wall masks or asset snapshots).

\begin{figure}[h]
    \centering
    \includegraphics[width=1.0\linewidth]{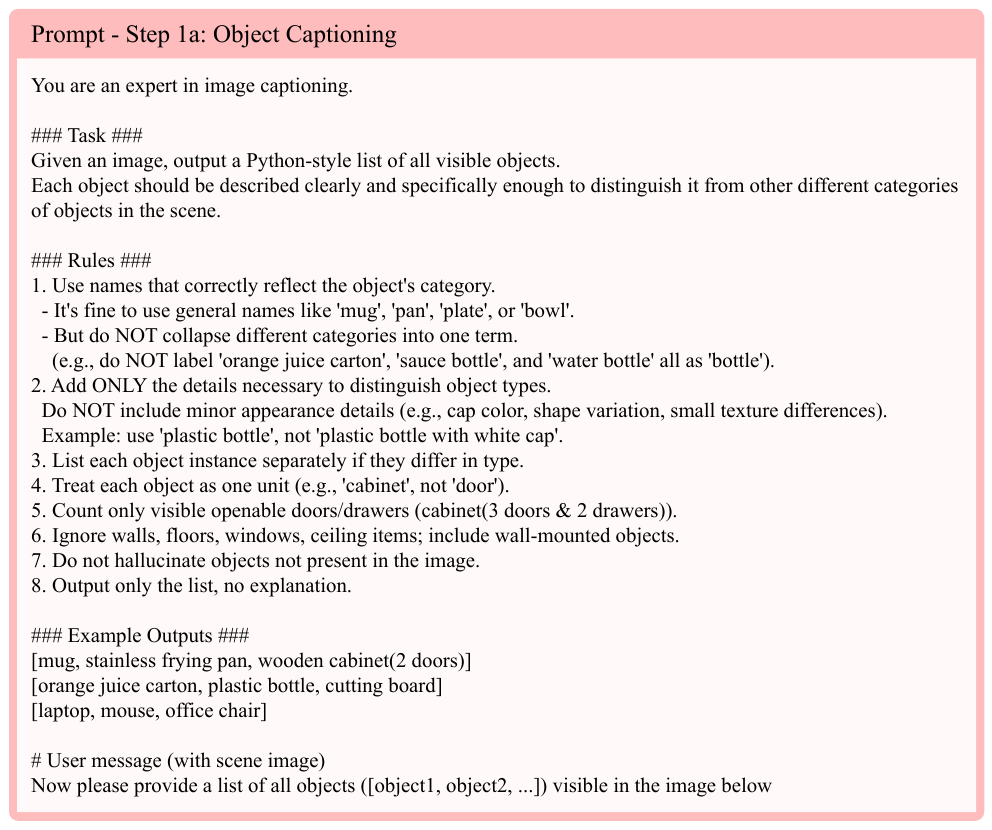}
    \caption{\textbf{Step~1a: Object captioning.} The VLM enumerates the objects present in the workspace image and produces a short caption for each.}
    \label{fig:prompt_step1a}
\end{figure}

\begin{figure}[h]
    \centering
    \includegraphics[width=1.0\linewidth]{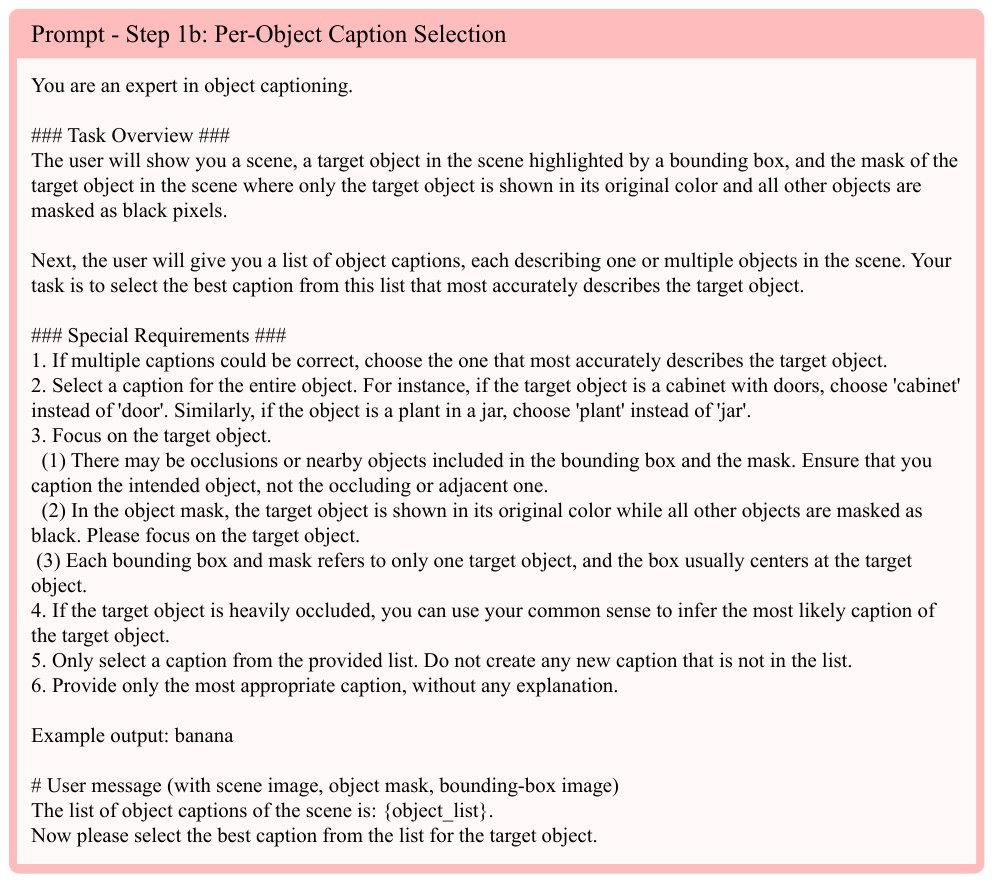}
    \caption{\textbf{Step~1b: Per-object caption selection.} Given several candidate captions for a single segmented object, the VLM selects the one that best matches its appearance.}
    \label{fig:prompt_step1b}
\end{figure}

\begin{figure}[h]
    \centering
    \includegraphics[width=1.0\linewidth]{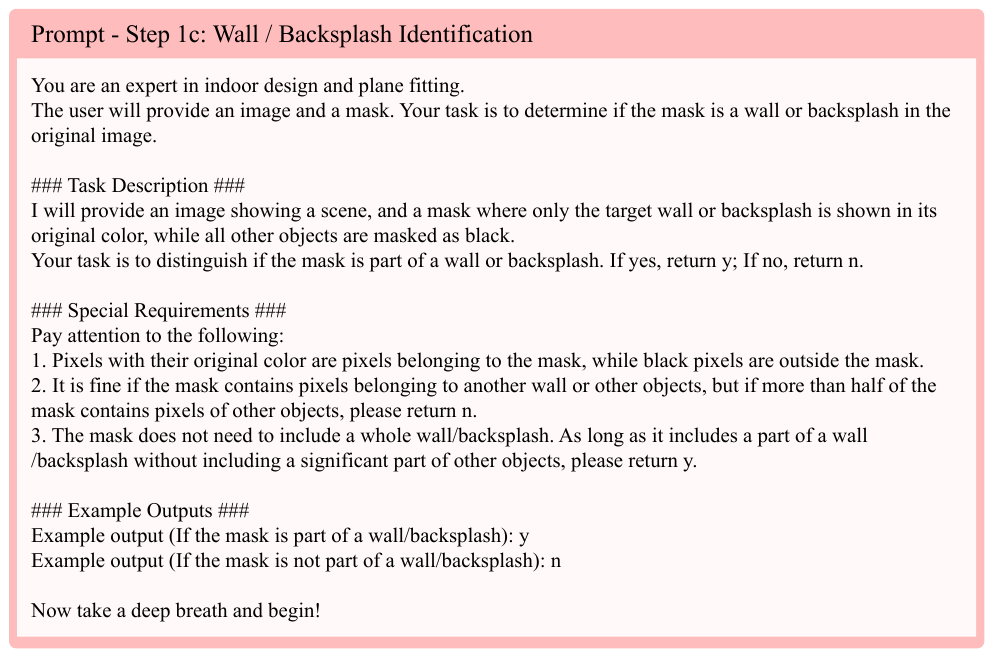}
    \caption{\textbf{Step~1c: Wall and backsplash identification.} The VLM determines which candidate planar regions correspond to walls or backsplashes.}
    \label{fig:prompt_step1c}
\end{figure}

\begin{figure}[h]
    \centering
    \includegraphics[width=1.0\linewidth]{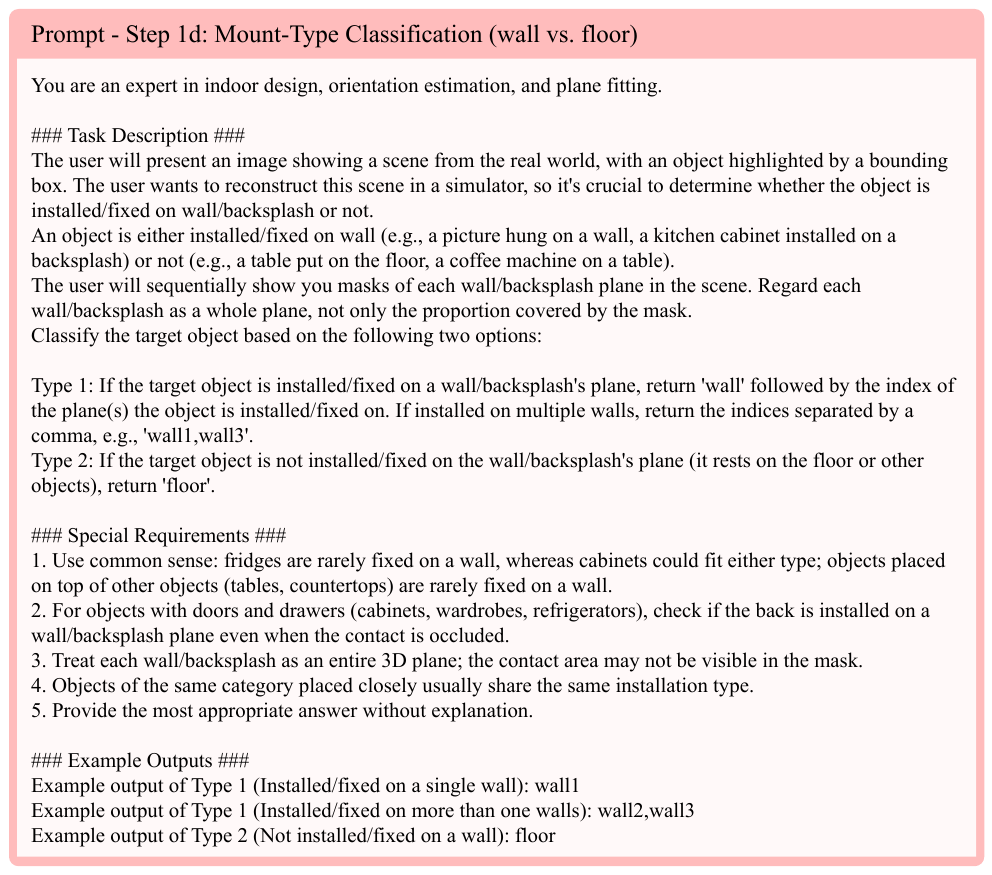}
    \caption{\textbf{Step~1d: Mount-type classification (wall vs.\ floor).} The VLM decides whether each object is supported by the floor or mounted on a wall.}
    \label{fig:prompt_step1d}
\end{figure}

\begin{figure}[h]
    \centering
    \includegraphics[width=1.0\linewidth]{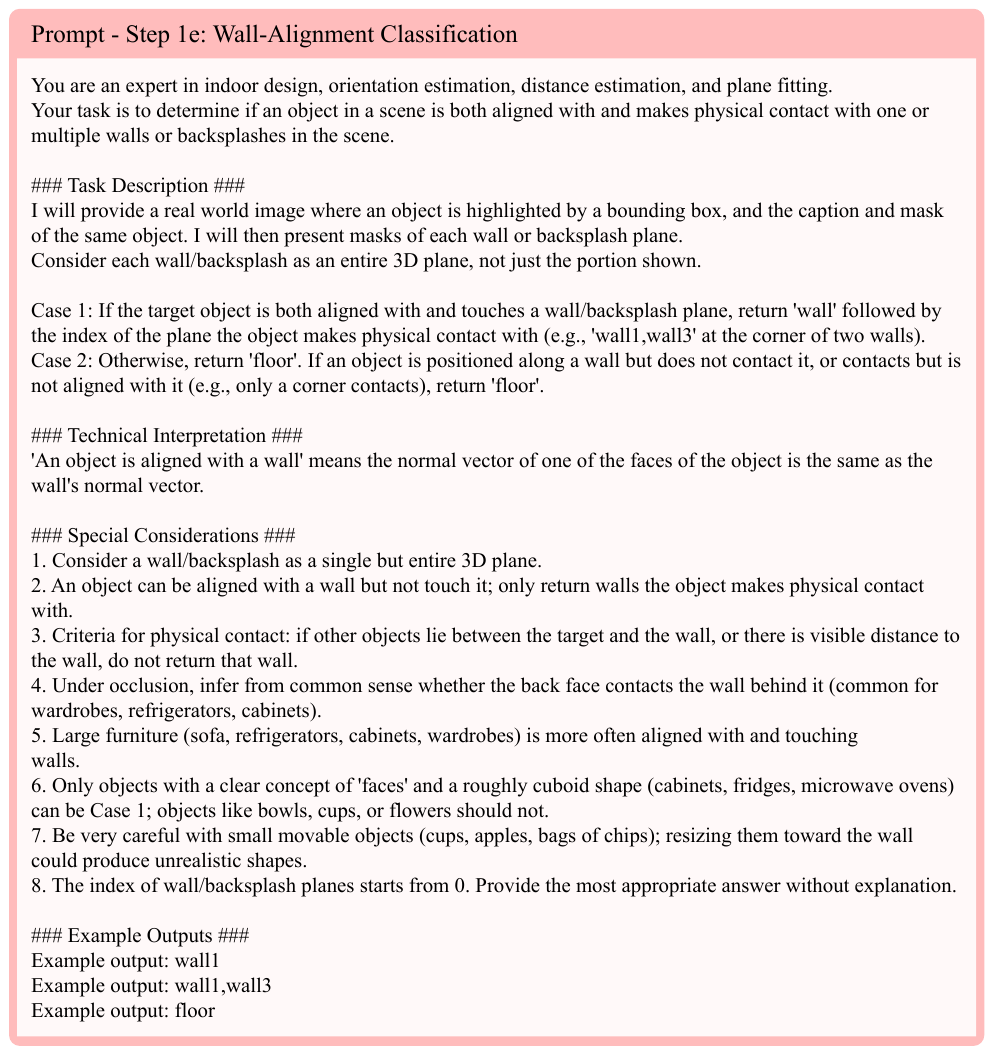}
    \caption{\textbf{Step~1e: Wall-alignment classification.} For wall-mounted objects, the VLM determines which wall the object is aligned against.}
    \label{fig:prompt_step1e}
\end{figure}

\begin{figure}[h]
    \centering
    \includegraphics[width=1.0\linewidth]{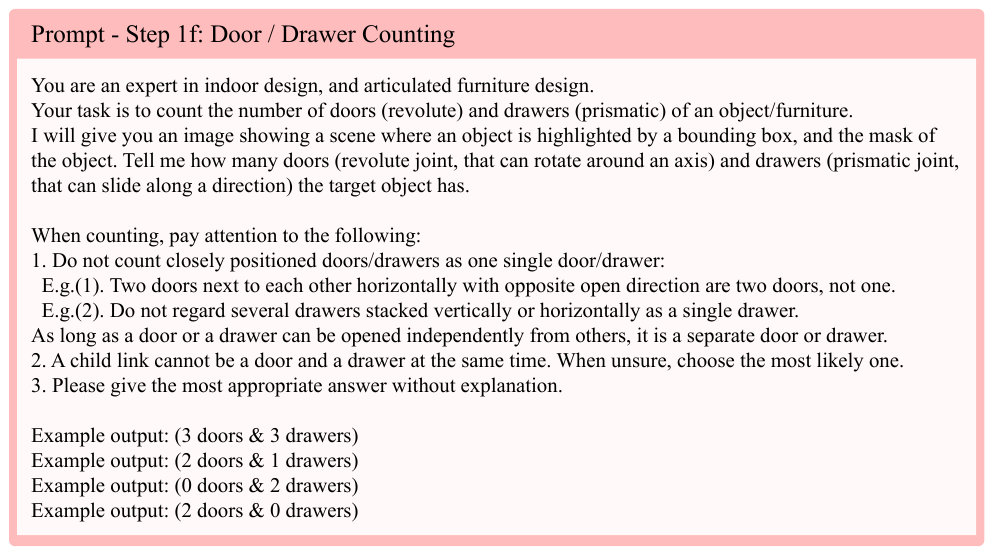}
    \caption{\textbf{Step~1f: Door and drawer counting.} For each articulated object, the VLM counts its doors and drawers.}
    \label{fig:prompt_step1f}
\end{figure}

\paragraph{Step~1: Real-World Extraction.}
Step~1 incrementally builds the scene graph through a sequence of VLM queries: it captions each detected object (1a, Figure~\ref{fig:prompt_step1a}), selects the best caption for every segmented object (1b, Figure~\ref{fig:prompt_step1b}), identifies wall and backsplash planes (1c, Figure~\ref{fig:prompt_step1c}), decides how each object is mounted (1d, Figure~\ref{fig:prompt_step1d}) and against which wall it is aligned (1e, Figure~\ref{fig:prompt_step1e}), and counts the doors and drawers of articulated objects (1f, Figure~\ref{fig:prompt_step1f}).

\begin{figure}[h]
    \centering
    \includegraphics[width=1.0\linewidth]{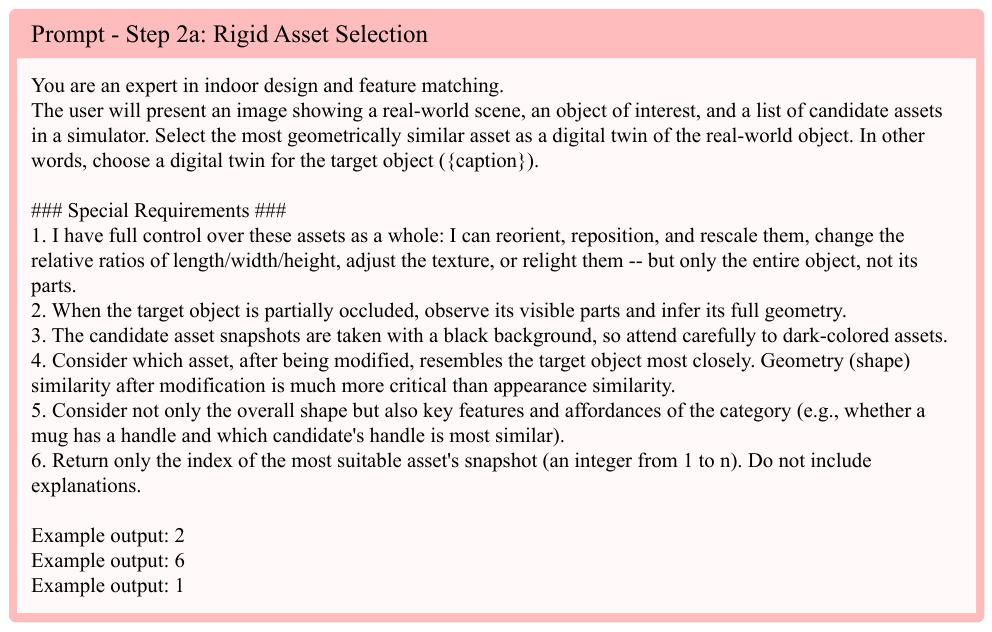}
    \caption{\textbf{Step~2a: Rigid asset selection.} From a set of candidate simulator assets, the VLM selects the one that best matches a rigid object in the observation.}
    \label{fig:prompt_step2a}
\end{figure}

\begin{figure}[h]
    \centering
    \includegraphics[width=1.0\linewidth]{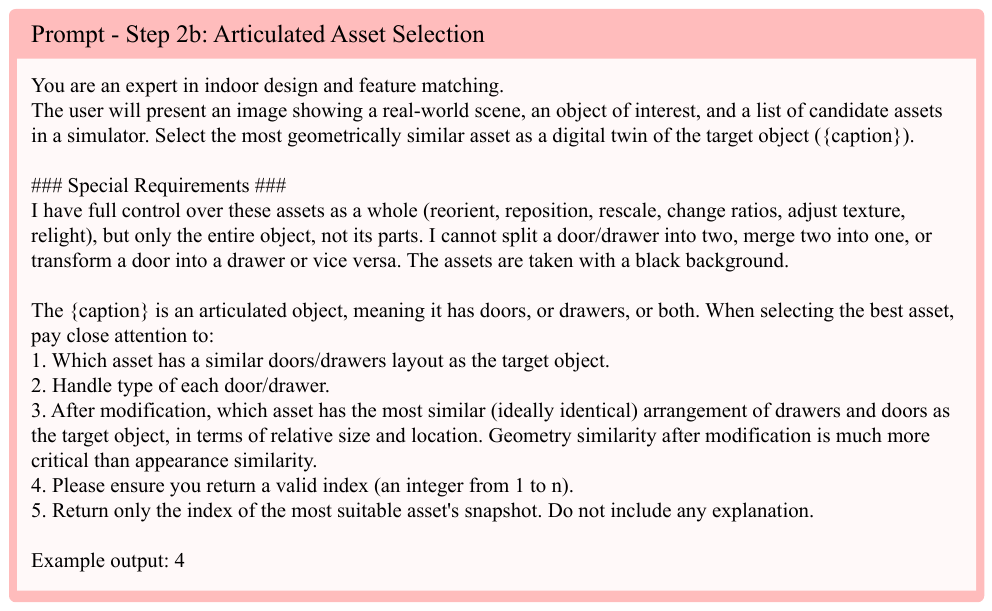}
    \caption{\textbf{Step~2b: Articulated asset selection.} The VLM selects the best-matching simulator asset for an articulated object.}
    \label{fig:prompt_step2b}
\end{figure}

\begin{figure}[h]
    \centering
    \includegraphics[width=1.0\linewidth]{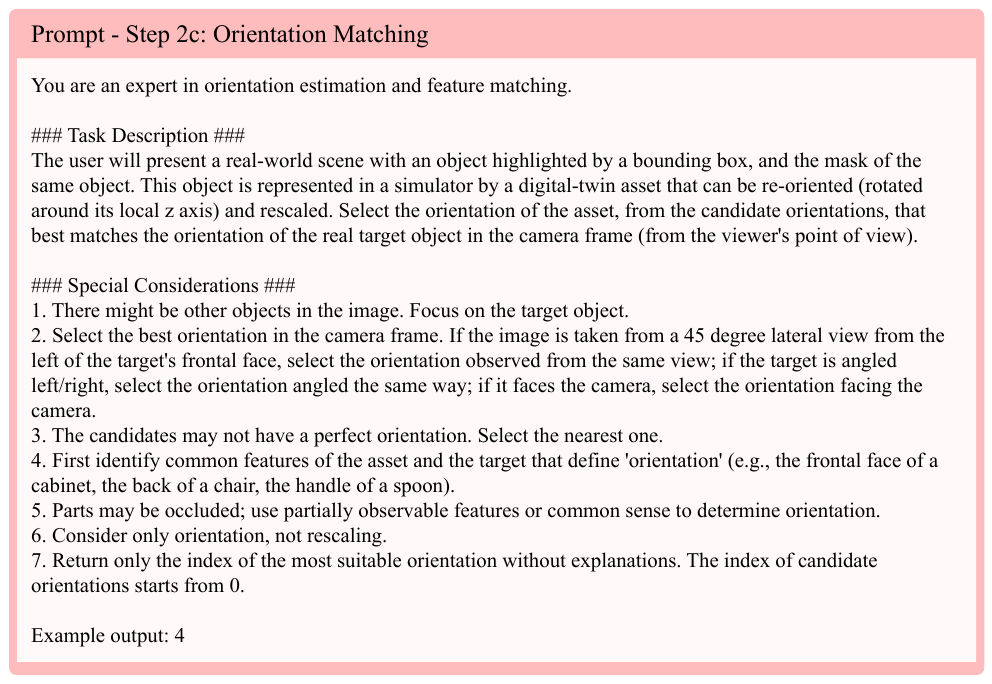}
    \caption{\textbf{Step~2c: Orientation matching.} The VLM selects the asset orientation that matches the object's pose in the camera frame.}
    \label{fig:prompt_step2c}
\end{figure}

\paragraph{Step~2: Digital Cousin Matching.}
For each extracted object, Step~2 retrieves the best-matching simulator asset, using separate prompts for rigid (2a, Figure~\ref{fig:prompt_step2a}) and articulated (2b, Figure~\ref{fig:prompt_step2b}) objects, and then selects the asset orientation matching the object's pose in the camera frame (2c, Figure~\ref{fig:prompt_step2c}).

\paragraph{Step~3: Simulated Scene Generation.}
The selected assets and the support/alignment relations from Steps~1-2 are composed into the final scene by a deterministic procedure (placement, support and collision resolution, and physics settling); this step uses no prompt.

\subsection{Demonstration Generation Prompt}
\label{sec:supp_demo_generation}

\begin{figure}[h]
    \centering
    \includegraphics[width=1.0\linewidth]{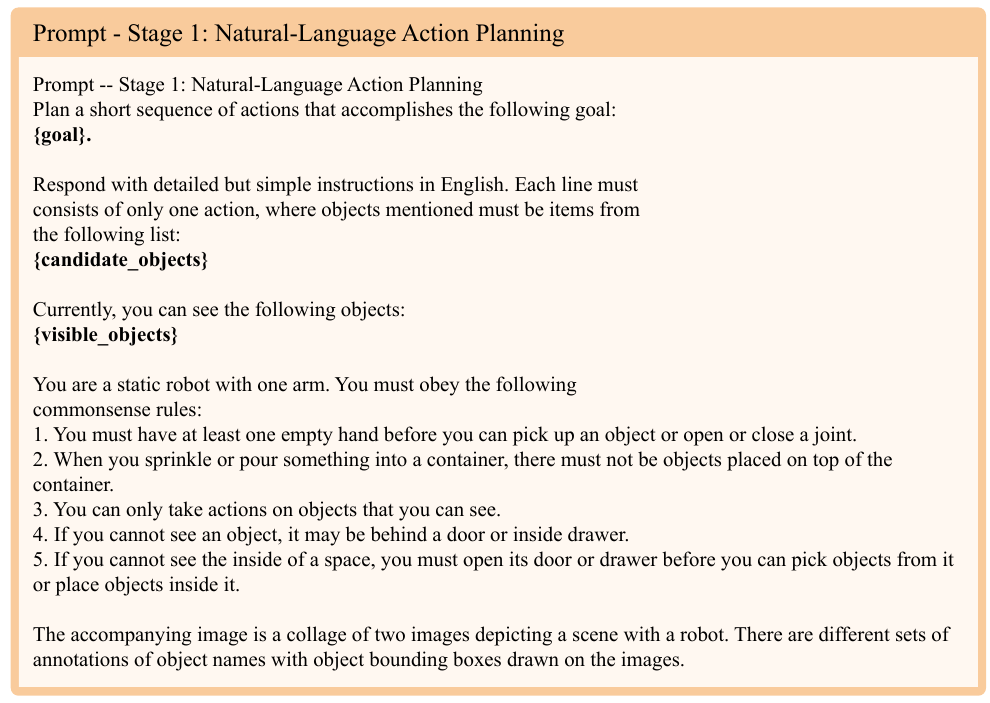}
    \caption{\textbf{Step~1: Natural-language action planning.} Given the task goal and the annotated scene image, the VLM is asked to produce a short, single-action-per-line plan that only references objects from the provided candidate list.}
    \label{fig:demo_generation_prompt_step1}
\end{figure}

\begin{figure}[h]
    \centering
    \includegraphics[width=1.0\linewidth]{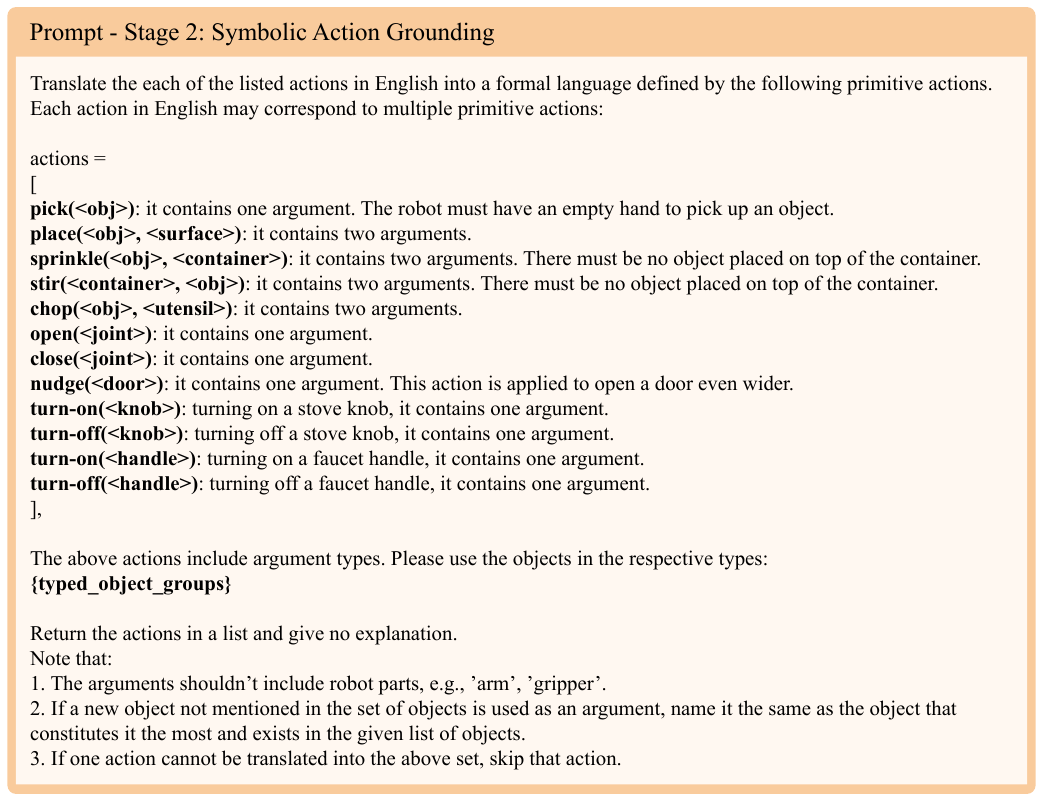}
    \caption{\textbf{Step~2: Symbolic action grounding.} The natural-language plan from Step~1 is translated into a sequence of typed primitive actions selected from a fixed action library, using only the objects listed under their respective argument types.}
    \label{fig:demo_generation_prompt_step2}
\end{figure}

\begin{figure}[h]
    \centering
    \includegraphics[width=1.0\linewidth]{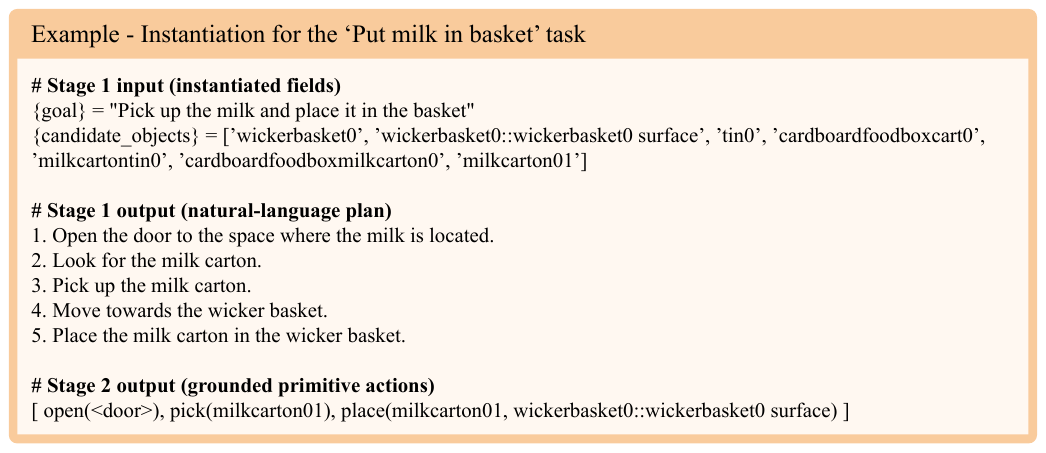}
    \caption{Example instantiation of the two-step prompt for the \textit{Put milk in basket} task, showing the filled-in input fields together with the natural-language plan and the grounded primitive-action sequence returned by the VLM.}
    \label{fig:prompt_example}
\end{figure}

To synthesize demonstrations automatically, PRISM queries the VLM inside the VLM-TAMP loop using a two-step prompting scheme.
The VLM is given the task goal together with a collage image of the scene, in which object names and bounding boxes are annotated.
In the first step, the VLM produces a high-level plan expressed in natural language (Figure~\ref{fig:demo_generation_prompt_step1}).
In the second step, this natural-language plan is grounded into a sequence of symbolic primitive actions drawn from a fixed action library (Figure~\ref{fig:demo_generation_prompt_step2}).
The grounded action sequence is then handed to the task-and-motion planner, and the resulting plan is replayed under randomized object poses and robot initial joint configurations to produce the recorded demonstrations.
In the prompts below, fields enclosed in \texttt{\{\,\}} are instantiated per task and per scene, while all remaining text is fixed.
Figure~\ref{fig:prompt_example} shows a concrete instantiation and the corresponding VLM outputs for the representative \textit{Put milk in basket} task.

%% file: references.bib
@inproceedings{mimicgen,
    title={MimicGen: A Data Generation System for Scalable Robot Learning using Human Demonstrations},
    author={Mandlekar, Ajay and Nasiriany, Soroush and Wen, Bowen and Akinola, Iretiayo and Narang, Yashraj and Fan, Linxi and Zhu, Yuke and Fox, Dieter},
    booktitle={7th Annual Conference on Robot Learning},
    year={2023}
}

@inproceedings{gensim2,
  title={GenSim2: Scaling Robot Data Generation with Multi-modal and Reasoning LLMs},
  author={Hua, Pu and Liu, Minghuan and Macaluso, Annabella and Lin, Yunfeng and Zhang, Weinan and Xu, Huazhe and Wang, Lirui},
  booktitle={8th Annual Conference on Robot Learning},
  year={2024}
}

@INPROCEEDINGS{demogen, 
    AUTHOR    = {Zhengrong Xue AND Shuying Deng AND Zhenyang Chen AND Yixuan Wang AND Zhecheng Yuan AND Huazhe Xu}, 
    TITLE     = {{DemoGen: Synthetic Demonstration Generation for Data-Efficient Visuomotor Policy Learning}}, 
    BOOKTITLE = {Proceedings of Robotics: Science and Systems}, 
    YEAR      = {2025}, 
    ADDRESS   = {LosAngeles, CA, USA}, 
    MONTH     = {June}, 
    DOI       = {10.15607/RSS.2025.XXI.157} 
}

@inproceedings{skillmimicgen,
    title={SkillMimicGen: Automated Demonstration Generation for Efficient Skill Learning and Deployment},
    author={Garrett, Caelan and Mandlekar, Ajay and Wen, Bowen and Fox, Dieter},
    booktitle={8th Annual Conference on Robot Learning},
    year={2024}
}

@article{dreamgen,
  title={DreamGen: Unlocking Generalization in Robot Learning through Video World Models},
  author={Jang, Joel and Ye, Seonghyeon and Lin, Zongyu and Xiang, Jiannan and Bjorck, Johan and Fang, Yu and Hu, Fengyuan and Huang, Spencer and Kundalia, Kaushil and Lin, Yen-Chen and others},
  journal={9th Annual Conference on Robot Learning},
  year={2025}
}

@inproceedings{dexmimicgen,
  title={Dexmimicgen: Automated data generation for bimanual dexterous manipulation via imitation learning},
  author={Jiang, Zhenyu and Xie, Yuqi and Lin, Kevin and Xu, Zhenjia and Wan, Weikang and Mandlekar, Ajay and Fan, Linxi Jim and Zhu, Yuke},
  booktitle={2025 IEEE International Conference on Robotics and Automation (ICRA)},
  pages={16923--16930},
  year={2025},
  organization={IEEE}
}

@article{robotwin2.0,
  title={Robotwin 2.0: A scalable data generator and benchmark with strong domain randomization for robust bimanual robotic manipulation},
  author={Chen, Tianxing and Chen, Zanxin and Chen, Baijun and Cai, Zijian and Liu, Yibin and Li, Zixuan and Liang, Qiwei and Lin, Xianliang and Ge, Yiheng and Gu, Zhenyu and others},
  journal={arXiv preprint arXiv:2506.18088},
  year={2025}
}

@article{acdc,
  title={Automated creation of digital cousins for robust policy learning},
  author={Dai, Tianyuan and Wong, Josiah and Jiang, Yunfan and Wang, Chen and Gokmen, Cem and Zhang, Ruohan and Wu, Jiajun and Fei-Fei, Li},
  journal={8th Annual Conference on Robot Learning},
  year={2024}
}

@inproceedings{robogen,
title={RoboGen: Towards Unleashing Infinite Data for Automated Robot Learning via Generative Simulation},
author={Yufei Wang and Zhou Xian and Feng Chen and Tsun-Hsuan Wang and Yian Wang and Katerina Fragkiadaki and Zackory Erickson and David Held and Chuang Gan},
booktitle={Forty-first Int. Conf. on Machine Learning},
year={2024},
}

@inproceedings{robotwin1.0,
  title={Robotwin: Dual-arm robot benchmark with generative digital twins},
  author={Mu, Yao and Chen, Tianxing and Chen, Zanxin and Peng, Shijia and Lan, Zhiqian and Gao, Zeyu and Liang, Zhixuan and Yu, Qiaojun and Zou, Yude and Xu, Mingkun and others},
  booktitle={Proceedings of the Computer Vision and Pattern Recognition Conference},
  pages={27649--27660},
  year={2025}
}

@inproceedings{humanoidgen,
    title={HumanoidGen: Data Generation for Bimanual Dexterous Manipulation via {LLM} Reasoning},
    author={Zhi Jing and Siyuan Yang and Jicong Ao and Ting Xiao and Yu-Gang Jiang and Chenjia Bai},
    booktitle={The Thirty-ninth Annual Conference on Neural Information Processing Systems},
    year={2025},
    url={https://openreview.net/forum?id=Mk9ykil8eP}
}

@inproceedings{gensim,
    title={GenSim: Generating Robotic Simulation Tasks via Large Language Models},
    author={Lirui Wang and Yiyang Ling and Zhecheng Yuan and Mohit Shridhar and Chen Bao and Yuzhe Qin and Bailin Wang and Huazhe Xu and Xiaolong Wang},
    booktitle={The Twelfth International Conference on Learning Representations},
    year={2024},
    url={https://openreview.net/forum?id=OI3RoHoWAN}
}

@inproceedings{rt2,
    title={{RT}-2: Vision-Language-Action Models Transfer Web Knowledge to Robotic Control},
    author={Brianna Zitkovich and Tianhe Yu and Sichun Xu and Peng Xu and Ted Xiao and Fei Xia and Jialin Wu and Paul Wohlhart and Stefan Welker and Ayzaan Wahid and Quan Vuong and Vincent Vanhoucke and Huong Tran and Radu Soricut and Anikait Singh and Jaspiar Singh and Pierre Sermanet and Pannag R Sanketi and Grecia Salazar and Michael S Ryoo and Krista Reymann and Kanishka Rao and Karl Pertsch and Igor Mordatch and Henryk Michalewski and Yao Lu and Sergey Levine and Lisa Lee and Tsang-Wei Edward Lee and Isabel Leal and Yuheng Kuang and Dmitry Kalashnikov and Ryan Julian and Nikhil J Joshi and Alex Irpan and brian ichter and Jasmine Hsu and Alexander Herzog and Karol Hausman and Keerthana Gopalakrishnan and Chuyuan Fu and Pete Florence and Chelsea Finn and Kumar Avinava Dubey and Danny Driess and Tianli Ding and Krzysztof Marcin Choromanski and Xi Chen and Yevgen Chebotar and Justice Carbajal and Noah Brown and Anthony Brohan and Montserrat Gonzalez Arenas and Kehang Han},
    booktitle={7th Annual Conference on Robot Learning},
    year={2023},
    url={https://openreview.net/forum?id=XMQgwiJ7KSX}
}

@inproceedings{openx,
  title={Open x-embodiment: Robotic learning datasets and rt-x models: Open x-embodiment collaboration 0},
  author={O’Neill, Abby and Rehman, Abdul and Maddukuri, Abhiram and Gupta, Abhishek and Padalkar, Abhishek and Lee, Abraham and Pooley, Acorn and Gupta, Agrim and Mandlekar, Ajay and Jain, Ajinkya and others},
  booktitle={2024 IEEE Int. Conf. on Robotics and Automation (ICRA)},
  pages={6892--6903},
  year={2024},
  organization={IEEE}
}

@inproceedings{openvla,
    title={Open{VLA}: An Open-Source Vision-Language-Action Model},
    author={Moo Jin Kim and Karl Pertsch and Siddharth Karamcheti and Ted Xiao and Ashwin Balakrishna and Suraj Nair and Rafael Rafailov and Ethan P Foster and Pannag R Sanketi and Quan Vuong and Thomas Kollar and Benjamin Burchfiel and Russ Tedrake and Dorsa Sadigh and Sergey Levine and Percy Liang and Chelsea Finn},
    booktitle={8th Annual Conference on Robot Learning},
    year={2024},
}

@inproceedings{pi05,
    title={\${\textbackslash}pi\_\{0.5\}\$: a Vision-Language-Action Model with Open-World Generalization},
    author={Kevin Black and Noah Brown and James Darpinian and Karan Dhabalia and Danny Driess and Adnan Esmail and Michael Robert Equi and Chelsea Finn and Niccolo Fusai and Manuel Y. Galliker and Dibya Ghosh and Lachy Groom and Karol Hausman and Brian Ichter and Szymon Jakubczak and Tim Jones and Liyiming Ke and Devin LeBlanc and Sergey Levine and Adrian Li-Bell and Mohith Mothukuri and Suraj Nair and Karl Pertsch and Allen Z. Ren and Lucy Xiaoyang Shi and Laura Smith and Jost Tobias Springenberg and Kyle Stachowicz and James Tanner and Quan Vuong and Homer Walke and Anna Walling and Haohuan Wang and Lili Yu and Ury Zhilinsky},
    booktitle={9th Annual Conference on Robot Learning},
    year={2025},
    url={https://openreview.net/forum?id=vlhoswksBO}
}

@INPROCEEDINGS{rt1, 
    AUTHOR    = {Anthony Brohan AND Noah Brown AND Justice Carbajal AND Yevgen Chebotar AND Joseph Dabis AND Chelsea Finn AND Keerthana Gopalakrishnan AND Karol Hausman AND Alexander Herzog AND Jasmine Hsu AND Julian Ibarz AND Brian Ichter AND Alex Irpan AND Tomas Jackson AND Sally Jesmonth AND Nikhil Joshi AND Ryan Julian AND Dmitry Kalashnikov AND Yuheng Kuang AND Isabel Leal AND Kuang-Huei Lee AND Sergey Levine AND Yao Lu AND Utsav Malla AND Deeksha Manjunath AND Igor Mordatch AND Ofir Nachum AND Carolina Parada AND Jodilyn  Peralta AND Emily Perez AND Karl Pertsch AND Jornell  Quiambao AND Kanishka Rao AND Michael S Ryoo AND Grecia  Salazar AND Pannag R Sanketi AND Kevin  Sayed AND Jaspiar  Singh AND Sumedh  Sontakke AND Austin  Stone AND Clayton  Tan AND Huong  Tran AND Vincent Vanhoucke AND Steve  Vega AND Quan H Vuong AND Fei Xia AND Ted Xiao AND Peng Xu AND Sichun Xu AND Tianhe Yu AND Brianna  Zitkovich}, 
    TITLE     = {{RT-1: Robotics Transformer for Real-World Control at Scale}}, 
    BOOKTITLE = {Proceedings of Robotics: Science and Systems}, 
    YEAR      = {2023}, 
    ADDRESS   = {Daegu, Republic of Korea}, 
    MONTH     = {July}, 
    DOI       = {10.15607/RSS.2023.XIX.025} 
}

@inproceedings{octo,
    title={Octo: An Open-Source Generalist Robot Policy},
    author = {{Octo Model Team} and Dibya Ghosh and Homer Walke and Karl Pertsch and Kevin Black and Oier Mees and Sudeep Dasari and Joey Hejna and Charles Xu and Jianlan Luo and Tobias Kreiman and {You Liang} Tan and Lawrence Yunliang Chen and Pannag Sanketi and Quan Vuong and Ted Xiao and Dorsa Sadigh and Chelsea Finn and Sergey Levine},
    booktitle = {Proceedings of Robotics: Science and Systems},
    address  = {Delft, Netherlands},
    year = {2024},
}

@inproceedings{libero,
title={{LIBERO}: Benchmarking Knowledge Transfer for Lifelong Robot Learning},
author={Bo Liu and Yifeng Zhu and Chongkai Gao and Yihao Feng and qiang liu and Yuke Zhu and Peter Stone},
booktitle={Thirty-seventh Conference on Neural Information Processing Systems Datasets and Benchmarks Track},
year={2023},
url={https://openreview.net/forum?id=xzEtNSuDJk}
}

@inproceedings{droid,
    title={{DROID}: A Large-Scale In-The-Wild Robot Manipulation Dataset},
    author={Alexander Khazatsky and Karl Pertsch and Suraj Nair and Ashwin Balakrishna and Sudeep Dasari and Siddharth Karamcheti and Soroush Nasiriany and Mohan Kumar Srirama and Lawrence Yunliang Chen and Kirsty Ellis and Peter David Fagan and Joey Hejna and Masha Itkina and Marion Lepert and Yecheng Jason Ma and Patrick Tree Miller and Jimmy Wu and Suneel Belkhale and Shivin Dass and Huy Ha and Arhan Jain and Abraham Lee and Youngwoon Lee and Marius Memmel and Sungjae Park and Ilija Radosavovic and Kaiyuan Wang and Albert Zhan and Kevin Black and Cheng Chi and Kyle Beltran Hatch and Shan Lin and Jingpei Lu and Jean Mercat and Abdul Rehman and Pannag R Sanketi and Archit Sharma and Cody Simpson and Quan Vuong and Homer Rich Walke and Blake Wulfe and Ted Xiao and Jonathan Heewon Yang and Arefeh Yavary and Tony Z. Zhao and Christopher Agia and Rohan Baijal and Mateo Guaman Castro and Daphne Chen and Qiuyu Chen and Trinity Chung and Jaimyn Drake and Ethan Paul Foster and Jensen Gao and David Antonio Herrera and Minho Heo and Kyle Hsu and Jiaheng Hu and Donovon Jackson and Charlotte Le and Yunshuang Li and Xinyu Lin and Zehan Ma and Abhiram Maddukuri and Suvir Mirchandani and Daniel Morton and Tony Khuong Nguyen and Abigail O'Neill and Rosario Scalise and Derick Seale and Victor Son and Stephen Tian and Emi Tran and Andrew E. Wang and Yilin Wu and Annie Xie and Jingyun Yang and Patrick Yin and Yunchu Zhang and Osbert Bastani and Glen Berseth and Jeannette Bohg and Ken Goldberg and Abhinav Gupta and Abhishek Gupta and Dinesh Jayaraman and Joseph J Lim and Jitendra Malik and Roberto Mart{\'\i}n-Mart{\'\i}n and Subramanian Ramamoorthy and Dorsa Sadigh and Shuran Song and Jiajun Wu and Michael C. Yip and Yuke Zhu and Thomas Kollar and Sergey Levine and Chelsea Finn},
    booktitle={Proceedings of Robotics: Science and Systems},
    year={2024},
    ADDRESS   = {Delft, Netherlands},
    url={https://openreview.net/forum?id=Ml2pTYLNLi}
}

@article{libero-plus,
    title={LIBERO-Plus: In-depth Robustness Analysis of Vision-Language-Action Models},
    author={Senyu Fei and Siyin Wang and Junhao Shi and Zihao Dai and Jikun Cai and Pengfang Qian and Li Ji and Xinzhe He and Shiduo Zhang and Zhaoye Fei and Jinlan Fu and Jingjing Gong and Xipeng Qiu},
    journal = {arXiv preprint arXiv:2510.13626},
    year={2025},
}

@article{dp,
  title={Diffusion policy: Visuomotor policy learning via action diffusion},
  author={Chi, Cheng and Xu, Zhenjia and Feng, Siyuan and Cousineau, Eric and Du, Yilun and Burchfiel, Benjamin and Tedrake, Russ and Song, Shuran},
  journal={The International Journal of Robotics Research},
  volume={44},
  number={10-11},
  pages={1684--1704},
  year={2025},
  publisher={Sage Publications Sage UK: London, England}
}

@inproceedings{roboturk,
  title={Scaling robot supervision to hundreds of hours with roboturk: Robotic manipulation dataset through human reasoning and dexterity},
  author={Mandlekar, Ajay and Booher, Jonathan and Spero, Max and Tung, Albert and Gupta, Anchit and Zhu, Yuke and Garg, Animesh and Savarese, Silvio and Fei-Fei, Li},
  booktitle={2019 IEEE/RSJ International Conference on Intelligent Robots and Systems (IROS)},
  pages={1048--1055},
  year={2019},
  organization={IEEE}
}

@inproceedings{urdformer,
    title={{URDF}ormer: Constructing interactive Realistic Scenes from Real Images via Simulation and Generative Modeling},
    author={Qiuyu Chen and Marius Memmel and Alex Fang and Aaron Walsman and Dieter Fox and Abhishek Gupta},
    booktitle={Towards Generalist Robots: Learning Paradigms for Scalable Skill Acquisition @ CoRL2023},
    year={2023},
}

@inproceedings{xsim,
    title={X-Sim: Cross-Embodiment Learning via Real-to-Sim-to-Real},
    author={Prithwish Dan and Kushal Kedia and Angela Chao and Edward Duan and Maximus Adrian Pace and Wei-Chiu Ma and Sanjiban Choudhury},
    booktitle={9th Annual Conference on Robot Learning},
    year={2025},
    url={https://openreview.net/forum?id=BO7qo66YJ2}
}

@inproceedings{dp3,
    title={3D Diffusion Policy: Generalizable Visuomotor Policy Learning via Simple 3D Representations},
    author = {Yanjie Ze AND Gu Zhang AND Kangning Zhang AND Chenyuan Hu AND Muhan Wang AND Huazhe Xu},
    booktitle = {Proceedings of Robotics: Science and Systems},
    address  = {Delft, Netherlands},
    year = {2024},
}

@article{gr00t,
  title={Gr00t n1: An open foundation model for generalist humanoid robots},
  author={Bjorck, Johan and Casta{\~n}eda, Fernando and Cherniadev, Nikita and Da, Xingye and Ding, Runyu and Fan, Linxi and Fang, Yu and Fox, Dieter and Hu, Fengyuan and Huang, Spencer and others},
  journal={arXiv preprint arXiv:2503.14734},
  year={2025}
}

@INPROCEEDINGS{umi, 
    AUTHOR    = {Cheng Chi AND Zhenjia Xu AND Chuer Pan AND Eric Cousineau AND Benjamin Burchfiel AND Siyuan Feng AND Russ Tedrake AND Shuran Song}, 
    TITLE     = {{Universal Manipulation Interface: In-The-Wild Robot Teaching Without In-The-Wild Robots}}, 
    BOOKTITLE = {Proceedings of Robotics: Science and Systems}, 
    YEAR      = {2024}, 
    ADDRESS   = {Delft, Netherlands}, 
    MONTH     = {July}, 
    DOI       = {10.15607/RSS.2024.XX.045} 
}

@INPROCEEDINGS{openvla-oft, 
    AUTHOR    = {Moo Jin Kim AND Chelsea Finn AND Percy Liang}, 
    TITLE     = {{Fine-Tuning Vision-Language-Action Models: Optimizing Speed and Success}}, 
    BOOKTITLE = {Proceedings of Robotics: Science and Systems}, 
    YEAR      = {2025}, 
    ADDRESS   = {LosAngeles, CA, USA}, 
    MONTH     = {June}, 
    DOI       = {10.15607/RSS.2025.XXI.017} 
}

@article{gaia,
  title={GAIA: Generating Task Instruction Aware Simulation Grounded in Real Contexts Using Vision-Language Models},
  author={Ko, Dogyu and Yeo, Chanyoung and Kim, Daeho and Kim, Jaeho and Hwang, Hyoseok},
  journal={IEEE Robotics and Automation Letters},
  year={2025},
  publisher={IEEE}
}

@article{grounded-sam,
      title={Grounded SAM: Assembling Open-World Models for Diverse Visual Tasks}, 
      author={Tianhe Ren and Shilong Liu and Ailing Zeng and Jing Lin and Kunchang Li and He Cao and Jiayu Chen and Xinyu Huang and Yukang Chen and Feng Yan and Zhaoyang Zeng and Hao Zhang and Feng Li and Jie Yang and Hongyang Li and Qing Jiang and Lei Zhang},
      year={2024},
      eprint={2401.14159},
      archivePrefix={arXiv},
      primaryClass={cs.CV}
}

@article{depth_anything,
  title={Depth anything v2},
  author={Yang, Lihe and Kang, Bingyi and Huang, Zilong and Zhao, Zhen and Xu, Xiaogang and Feng, Jiashi and Zhao, Hengshuang},
  journal={Advances in Neural Information Processing Systems},
  volume={37},
  pages={21875--21911},
  year={2024}
}

@InProceedings{perspective-fields,
    author    = {Jin, Linyi and Zhang, Jianming and Hold-Geoffroy, Yannick and Wang, Oliver and Blackburn-Matzen, Kevin and Sticha, Matthew and Fouhey, David F.},
    title     = {Perspective Fields for Single Image Camera Calibration},
    booktitle = {Proceedings of the IEEE/CVF Conference on Computer Vision and Pattern Recognition (CVPR)},
    month     = {June},
    year      = {2023},
    pages     = {17307-17316}
}

@inproceedings{behavior,
title={{BEHAVIOR}-1K: A Benchmark for Embodied {AI} with 1,000 Everyday Activities and Realistic Simulation},
author={Chengshu Li and Ruohan Zhang and Josiah Wong and Cem Gokmen and Sanjana Srivastava and Roberto Mart{\'\i}n-Mart{\'\i}n and Chen Wang and Gabrael Levine and Michael Lingelbach and Jiankai Sun and Mona Anvari and Minjune Hwang and Manasi Sharma and Arman Aydin and Dhruva Bansal and Samuel Hunter and Kyu-Young Kim and Alan Lou and Caleb R Matthews and Ivan Villa-Renteria and Jerry Huayang Tang and Claire Tang and Fei Xia and Silvio Savarese and Hyowon Gweon and Karen Liu and Jiajun Wu and Li Fei-Fei},
booktitle={6th Annual Conference on Robot Learning},
year={2022},
url={https://openreview.net/forum?id=_8DoIe8G3t}
}

@article{
dinov2,
title={{DINO}v2: Learning Robust Visual Features without Supervision},
author={Maxime Oquab and Timoth{\'e}e Darcet and Th{\'e}o Moutakanni and Huy V. Vo and Marc Szafraniec and Vasil Khalidov and Pierre Fernandez and Daniel HAZIZA and Francisco Massa and Alaaeldin El-Nouby and Mido Assran and Nicolas Ballas and Wojciech Galuba and Russell Howes and Po-Yao Huang and Shang-Wen Li and Ishan Misra and Michael Rabbat and Vasu Sharma and Gabriel Synnaeve and Hu Xu and Herve Jegou and Julien Mairal and Patrick Labatut and Armand Joulin and Piotr Bojanowski},
journal={Transactions on Machine Learning Research},
issn={2835-8856},
year={2024},
note={Featured Certification}
}

@INPROCEEDINGS{vlm-tamp,
  author={Yang, Zhutian and Garrett, Caelan and Fox, Dieter and Lozano-Pérez, Tomás and Kaelbling, Leslie Pack},
  booktitle={2025 IEEE International Conference on Robotics and Automation (ICRA)}, 
  title={Guiding Long-Horizon Task and Motion Planning with Vision Language Models}, 
  year={2025},
  volume={},
  number={},
  pages={16847-16853},
  doi={10.1109/ICRA55743.2025.11128705}}

@INPROCEEDINGS{piginet, 
    AUTHOR    = {Zhutian  Yang AND Caelan R Garrett AND Tomas Lozano-Perez AND Leslie Kaelbling AND Dieter Fox}, 
    TITLE     = {{Sequence-Based Plan Feasibility Prediction for Efficient Task and Motion Planning}}, 
    BOOKTITLE = {Proceedings of Robotics: Science and Systems}, 
    YEAR      = {2023}, 
    ADDRESS   = {Daegu, Republic of Korea}, 
    MONTH     = {July}, 
    DOI       = {10.15607/RSS.2023.XIX.061} 
}

@InProceedings{sapien,
author = {Xiang, Fanbo and Qin, Yuzhe and Mo, Kaichun and Xia, Yikuan and Zhu, Hao and Liu, Fangchen and Liu, Minghua and Jiang, Hanxiao and Yuan, Yifu and Wang, He and Yi, Li and Chang, Angel X. and Guibas, Leonidas J. and Su, Hao},
title = {{SAPIEN}: A SimulAted Part-based Interactive ENvironment},
booktitle = {The IEEE Conference on Computer Vision and Pattern Recognition (CVPR)},
month = {June},
year = {2020}}

@inproceedings{diffusionpolicy,
	title={Diffusion Policy: Visuomotor Policy Learning via Action Diffusion},
	author={Chi, Cheng and Feng, Siyuan and Du, Yilun and Xu, Zhenjia and Cousineau, Eric and Burchfiel, Benjamin and Song, Shuran},
	booktitle={Proceedings of Robotics: Science and Systems (RSS)},
	year={2023}}

@inproceedings{compare_ft_ml,
  title={Comparing the efficacy of fine-tuning and meta-learning for few-shot policy imitation},
  author={Patacchiola, Massimiliano and Sun, Mingfei and Hofmann, Katja and Turner, Richard E},
  booktitle={Conference on Lifelong Learning Agents},
  pages={878--908},
  year={2023},
  organization={PMLR}
}

@inproceedings{gen2sim,
  title={Gen2sim: Scaling up robot learning in simulation with generative models},
  author={Katara, Pushkal and Xian, Zhou and Fragkiadaki, Katerina},
  booktitle={2024 IEEE International Conference on Robotics and Automation (ICRA)},
  pages={6672--6679},
  year={2024},
  organization={IEEE}
}

@article{DDIM,
  title={Denoising diffusion implicit models},
  author={Song, Jiaming and Meng, Chenlin and Ermon, Stefano},
  journal={arXiv preprint arXiv:2010.02502},
  year={2020}
}

@inproceedings{FiLM,
  title={Film: Visual reasoning with a general conditioning layer},
  author={Perez, Ethan and Strub, Florian and De Vries, Harm and Dumoulin, Vincent and Courville, Aaron},
  booktitle={Proceedings of the AAAI conference on artificial intelligence},
  volume={32},
  number={1},
  year={2018}
}

@inproceedings{lora,
title={Lo{RA}: Low-Rank Adaptation of Large Language Models},
author={Edward J Hu and Yelong Shen and Phillip Wallis and Zeyuan Allen-Zhu and Yuanzhi Li and Shean Wang and Lu Wang and Weizhu Chen},
booktitle={International Conference on Learning Representations},
year={2022},
url={https://openreview.net/forum?id=nZeVKeeFYf9}
}

@inproceedings{mujoco,
  title={MuJoCo: A physics engine for model-based control},
  author={Todorov, Emanuel and Erez, Tom and Tassa, Yuval},
  booktitle={2012 IEEE/RSJ International Conference on Intelligent Robots and Systems},
  pages={5026--5033},
  year={2012},
  organization={IEEE},
  doi={10.1109/IROS.2012.6386109}
}

@inproceedings{gaussiansplatting2d,
    title={2D Gaussian Splatting for Geometrically Accurate Radiance Fields},
    author={Huang, Binbin and Yu, Zehao and Chen, Anpei and Geiger, Andreas and Gao, Shenghua},
    publisher = {Association for Computing Machinery},
    booktitle = {SIGGRAPH 2024 Conference Papers},
    year      = {2024},
    doi       = {10.1145/3641519.3657428}
}

@InProceedings{foundationpose,
    author    = {Wen, Bowen and Yang, Wei and Kautz, Jan and Birchfield, Stan},
    title     = {FoundationPose: Unified 6D Pose Estimation and Tracking of Novel Objects},
    booktitle = {Proceedings of the IEEE/CVF Conference on Computer Vision and Pattern Recognition (CVPR)},
    month     = {June},
    year      = {2024},
    pages     = {17868-17879}
}

@article{ppo,
  title={Proximal policy optimization algorithms},
  author={Schulman, John and Wolski, Filip and Dhariwal, Prafulla and Radford, Alec and Klimov, Oleg},
  journal={arXiv preprint arXiv:1707.06347},
  year={2017}
}

@misc{isaacsim,
  author = {{NVIDIA}},
  title = {{NVIDIA Isaac Sim}},
  howpublished = {\url{https://developer.nvidia.com/isaac-sim}},
  year = {2025},
  note = {Accessed: 2026-05-01}
}

@misc{rodin,
  author       = {{Deemos}},
  title        = {{Rodin}},
  howpublished = {\url{https://hyperhuman.deemos.com/rodin}},
  year         = {2024},
  note         = {Accessed: 2026-05-01}
}

@misc{polycam,
  author       = {{Polycam}},
  title        = {{Polycam}},
  year         = {2020},
  howpublished = {\url{https://poly.cam}},
  note         = {Accessed: 20226-04-23}
}

@Manual{blender,
  title        = {Blender - a 3D modelling and rendering package},
  author       = {{Blender Online Community}},
  organization = {Blender Foundation},
  address      = {Stichting Blender Foundation, Amsterdam},
  year         = {2024},
  url          = {http://www.blender.org}
}

@inproceedings{siglip,
  title={Sigmoid loss for language image pre-training},
  author={Zhai, Xiaohua and Mustafa, Basil and Kolesnikov, Alexander and Beyer, Lucas},
  booktitle={Proceedings of the IEEE/CVF international conference on computer vision},
  pages={11975--11986},
  year={2023}
}
